\documentclass[letterpaper]{article}
\usepackage[preprint]{aaai2027}
\usepackage[hyphens]{url}
\usepackage{graphicx}
\usepackage{natbib}
\usepackage{caption}
\usepackage{makecell}
\usepackage{multirow}
\usepackage{amsmath,amssymb,amsfonts}
\usepackage{amsthm}
\usepackage{mathrsfs}
\usepackage{xcolor}
\usepackage{colortbl}
\usepackage{textcomp}
\usepackage{manyfoot}
\usepackage{booktabs}
\usepackage{algorithm}
\usepackage{algorithmicx}
\usepackage{algpseudocode}
\usepackage{listings}
\usepackage{xspace}
\usepackage{bm}
\usepackage{rotating}
\usepackage{subcaption}
\usepackage{nicematrix}

\theoremstyle{plain}

\theoremstyle{definition}

\theoremstyle{remark}

\definecolor{bestred}{HTML}{F7B6B6}
\definecolor{secondyellow}{HTML}{FFF3A6}
\newcommand{\best}[1]{\cellcolor{bestred}#1}
\newcommand{\second}[1]{\cellcolor{secondyellow}#1}
\newcommand{\besttext}[1]{\textcolor{red}{#1}}
\newcommand{\secondtext}[1]{\textcolor{yellow}{#1}}

\newcommand{\Tref}[1]{Table~\ref{#1}}

\newcommand{\fref}[1]{Fig.~\ref{#1}}
\newcommand{\Fref}[1]{Figure~\ref{#1}}

\newcommand{\Sref}[1]{Section~\ref{#1}}

\newcounter{todos}
\AtEndDocument{\ifnum\value{todos}>0 \PackageWarning{TODOS}{There are \arabic{todos} todos left in this paper! Fix them before submitting the paper!} \fi}

\makeatletter
\DeclareRobustCommand\onedot{\futurelet\@let@token\@onedot}
\def\@onedot{\ifx\@let@token.\else.\null\fi\xspace}
\def\eg{\emph{e.g}\onedot}

\makeatother

\newcommand{\anysplat}{AnySplat~\cite{jiang2025anysplat}\xspace}
\newcommand{\any}{AnySplat~\cite{jiang2025anysplat}\xspace}
\newcommand{\wm}{WorldMirror~\cite{liu2025worldmirror}\xspace}
\newcommand{\strgs}{StreamGS~\cite{li2025streamgs}\xspace}
\newcommand{\strsplat}{StreamSplat~\cite{wu2025streamsplat}\xspace}

\newcommand{\flare}{FLARE~\cite{zhang2025flare}\xspace}
\newcommand{\vggt}{VGGT~\cite{wang2025vggt}\xspace}
\newcommand{\svggt}{StreamVGGT~\cite{zhuo2025streaming}\xspace}
\newcommand{\onthefly}{OnTheFly-NVS~\cite{meuleman2025onthefly}\xspace}
\newcommand{\ours}{OF$^3$GS\xspace}
\newcommand{\sgs}{SVGGT+GS\xspace}
\newcommand{\dl}{DL3DV~\cite{ling2024dl3dv}\xspace}
\newcommand{\dlbench}{DL3DV-140~\cite{ling2024dl3dv}\xspace}
\newcommand{\re}{RE10K~\cite{zhou2018stereo}\xspace}
\newcommand{\nyu}{NYUv2~\cite{silberman2012nyuv2}\xspace}

\title{\ours: \underline{O}n-the-\underline{F}ly \underline{F}eed-\underline{F}orward 3D \underline{G}aussian \underline{S}platting from Unposed Images}
\author{Ruiyang Chen\textsuperscript{1},
Feiran Li\textsuperscript{2},
Chu Zhou\textsuperscript{3},
Zonglin Li\textsuperscript{1},
Zhanyu Ma\textsuperscript{1},
and Heng~Guo\textsuperscript{1}\corresponding}
\affiliations{
	\textsuperscript{1}Beijing University of Posts and Telecommunications, China\\
	\textsuperscript{2}Independent Researcher\\
	\textsuperscript{3}National Institute of Informatics, Japan\\
	\texttt{\{chenruiyang,zonglinli,mazhanyu,guoheng\}@bupt.edu.cn},\\
	\texttt{frl1994@hotmail.com},\\
	\texttt{zhou\_chu@hotmail.com}}

\begin{document}

\maketitle

\begin{abstract}
Feed-forward 3D Gaussian Splatting~(3DGS) enables efficient and high-fidelity novel view synthesis~(NVS) from offline image sequences. However, achieving on-the-fly NVS from unposed images remains challenging: the system must reconstruct renderable 3D Gaussians as images arrive, without access to future observations. Although online feed-forward geometry methods have been developed for causal depth and point-cloud recovery, directly adapting them to NVS often leads to severe rendering artifacts because Gaussian-based rendering demands stricter multi-view consistency in primitive scale and pose-geometry alignment. Even minor deviations can accumulate under causal inference and visibly degrade rendering quality. To this end, we propose \textit{\ours}, a feed-forward framework for efficient and high-quality on-the-fly NVS from sparse-view unposed images under causal constraints. We introduce two mechanisms for causal geometric stability: a \textit{Decoupled Intrinsic Recovery Head} that mitigates cumulative camera-intrinsic bias and scene-scale jitter, and \textit{Dynamic Point Refinement Offsets} that relax rigid unprojection to compensate for coupled pose-depth drift. Extensive experiments show that \ours outperforms online baselines and approaches offline feed-forward 3DGS methods under comparable sparse-input settings. It also remains memory-feasible with denser inputs. Homepage: \url{https://richardchen225.github.io/of3gs/}

\end{abstract}

\section{Introduction}

3D Gaussian Splatting (3DGS) has evolved from per-scene optimization to feed-forward networks that regress Gaussian parameters from offline image sequences. Moving beyond this offline paradigm, a more challenging yet practical setting is on-the-fly 3DGS reconstruction from unposed images, where a feed-forward system incrementally constructs a 3DGS representation without access to future frames while maintaining efficiency and novel view synthesis~(NVS) quality. This setting naturally involves two largely separable challenges: \textit{sparse-view causal reconstruction}, where future views are unavailable and past predictions cannot be revised, and \textit{dense-view scaling}, where redundant observations increase the active state size and the lifetime of accumulated errors. Such capability is crucial for latency-sensitive applications, including on-the-fly video stabilization~\cite{lin2025longsplat,you2025gavs}, on-the-fly sparse-view scene capture~\cite{quesado2026livestre4m,zhan2024ontheflysfm}, and AR/VR interaction~\cite{xie2024physgaussian,jiang2024vr}.
\begin{figure}[!t]
    \centering
    \includegraphics[width=\linewidth]{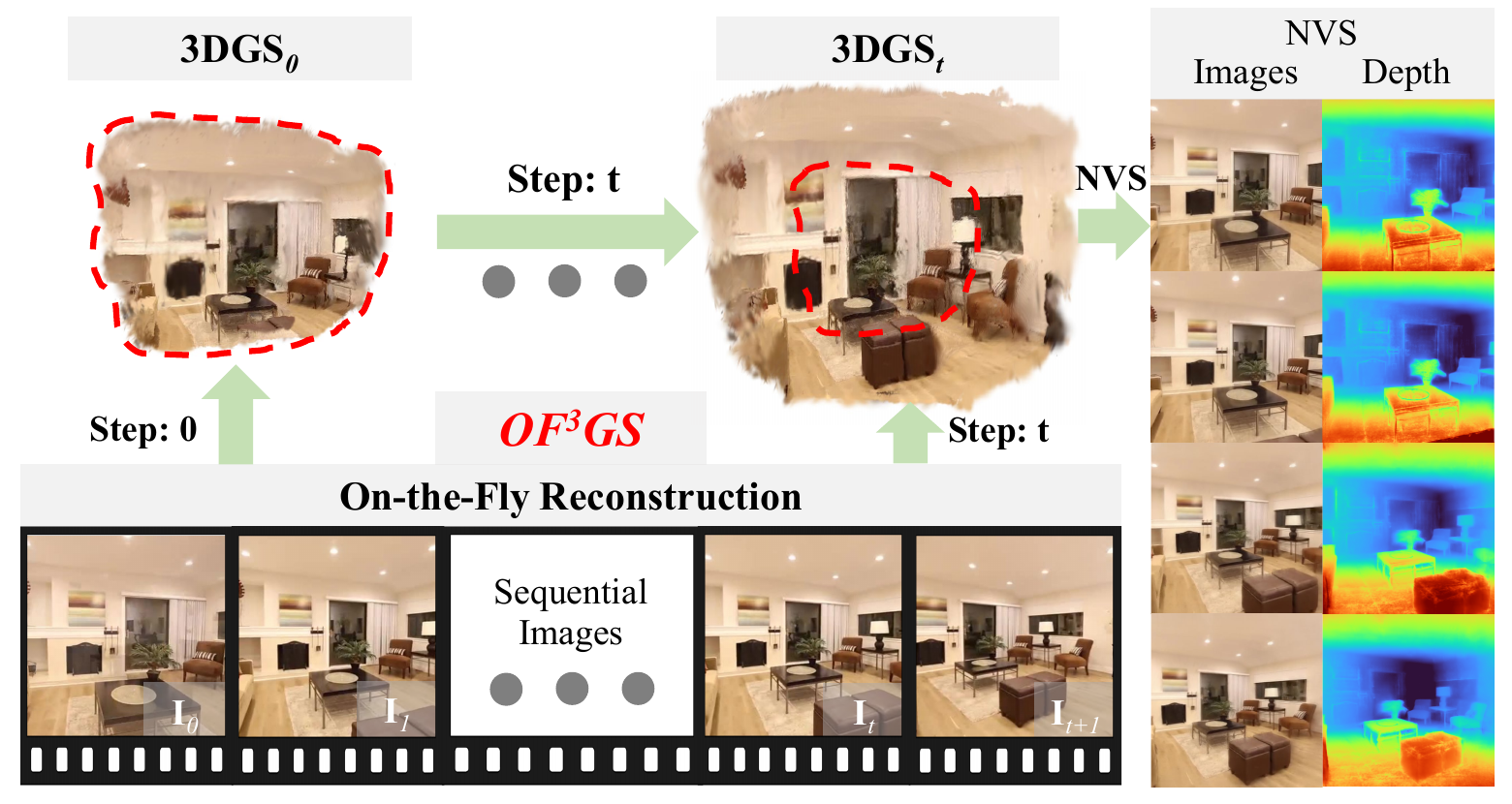}
    \caption{\ours is a feed-forward framework that incrementally reconstructs high-quality 3D Gaussians from incoming unposed images, enabling low-latency on-the-fly novel view synthesis under causal constraints.}
    \label{fig:fig_teaser}
    \vspace{-10pt}
\end{figure}

To build high-quality online 3DGS under sparse causal inputs, optimization-based frameworks~\cite{meuleman2025onthefly, li2025artdeco, lin2025longsplat} perform per-frame iterative refinement to maintain multi-view consistency and rendering fidelity. However, these approaches require iterative updates, which inevitably incur substantial computational overhead, leading to prohibitive latency for applications that require immediate feedback.

To improve efficiency, feed-forward approaches offer a natural alternative. Recent methods enable real-time online geometry recovery~\cite{zhuo2025streaming, stream3r2025} or focus on offline feed-forward 3DGS reconstruction~\cite{jiang2025anysplat,liu2025worldmirror}. However, these approaches do not directly extend to on-the-fly NVS. Unlike depth or point cloud estimation, 3DGS-based NVS is highly sensitive to cross-view misalignment; small pose or scale errors can accumulate over time and progressively degrade rendering quality. Consequently, achieving both high-quality rendering and low-latency 3DGS reconstruction from incoming unposed images remains unresolved.

\begin{figure}[!t]
    \centering
    \includegraphics[width=\linewidth]{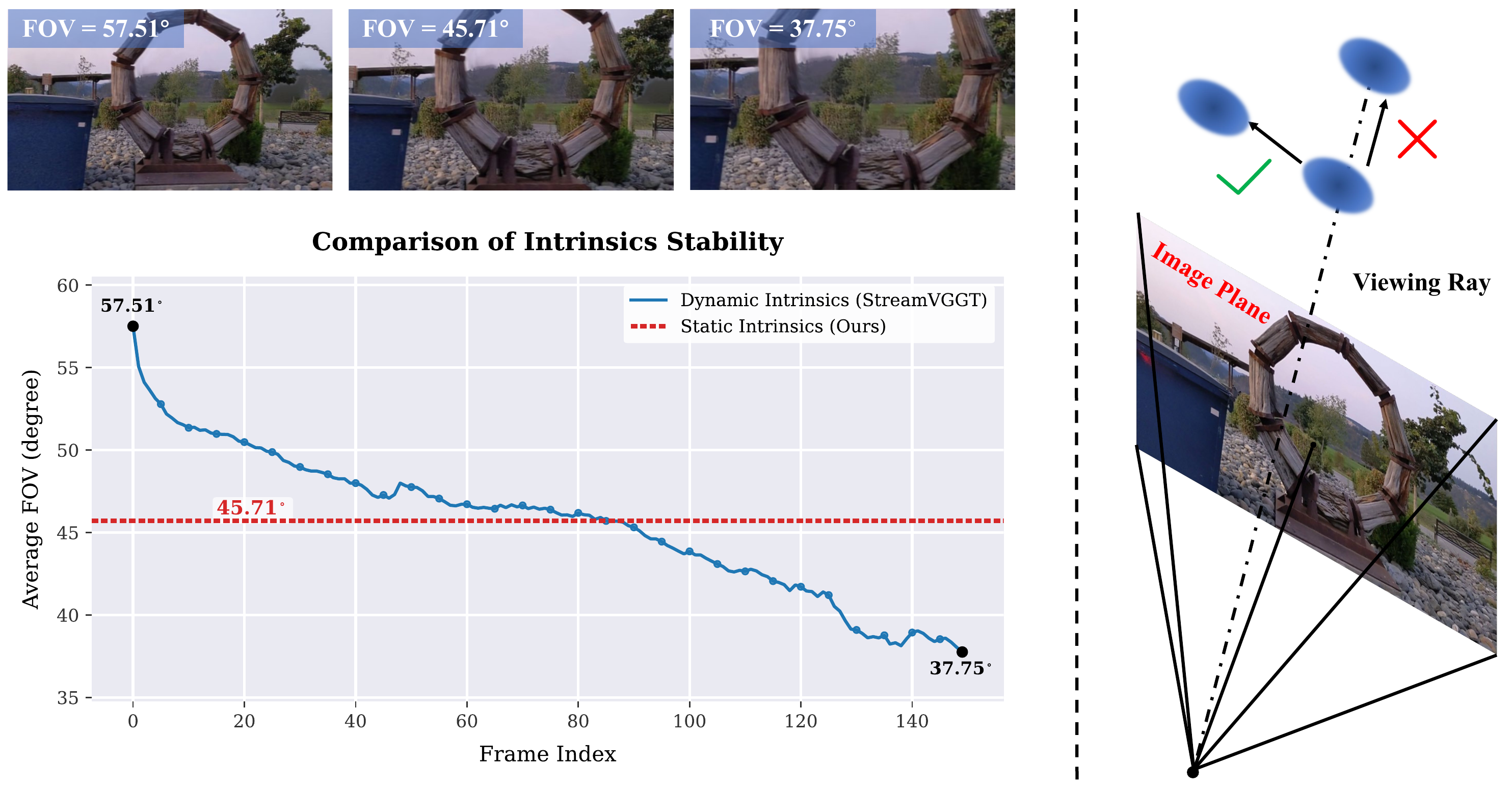}
    \caption{Two geometric inconsistencies in online 3DGS. (Left) Intrinsics drift induces global scale inconsistency. (Right) Rigid viewing-ray constraints lead to distorted Gaussian placement under coupled errors.}
    \label{fig:fig_insight}
    \vspace{-10pt}
\end{figure}
To address this problem, we propose \textit{\ours}, a feed-forward 3DGS framework that achieves high-quality on-the-fly NVS directly from incoming unposed images (\fref{fig:fig_teaser}). We observe that lifting offline feed-forward 3DGS to online fundamentally weakens the implicit regularization of multi-view geometric consistency, as future observations are unavailable during inference. As a result, both geometric reconstruction and camera estimation become progressively unstable, degrading NVS quality. Specifically, (1) frame-wise intrinsic prediction drifts, introducing global scale inconsistency and cross-frame misalignment of reconstructed Gaussians (left side of \fref{fig:fig_insight}); and (2) per-pixel 3DGS constrains Gaussian updates along viewing rays, making the reconstruction sensitive to coupled pose-depth errors that accumulate and distort Gaussian placement over time (right side of \fref{fig:fig_insight}). These failures already arise in short input sequences, making sparse-view causal reconstruction quality a prerequisite for dense-view scaling. Therefore, this paper focuses on sparse-view causal reconstruction as the primary task, while treating dense-view reconstruction as a validation of the framework's resilience against catastrophic geometric collapse and its ability to mitigate error accumulation under denser input sequences.

To address these challenges, \ours incorporates two synergistic mechanisms designed to restore geometric stability under causal inference. First, we propose \textit{Decoupled Intrinsic Recovery Head (DIR-Head)}. With the observation that many capture settings use fixed camera intrinsics and by anchoring intrinsics across time, DIR-Head enables multiple frames to share a unified estimate, mitigating cumulative scale drift and enforcing global scale consistency across per-frame 3DGS predictions. Second, to address inaccuracies in per-pixel 3DGS caused by pose-depth drift, we further introduce \textit{Dynamic Point Refinement Offsets (DPR-Offsets)}, which refine Gaussian positions by compensating for coupled pose-depth errors and relaxing rigid pixel-to-3DGS unprojection constraints.

In summary, our main contributions include:
\begin{enumerate}
    \item \textit{\ours}, a feed-forward framework for high-quality on-the-fly 3DGS reconstruction from sparse-view unposed images under causal constraints;
    \item \textit{DIR-Head}, motivated by the observation that frame-wise intrinsic estimation often suffers from drastic drifts and recursive bias; it stabilizes intrinsic estimation to mitigate global scale inconsistency;
    \item \textit{DPR-Offsets}, a geometric compensator that provides the necessary spatial relaxation to break free from rigid viewing-direction constraints, thereby holistically rectifying the coupled drift of camera poses and depth to ensure precise alignment of 3D Gaussian primitives.
\end{enumerate}
Experiments show that \ours improves sparse-view causal NVS quality over online baselines and approaches offline feed-forward methods under comparable sparse-input settings, with additional dense-view evaluations validating memory-feasible inference.

\section{Related Work}

\noindent\textbf{Feed-forward-based Offline 3DGS} 
leverages neural networks to directly regress 3D Gaussian primitives, bypassing the time-consuming per-scene optimization of the original 3DGS. Early methods~\cite{xu2025depthsplat, wang2024freesplat, zheng2024gps, charatan2024pixelsplat, chen2024mvsplat} focused on sparse-view generalizable reconstruction with known poses, while subsequent works~\cite{ye2024no, hong2024pf3plat} extended this paradigm to unposed inputs through canonical anchoring or pose-aware representations. More recent universal architectures~\cite{jiang2025anysplat, li2025vicasplat, liu2025worldmirror, depthanything3} further exploit powerful transformer backbones to jointly infer cameras and 3D scenes from unconstrained images. However, despite their impressive visual quality, these methods rely on non-causal processing of the entire image sequence to extract global context, rendering them ill-suited for on-the-fly causal NVS.

\noindent\textbf{Optimization-based Online 3DGS} methods achieve high-fidelity causal reconstruction by jointly performing pose tracking and iterative optimization of 3D Gaussian primitives. Representative works~\cite{meuleman2025onthefly, huang2025longsplatonlinegeneralizable3d, xu2025gaussian, li2025artdeco}, such as \onthefly, obtain strong visual quality but depend on costly per-frame optimization, limiting their applicability in low-latency scenarios requiring immediate feedback. In contrast, \ours bypasses this overhead with feed-forward reconstruction under on-the-fly causal constraints, avoiding iterative refinement at inference time.

\noindent\textbf{Feed-forward-based Online 3DGS} avoid iterative optimization by performing incremental reconstruction in a single forward pass. Representative methods~\cite{li2025streamgs, xia2026onlinex, wu2025streamsplat, guo2025salon3r} such as \strgs and OnlineX~\cite{xia2026onlinex} establish correspondences using adjacent frame pairs to reduce redundancy. However, this pairwise paradigm provides only a limited temporal receptive field, preventing effective long-term context aggregation and leading to geometric drift over time.
To address this, we leverage full-history causal attention to expand the temporal receptive field and introduce a decoupled intrinsic head with dynamic offsets to rectify geometric instabilities.


\begin{figure*}[t]
    \centering
    \includegraphics[width=\textwidth]{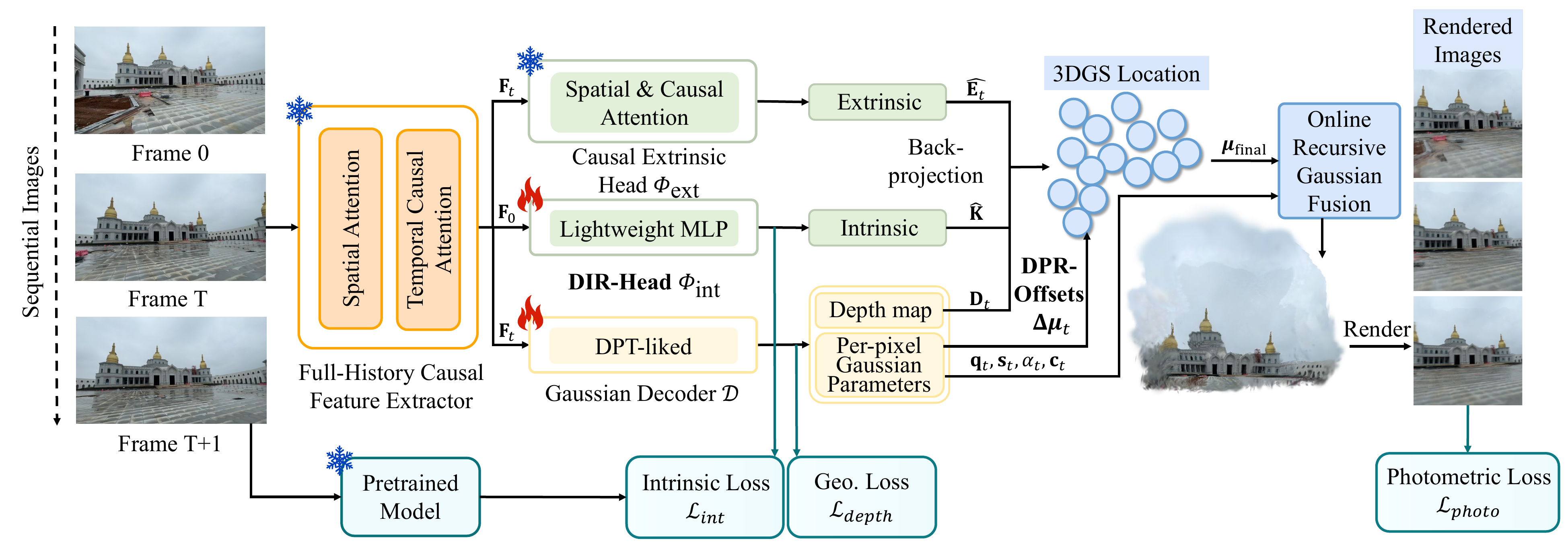}
    \caption{Overview of \ours. A causal extractor with KV caching leverages historical context from unposed images. Decoupled camera recovery heads then predict camera parameters (Causal Extrinsic Head \& DIR-Head) and 3D Gaussian primitives (Gaussian decoder) refined by DPR-Offsets to rectify spatial drift. These primitives are consolidated via Online Recursive Gaussian Fusion for rendering. Training utilizes pretrained model distillation and losses to ensure geometric stability.}
    \vspace{-15pt}
    \label{fig:fig_pipe}
\end{figure*}

\section{Background: On-the-Fly 3DGS Reconstruction}
\label{back}
Given incoming unposed images $\mathcal{I} = \{ \mathbf{I}_t \in \mathbb{R}^{H \times W \times 3} \}_{t=0}^{N-1}$, our goal is to incrementally reconstruct the 3D scene and estimate camera parameters at each time step $t$, utilizing only the observed history $\mathcal{I}_{0:t}$.

Formally, we learn an incremental mapping $f_\theta$ that updates the scene representation based on the current image $\mathbf{I}_t$ and a cached historical state $\mathcal{H}_{t-1}$:
\begin{equation}
    (\mathcal{G}_t, \mathcal{P}_t, \mathcal{H}_t) = f_\theta(\mathbf{I}_t, \mathcal{H}_{t-1}),
\end{equation}
where $\mathcal{P}_t = \{ \hat{\mathbf{K}}_t, \hat{\mathbf{R}}_t, \hat{\mathbf{t}}_t \}$ denotes the estimated camera intrinsics and extrinsics. $\mathcal{H}_t$ represents the updated historical state. The term $\mathcal{G}_t$ represents the collection of 3D Gaussian primitives, each characterized by a set of learned attributes $\{ \mathbf{D}_t, \mathbf{q}_t, \mathbf{s}_t, \alpha_t, \mathbf{c}_t\}$. Here, $\mathbf{D}$ is the per-pixel depth map used for geometric lifting, while $\mathbf{q}, \mathbf{s}, \alpha$, and $\mathbf{c}$ represent the rotation (quaternion), scale, opacity, and color, respectively. 

To ensure geometric consistency, for each pixel coordinate $\mathbf{u} = (i, j)$, a corresponding 3D Gaussian primitive is instantiated. The location $\bm{\mu}_{\mathbf{u}}$ of each primitive is explicitly unprojected from its predicted per-pixel depth $\mathbf{D}[\mathbf{u}]$ into the world coordinate system using the estimated camera parameters:
\begin{equation}
    \bm{\mu}_{\mathbf{u}} = \hat{\mathbf{R}}_t^{-1} \left( \mathbf{D}_t[\mathbf{u}] \cdot \hat{\mathbf{K}}_t^{-1} \tilde{\mathbf{u}} - \hat{\mathbf{t}}_t \right),
    \label{eq:unproject}
\end{equation}
where $\tilde{\mathbf{u}} = [i, j, 1]^\top$ is the homogeneous pixel coordinate and $\mathbf{D}[\mathbf{u}]$ is the scalar depth value at that pixel. For NVS, these aggregated primitives are rendered from arbitrary viewpoints using differentiable rasterization.


\section{Proposed Method: \ours}
\ours is a feed-forward framework for high-quality on-the-fly 3DGS reconstruction from unposed images under causal constraints (\fref{fig:fig_pipe}). It decouples frozen pretrained visual priors from trainable causal geometric stabilization: the former provides feature extraction and camera pose recovery, while our design focuses on stabilizing online 3DGS reconstruction under unposed causal inference. Our pipeline is organized into four sequential stages:
(1) \textit{Decoupled Extrinsic and Intrinsic Recovery} aggregates temporal context and leverages the Decoupled Intrinsic Recovery Head (DIR-Head) for stable intrinsic estimation;
(2) \textit{Gaussian Primitive Recovery} generates explicit 3D Gaussian primitives refined by Dynamic Point Refinement Offsets (DPR-Offsets) under geometric supervision;
(3) \textit{Online Recursive Gaussian Fusion} incrementally consolidates incoming primitives into a consistent global representation;
and (4) a tailored \textit{Training Objective} designs a causal-aware loss suite to enforce multi-view consistency and ensure high-quality rendering.

\subsection{Decoupled Extrinsic and Intrinsic Recovery}
\label{sec:dir}
As shown in \fref{fig:fig_pipe}, we first extract a causal feature map $\mathbf{F}_t$ with a full-history backbone based on \svggt. With full-history causal attention and efficient KV caching, the backbone preserves long-term temporal consensus across frames. Based on these features, a causal extrinsic head $\Phi_{\text{ext}}$ queries $\mathbf{F}_t$ and the cached historical state $\mathcal{H}_{t-1}$ to predict the relative camera extrinsic matrix $\hat{\mathbf{E}}_t = [\hat{\mathbf{R}}_t \,|\, \hat{\mathbf{t}}_t]$, enabling smooth trajectory recovery relative to the initial observation.

Second, while camera extrinsics are inherently relative and temporal, intrinsics are physically static properties of the lens. To address the drift inherent in frame-wise intrinsic estimation, we propose the \textit{Decoupled Intrinsic Recovery Head (DIR-Head)}. We enforce a physically grounded constraint by assuming a square pixel aspect ratio and a globally consistent focal length for the entire sequence ($f_x = f_y = f$). As illustrated in \fref{fig:fig_pipe}, the DIR-Head is implemented via a lightweight MLP $\Phi_{\text{int}}$ that operates exclusively on the first frame feature $\mathbf{F}_0$ to anchor the global coordinate system. To ensure this module remains resolution-agnostic, the MLP predicts a normalized focal length ratio, which is scaled by the image width $W$ to obtain the absolute focal length $\hat{f}$ in pixel units:
\begin{equation}
    \hat{f} = \Phi_{\text{int}}(\mathbf{F}_0) \cdot W.
\end{equation}
By predicting a normalized ratio rather than absolute pixel values, DIR-Head achieves better generalization across different image scales and stabilizes the numerical range during training. The final intrinsic matrix $\hat{\mathbf{K}}$ is constructed as:
\begin{equation}
    \hat{\mathbf{K}} = \begin{bmatrix} 
    \hat{f} & 0 & W/2 \\
    0 & \hat{f} & H/2 \\
    0 & 0 & 1 
    \end{bmatrix},
\end{equation}
where $W$ and $H$ denote the image width and height, respectively. Training the DIR-Head from scratch in an unposed setting is often unstable due to the cold-start problem, where early focal errors can disrupt downstream geometric learning. To stabilize this process, we use teacher forcing with a frozen \vggt teacher to provide a stable reference focal length $f_{\text{teacher}}$. During training, we feed this stable $f_{\text{teacher}}$ into the unprojection module and supervise the student with an L2 distillation loss $\mathcal{L}_{\text{int}}$:
\begin{equation}
    \mathcal{L}_{\text{int}} = \| \hat{f} - \text{sg}(f_{\text{teacher}}) \|^2,
\end{equation}
where $\text{sg}(\cdot)$ denotes the stop-gradient operation. At inference time, the teacher is discarded and the trained student DIR-Head alone provides low-latency parameter recovery.

\subsection{Gaussian Primitive Recovery}
\label{sec:dpr}
With the causal features $\mathbf{F}_t$ extracted and camera parameters $\mathcal{P}_t$ recovered, our framework transforms these latent representations into explicit 3D Gaussian primitives, as illustrated in \fref{fig:fig_pipe}. Specifically, we employ a DPT-like~\cite{ranftl2021vision} convolutional Gaussian decoder $\mathcal{D}$ to map $\mathbf{F}_t$ into the per-pixel Gaussian attributes defined in the background section. 

We observe that the standard per-pixel 3DGS representation constrains Gaussian position updates primarily along the viewing direction through depth lifting. In a causal setting, this constraint causes small depth and pose errors to accumulate and distort Gaussian placement, thereby significantly degrading rendering quality. To resolve this, we propose \textit{Dynamic Point Refinement Offsets (DPR-Offsets)}. This module enables the decoder $\mathcal{D}$ to additionally predict a per-pixel 3D residual map $\boldsymbol{\Delta \mu}_t \in \mathbb{R}^{H \times W \times 3}$, which provides the necessary spatial relaxation to rectify joint pose-depth drift. Specifically, for each pixel $\mathbf{u}=(i,j)$, we first compute an initial 3D position $\bm{\mu}_{\mathbf{u}, \text{init}}$ through unprojection using the predicted scalar depth $\mathbf{D}_t[\mathbf{u}]$:
\begin{equation}
    \bm{\mu}_{\mathbf{u}, \text{init}} = \hat{\mathbf{R}}_t^{-1} \left( \mathbf{D}_t[\mathbf{u}] \cdot \hat{\mathbf{K}}^{-1} \tilde{\mathbf{u}} - \hat{\mathbf{t}}_t \right).
\end{equation}
The final refined Gaussian center is then obtained by applying the predicted offsets: $\bm{\mu}_{\mathbf{u}, \text{final}} = \bm{\mu}_{\mathbf{u}, \text{init}} + \boldsymbol{\Delta \mu}_t[\mathbf{u}]$. This mechanism effectively detaches primitive placement from rigid viewing-ray constraints, allowing the network to dynamically align new Gaussians with the existing scene structure.

To ensure that this geometric refinement operates on a stable and physically plausible baseline, we incorporate scale-shift invariant geometric supervision to prevent the underlying depth map $\mathbf{D}_t$ from drifting during training. As illustrated in \fref{fig:fig_pipe}, we utilize a pretrained Depth Anything V3~\cite{depthanything3} model to generate pseudo-ground-truth depth priors, denoted as $\mathbf{D}^{\text{prior}}_t$. The predicted depth map from the Gaussian head $\mathcal{D}$ is compared against these priors. To supervise structural consistency while disregarding absolute scale, we solve for the optimal per-frame scale $\gamma_t$ and shift $\beta_t$ via least-squares alignment (detailed in Suppl.) before computing the geometric loss:
\begin{equation}
    \mathcal{L}_{\text{depth}} = \| (\gamma_t \cdot \mathbf{D}_t + \beta_t) - \mathbf{D}^{\text{prior}}_t \|^2.
\end{equation}
By stabilizing the depth baseline through this invariant constraint, the framework can effectively focus on recovering accurate relative scene structures, while the DPR-Offsets handle the fine-grained spatial corrections necessitated by causal drift.

\subsection{Online Recursive Gaussian Fusion}
\label{sec:agg}
Once refined and unprojected, per-frame Gaussian primitives must be consolidated into a compact global representation for efficient, high-quality rendering. Directly accumulating all per-pixel primitives would introduce redundancy and rendering noise. To resolve this, inspired by \anysplat, we aggregate geometrically proximate primitives in a voxelized radiance field using the decoder confidence scores $w_t$. While standard offline methods perform this fusion via confidence-weighted batch averaging across all views, such a strategy violates the strict causality of our online setting.

Specifically, \textit{Online Recursive Gaussian Fusion} maintains a global Gaussian state cache, where each active voxel $v$ stores an aggregated attribute vector $\mathbf{H}_{t-1}^{(v)}$ and an accumulated confidence weight $\Omega_{t-1}^{(v)}$. When frame $t$ arrives, pixels mapped to voxel $v$ form $\mathcal{U}_t^{(v)} = \{ \mathbf{u} \mid \text{proj}(\mathbf{u}, \mathcal{P}_t, \mathbf{D}_t) \in v \}$, and each pixel $\mathbf{u}$ is assigned a decoder confidence score $w_t[\mathbf{u}]$ that weights its contribution. The state is then updated by merging these confidence-weighted attributes with the historical cache:
\begin{align}
    \label{eq:recursive_weight}
    \Omega_t^{(v)} &= \Omega_{t-1}^{(v)} + \sum_{\mathbf{u} \in \mathcal{U}_t^{(v)}} \exp(w_t[\mathbf{u}]), \\
    \label{eq:recursive_attr}
    \mathbf{H}_t^{(v)} &= \frac{\Omega_{t-1}^{(v)} \cdot \mathbf{H}_{t-1}^{(v)} + \sum_{\mathbf{u} \in \mathcal{U}_t^{(v)}} \exp(w_t[\mathbf{u}]) \cdot \mathbf{a}_t[\mathbf{u}]}{\Omega_t^{(v)}},
\end{align}
where $\mathbf{a}_t[\mathbf{u}] = \{ \bm{\mu}_{\mathbf{u}, \text{final}}, \mathbf{q}_t, \mathbf{s}_t, \alpha_t, \mathbf{c}_t \}$ denotes the Gaussian attributes of pixel $\mathbf{u}$, and $\Omega_t^{(v)}$ is the cumulative confidence weight stored in voxel $v$ at time $t$. This recursive update preserves the aggregation behavior of offline confidence-weighted fusion while requiring only the cached voxel state under causal inference.

\begin{table*}[!t]
    \centering
    \resizebox{\linewidth}{!}{
        \begin{tabular}{@{}c l c | ccc | ccc | ccc | ccc@{}}
            \toprule
            \multirow{2}{*}{\textbf{Dataset}} & \multirow{2}{*}{\textbf{Method}} &
            \multirow{2}{*}{Online?} &
            \multicolumn{3}{c|}{Input Views: 5} &
            \multicolumn{3}{c|}{Input Views: 10} &
            \multicolumn{3}{c|}{Input Views: 64} &
            \multicolumn{3}{c}{Input Views: 128} \\
            \cmidrule(lr){4-6} \cmidrule(lr){7-9} \cmidrule(lr){10-12} \cmidrule(l){13-15}
             & & &
             PSNR $\uparrow$ & SSIM $\uparrow$ & LPIPS $\downarrow$ &
             PSNR $\uparrow$ & SSIM $\uparrow$ & LPIPS $\downarrow$ &
             PSNR $\uparrow$ & SSIM $\uparrow$ & LPIPS $\downarrow$ &
             PSNR $\uparrow$ & SSIM $\uparrow$ & LPIPS $\downarrow$ \\
            \midrule
            \multirow{6}{*}{\rotatebox[origin=c]{90}{\textit{DL3DV-140}}}
            & \flare$^\star$ & $\times$ &
            13.984 & 0.580 & 0.420 &
            13.720 & 0.610 & 0.422 &
            \multicolumn{3}{c|}{OOM} &
            \multicolumn{3}{c}{OOM} \\
            & \anysplat$^\star$ & $\times$ &
            19.139 & 0.577 & 0.388 &
            18.268 & 0.563 & 0.422 &
            \multicolumn{3}{c|}{OOM} &
            \multicolumn{3}{c}{OOM} \\
            & \wm$^\star$ & $\times$ &
            \second{21.559} & \second{0.673} & \second{0.285} &
            \best{20.503} & \best{0.685} & \best{0.320} &
            \multicolumn{3}{c|}{OOM} &
            \multicolumn{3}{c}{OOM} \\
            \cmidrule{2-15}
            & \onthefly$^\dag$ & \checkmark &
            21.030 & 0.644 & 0.325 &
            19.470 & 0.612 & 0.393 &
            \best{18.478} & \best{0.610} & \best{0.455} &
            \best{17.259} & \best{0.566} & \best{0.424} \\
            
            & \sgs$^\star$ & \checkmark &
            19.726 & 0.599 & 0.338 &
            17.614 & 0.532 & 0.443 &
            16.132 & 0.467 & 0.535 &
            15.129 & 0.486 & 0.514 \\

            & \textbf{Ours}$^\star$ & \checkmark &
            \best{21.884} & \best{0.688} & \best{0.273} &
            \second{19.952} & \second{0.629} & \second{0.373} &
            \second{17.190} & \second{0.558} & \second{0.485} &
            \second{16.377} & \second{0.530} & \second{0.461} \\
            \midrule
            \midrule
            \multirow{6}{*}{\rotatebox[origin=c]{90}{\textit{RE10K}}}
            & \flare$^\star$ & $\times$ &
            13.485 & 0.262 & 0.670 &
            13.244 & 0.263 & 0.676 &
            \multicolumn{3}{c|}{OOM} &
            \multicolumn{3}{c}{OOM} \\
            & \anysplat$^\star$ & $\times$ &
            23.034 & 0.753 & 0.251 &
            22.189 & 0.752 & 0.288 &
            \multicolumn{3}{c|}{OOM} &
            \multicolumn{3}{c}{OOM} \\
            & \wm$^\star$ & $\times$ &
            \second{25.419} & \second{0.820} & \second{0.203} &
            \best{25.158} & \best{0.828} & \best{0.216} &
            \multicolumn{3}{c|}{OOM} &
            \multicolumn{3}{c}{OOM} \\
            \cmidrule{2-15}
            & \onthefly$^\dag$ & \checkmark &
            20.829 & 0.815 & 0.290 &
            22.881 & 0.759 & 0.329 &
            \second{20.985} & \best{0.767} & \second{0.303} &
            \second{20.071} & \second{0.725} & \second{0.342} \\

            & \sgs$^\star$ & \checkmark &
            23.767 & 0.776 & 0.234 &
            22.427 & 0.742 & 0.283 &
            19.431 & 0.698 & 0.323 &
            19.245 & 0.675 & 0.354 \\

            & \textbf{Ours}$^\star$ & \checkmark &
            \best{25.797} & \best{0.833} & \best{0.200} &
            \second{24.536} & \second{0.801} & \second{0.241} &
            \best{21.562} & \second{0.749} & \best{0.297} &
            \best{20.598} & \best{0.732} & \best{0.322} \\
            \midrule
            \midrule
            \multirow{6}{*}{\rotatebox[origin=c]{90}{\textit{NYUv2}}}
            & \flare$^\star$ & $\times$ &
            14.643 & 0.505 & 0.619 &
            14.598 & 0.504 & 0.622 &
            \multicolumn{3}{c|}{OOM} &
            \multicolumn{3}{c}{OOM} \\
            & \anysplat$^\star$ & $\times$ &
            22.047 & 0.648 & 0.319 &
            22.024 & 0.660 & 0.334 &
            \multicolumn{3}{c|}{OOM} &
            \multicolumn{3}{c}{OOM} \\
            & \wm$^\star$ & $\times$ &
            \second{24.750} & \second{0.713} & \second{0.303} &
            \best{24.212} & \best{0.694} & \best{0.297} &
            \multicolumn{3}{c|}{OOM} &
            \multicolumn{3}{c}{OOM} \\
            \cmidrule{2-15}
            & \onthefly$^\dag$ & \checkmark &
            20.875 & 0.635 & 0.376 &
            22.515 & 0.629 & 0.323 &
            \best{23.174} & \best{0.679} & \best{0.352} &
            \best{21.454} & \best{0.690} & \best{0.351} \\

            & \sgs$^\star$ & \checkmark &
            22.892 & 0.674 & 0.318 &
            21.413 & 0.633 & 0.331 &
            19.246 & 0.598 & 0.385 &
            18.385 & 0.576 & 0.401 \\

            & \textbf{Ours}$^\star$ & \checkmark &
            \best{24.875} & \best{0.717} & \best{0.291} &
            \second{23.215} & \second{0.684} & \second{0.307} &
            \second{21.253} & \second{0.652} & \second{0.361} &
            \second{20.427} & \second{0.632} & \second{0.382} \\
            \bottomrule
        \end{tabular}
    }
    \caption{Quantitative comparison on \dlbench, \re, and \nyu datasets. $^\star$ and $^\dag$ denote feed-forward and optimization-based methods, respectively. OOM indicates out-of-memory failure. The best and second-best results are highlighted in \besttext{red} and \secondtext{yellow}, respectively.}
    \label{tab:tab_nvs_main}
    \vspace{-10pt}
\end{table*}

\subsection{Training Objective}
\label{sec:loss}
To ensure high-fidelity scene reconstruction and accurate parameter recovery, the trainable components of our framework are optimized jointly using a causal-aware multi-task objective, as denoted by \fref{fig:fig_pipe}. The final training objective $\mathcal{L}_{\text{total}}$ is a weighted sum of the DIR-Head intrinsic distillation loss $\mathcal{L}_{\text{int}}$, the scale-invariant geometric loss $\mathcal{L}_{\text{depth}}$ for DPR-Offsets, and a specialized generalization-weighted photometric loss $\mathcal{L}_{\text{photo}}$:
\begin{equation}
    \mathcal{L}_{\text{total}} = \mathcal{L}_{\text{photo}} + \lambda_{\text{depth}} \mathcal{L}_{\text{depth}} + \lambda_{\text{int}} \mathcal{L}_{\text{int}}.
\end{equation}
For the photometric component, we compute a reconstruction loss term $\mathcal{L}_{\text{rec}}$ that combines MSE, SSIM, and LPIPS metrics: $\mathcal{L}_{\text{rec}}(\hat{\mathbf{I}}, \mathbf{I}) = \mathcal{L}_{\text{MSE}} + \lambda_1 \mathcal{L}_{\text{SSIM}} + \lambda_2 \mathcal{L}_{\text{LPIPS}}$, where $\hat{\mathbf{I}}$ and $\mathbf{I}$ denote the rendered image and the corresponding ground truth. Crucially, we observe that standard feed-forward models tend to overfit to the input views, behaving as latent auto-encoders that fail to capture the true underlying 3D structure. To counteract this and enforce multi-view geometric consistency, we propose Novel-View-Weighted Supervision (NV-Sup), which explicitly assigns a higher weight $\lambda_{\text{novel}}$ to unseen novel views:
\begin{equation}
    \mathcal{L}_{\text{photo}} = \sum_{\mathbf{I} \in \mathcal{V}_{\text{ctx}}} \mathcal{L}_{\text{rec}}(\hat{\mathbf{I}}, \mathbf{I}) + \lambda_{\text{novel}} \sum_{\mathbf{I}' \in \mathcal{V}_{\text{novel}}} \mathcal{L}_{\text{rec}}(\hat{\mathbf{I}}', \mathbf{I}').
\end{equation}
Here, $\mathcal{V}_{\text{ctx}}$ and $\mathcal{V}_{\text{novel}}$ denote the sets of input views and unseen target novel views, respectively. This asymmetric weighting strategy forces the network to rely on the aggregated multi-view geometry rather than single-view feature encoding, significantly improving the rendering quality.

\section{Experiments}
\subsection{Experimental Setup}

\noindent\textbf{Baselines.}
We compare \ours with online optimization-based 3DGS (\onthefly) and SOTA offline feed-forward 3DGS (\flare, \anysplat, \wm), covering both causal quality-efficiency and offline upper-bound references.\footnote{We do not compare with OnlineX~\cite{xia2026onlinex} or \strgs because their official implementations are unavailable, and we exclude \strsplat because it relies on future endpoint frames and breaks our strict-causal online protocol.} Since no public feed-forward online 3DGS baseline is available for direct comparison, we additionally construct \sgs by combining \svggt with \anysplat-style Gaussian prediction and fusion, providing a reproducible feed-forward online reference. \sgs is trained under the same protocol as \ours, making it a matched reference for evaluating our proposed modules. For offline feed-forward baselines, we use their official all-input protocols without manual batching, tiling, chunking, or recursive state updates; OOM in dense-view settings is reported under this protocol. Baseline protocols are detailed in the supplementary material.

\noindent\textbf{Datasets and Evaluation Metrics.} 
\ours is trained on $10,000$ \dl sequences and $10,000$ randomly sampled \re trajectories. We evaluate on \dlbench, $200$ \re test scenes, and $100$ unseen \nyu scenes using PSNR, SSIM, and LPIPS.

\noindent\textbf{Implementation Details.}
We implement \ours in PyTorch and jointly train the proposed online 3DGS modules for 5 epochs on four NVIDIA RTX 5880 Ada GPUs. The full-history causal feature extractor and causal extrinsic head are initialized from \svggt and kept frozen, while the proposed online 3DGS modules are optimized for unposed causal reconstruction. More details are provided in the supplementary material.

\subsection{Experimental Results}

\begin{figure*}[t!]
    \makeatletter\setlength{\@fptop}{0pt}\makeatother
    \centering
    \newcommand{\imgsize}{0.14\textwidth} 
    \setlength{\tabcolsep}{1pt} 
    \renewcommand{\arraystretch}{0.5} 
    
    \begin{NiceTabular}{ccccccc}
        \includegraphics[width=\imgsize]{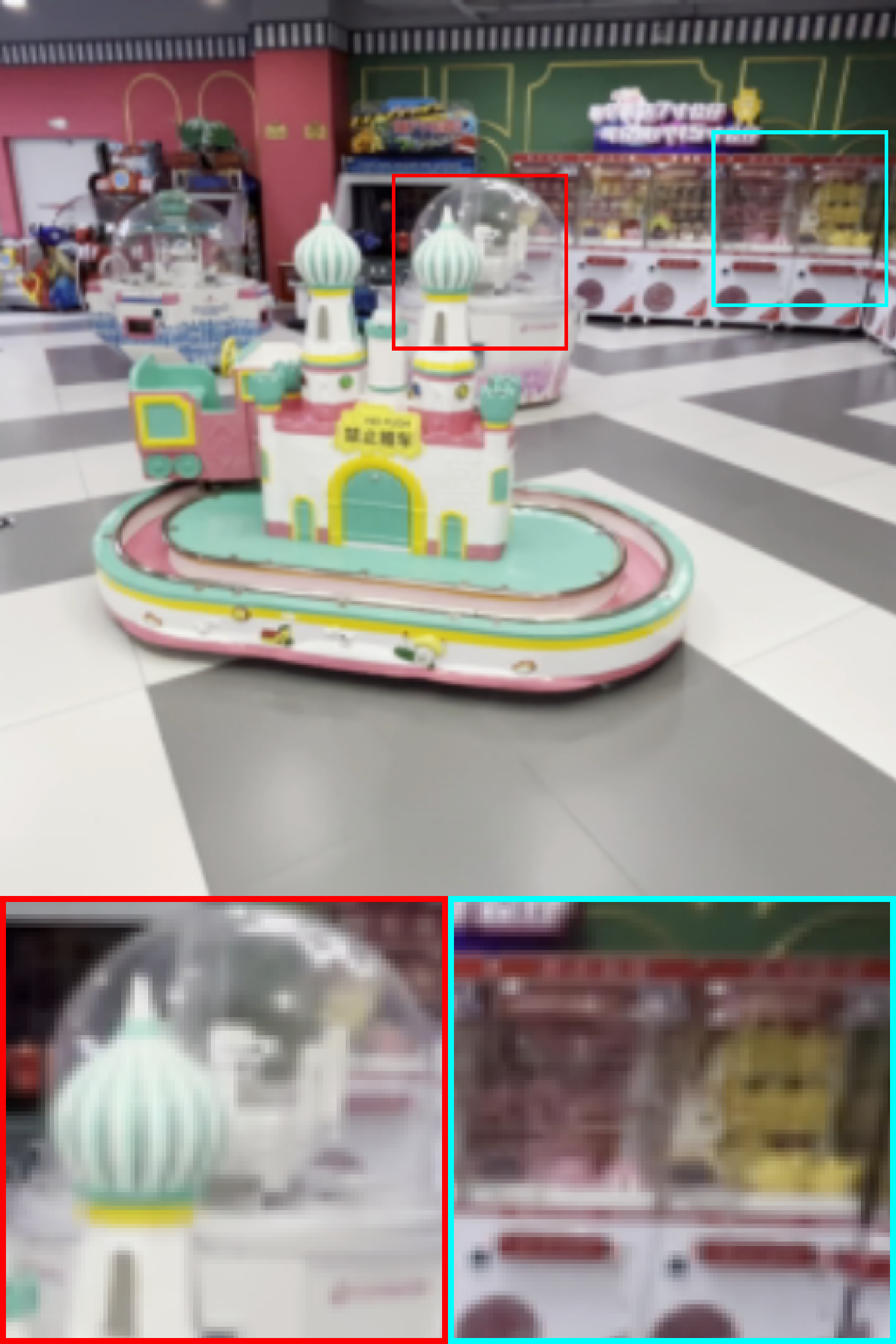} &
        \includegraphics[width=\imgsize]{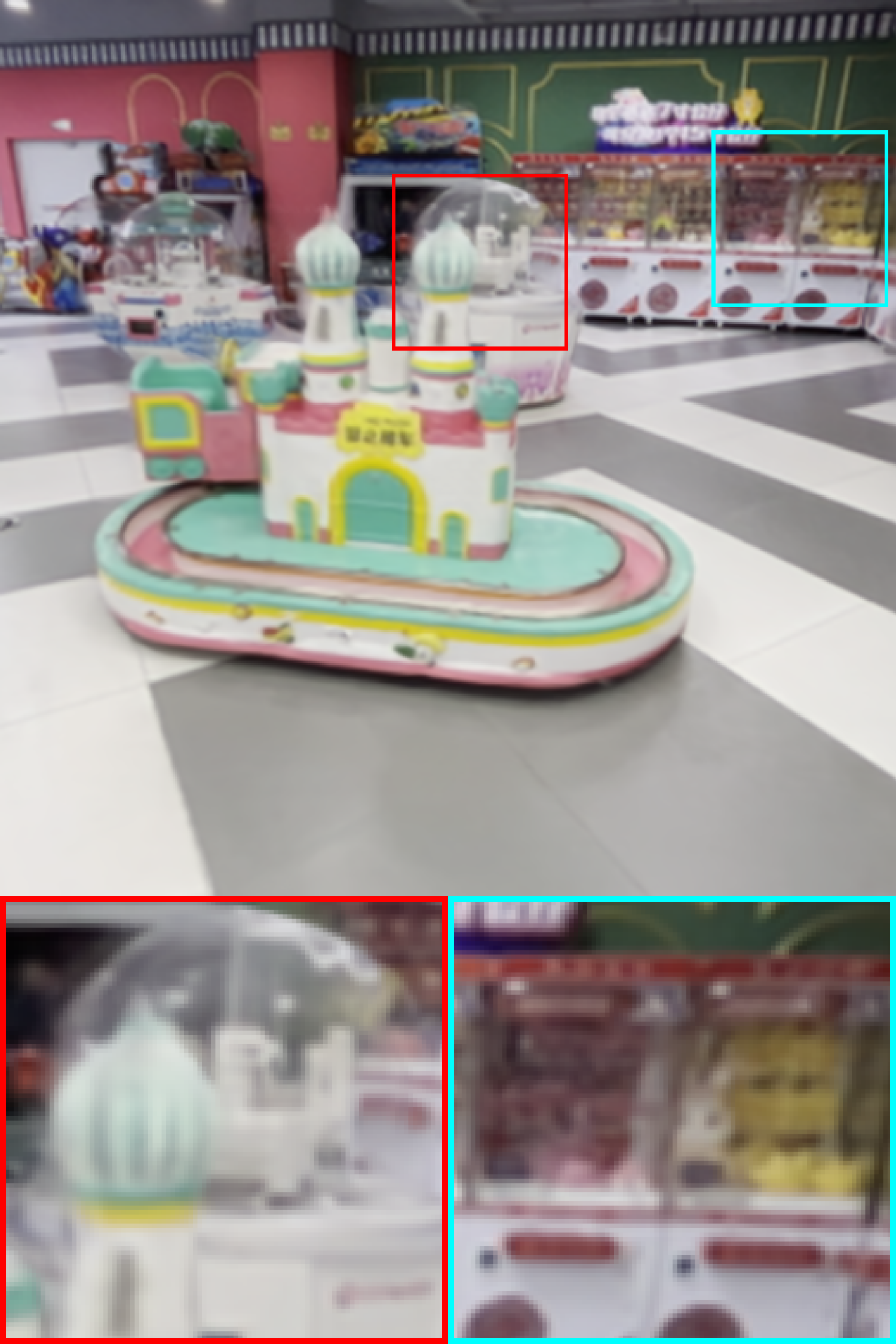} &
        \includegraphics[width=\imgsize]{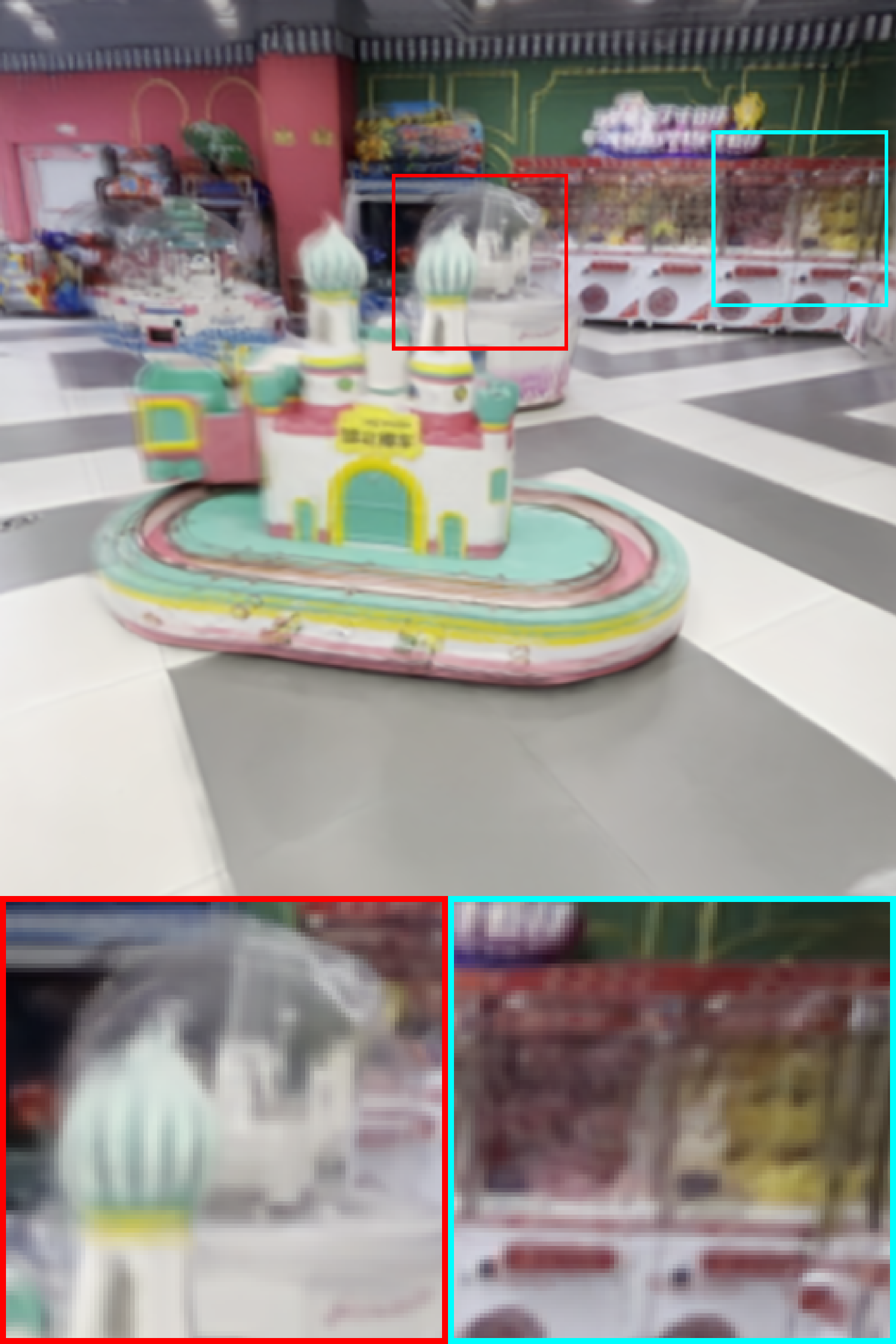} &
        \includegraphics[width=\imgsize]{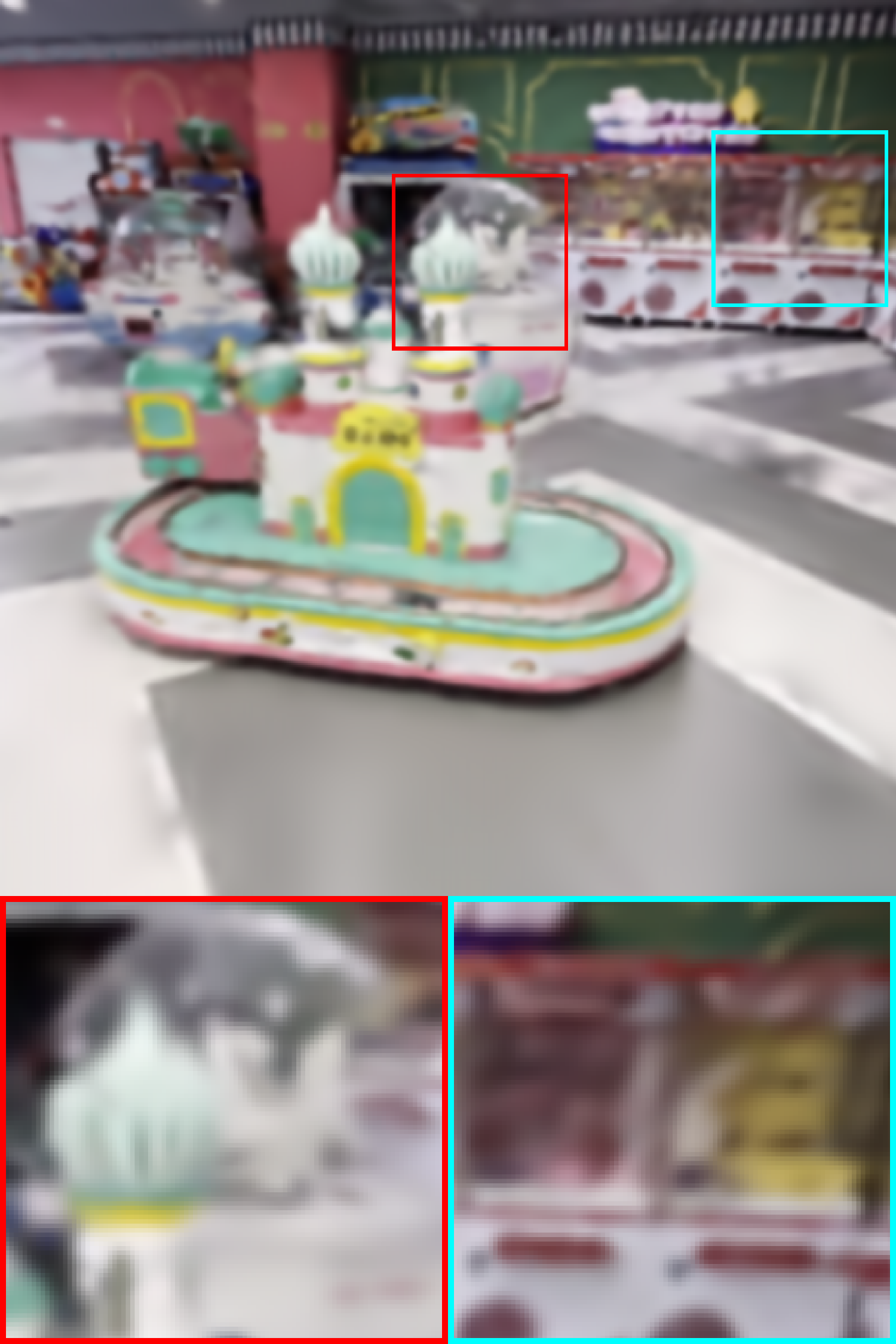} &
        \includegraphics[width=\imgsize]{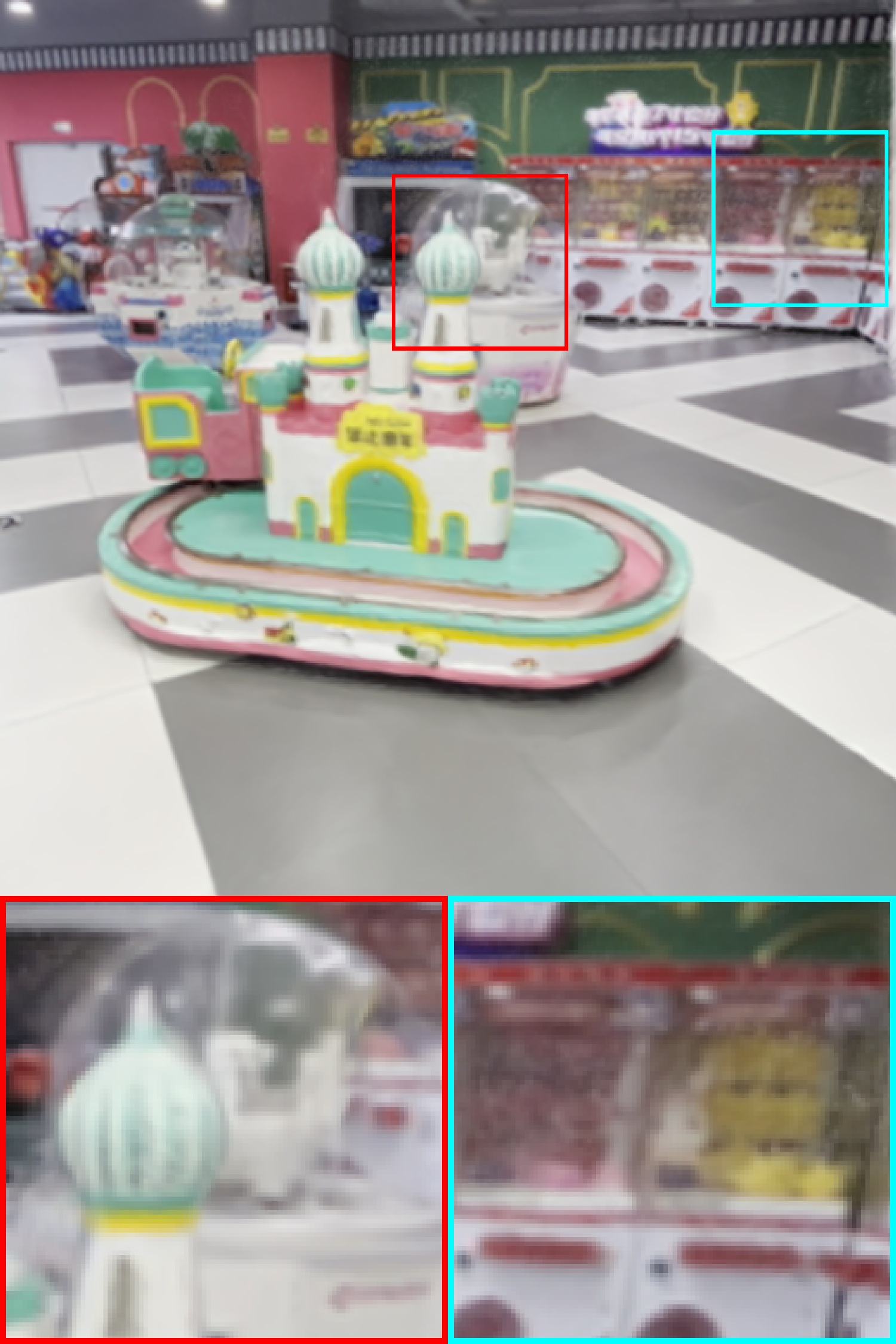} &
        \includegraphics[width=\imgsize]{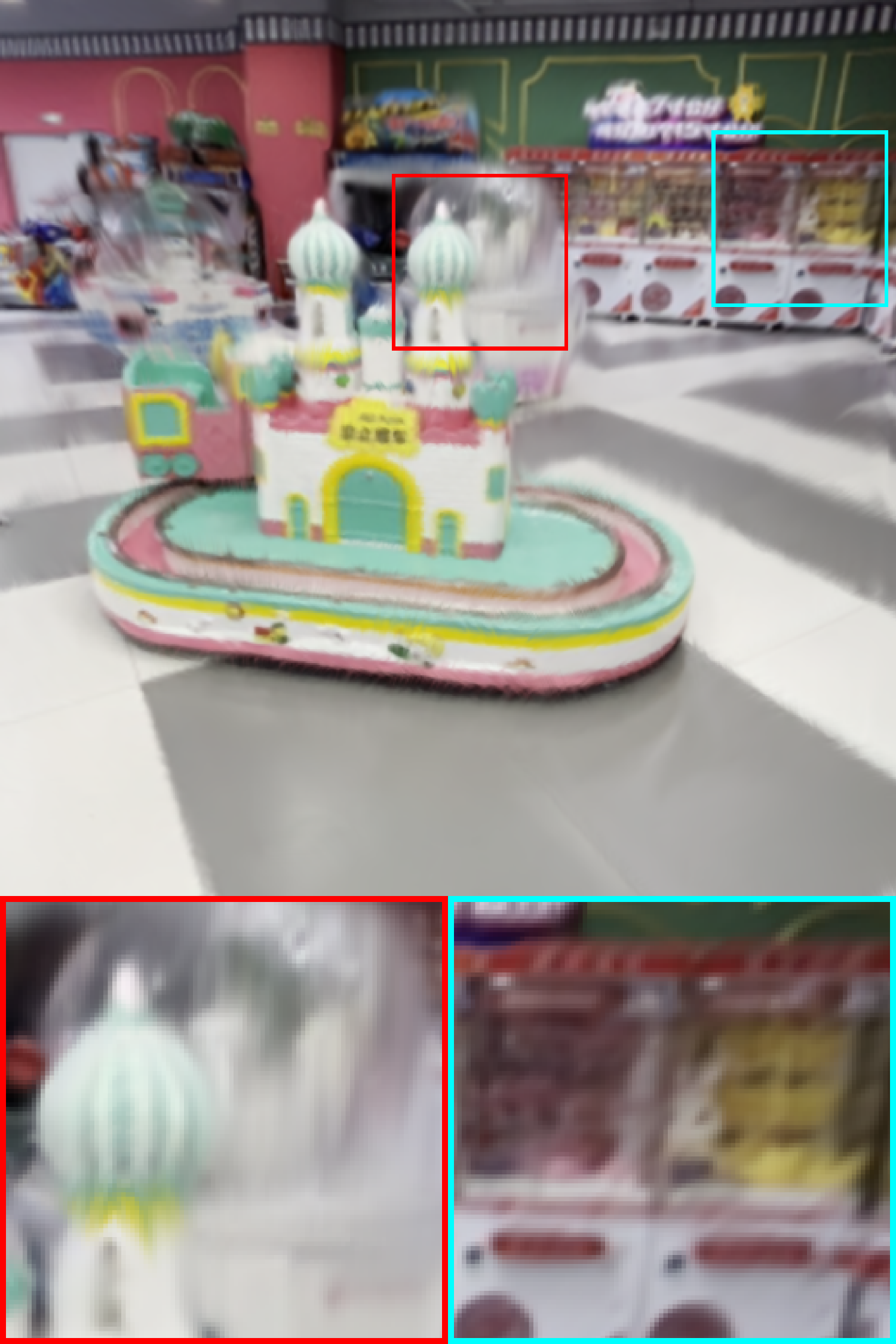} &
        \includegraphics[width=\imgsize]{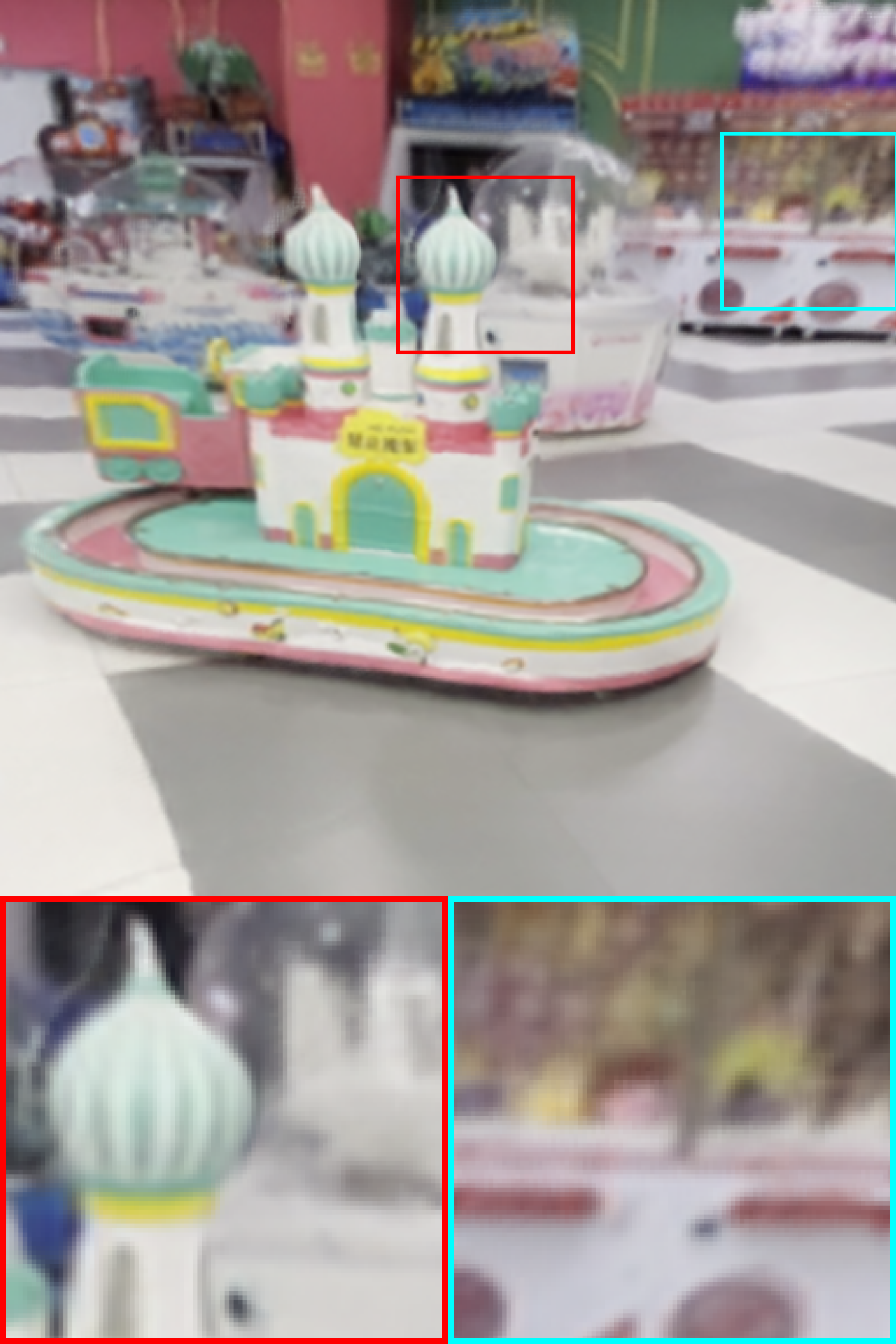} \\
        \includegraphics[width=\imgsize]{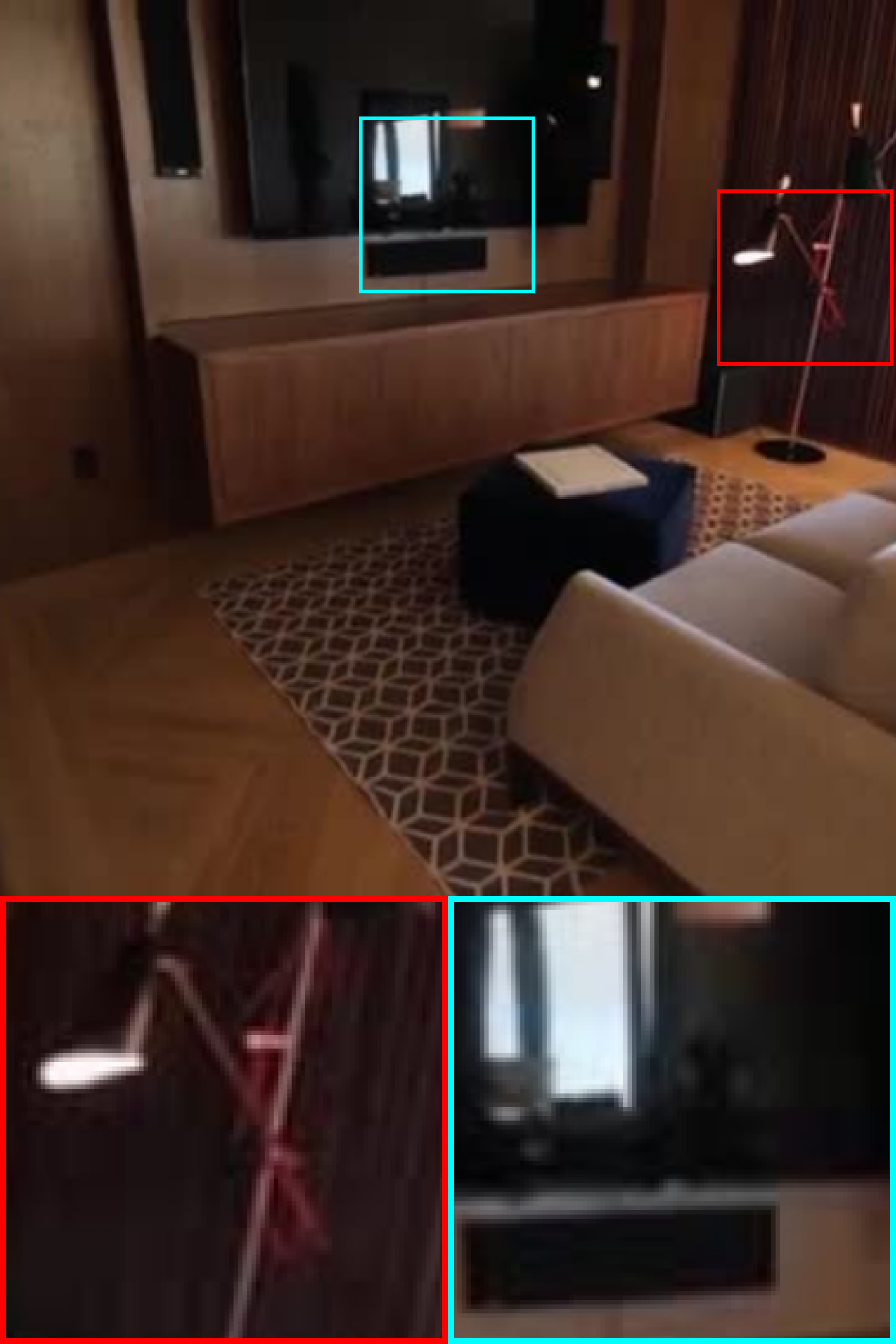} &
        \includegraphics[width=\imgsize]{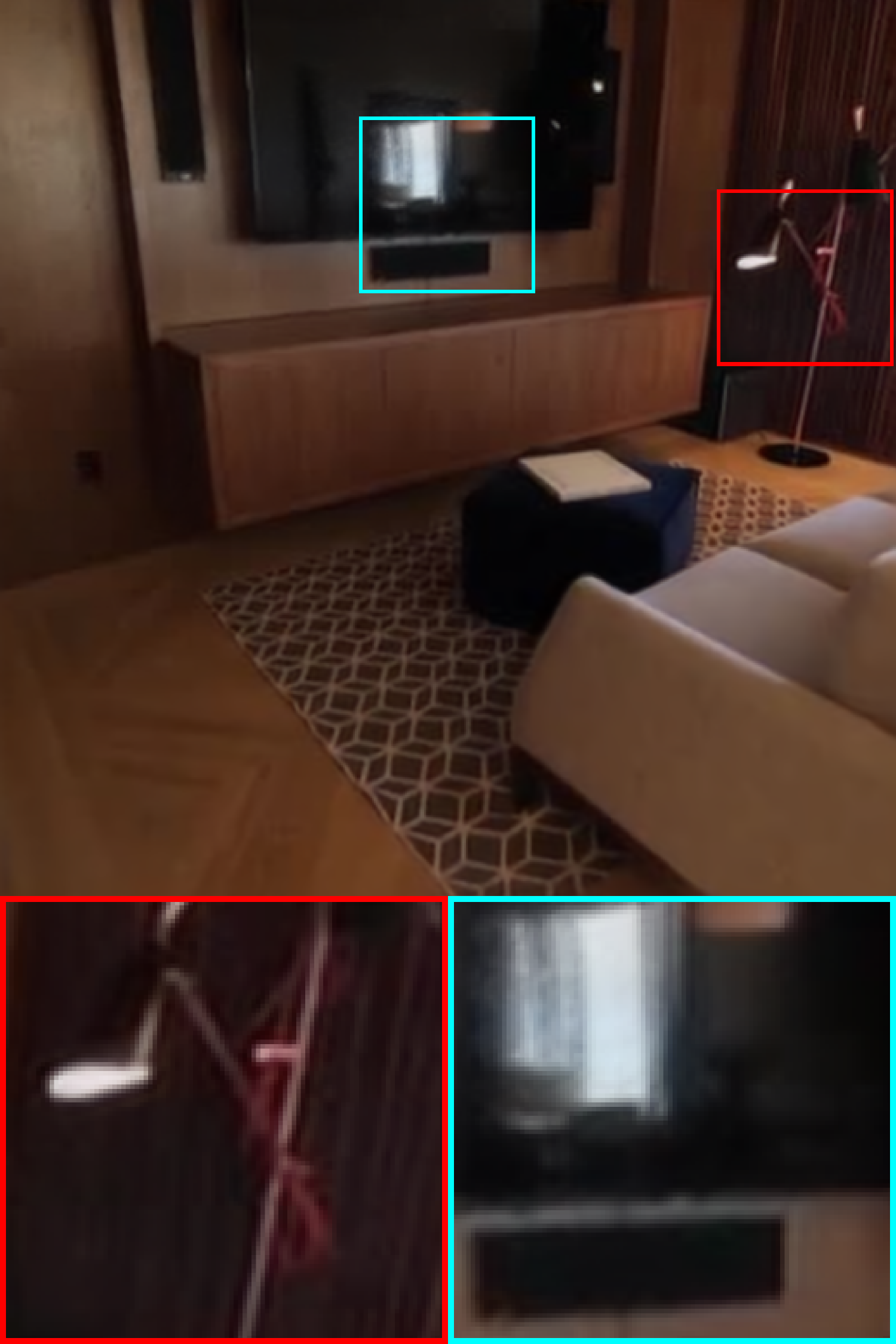} &
        \includegraphics[width=\imgsize]{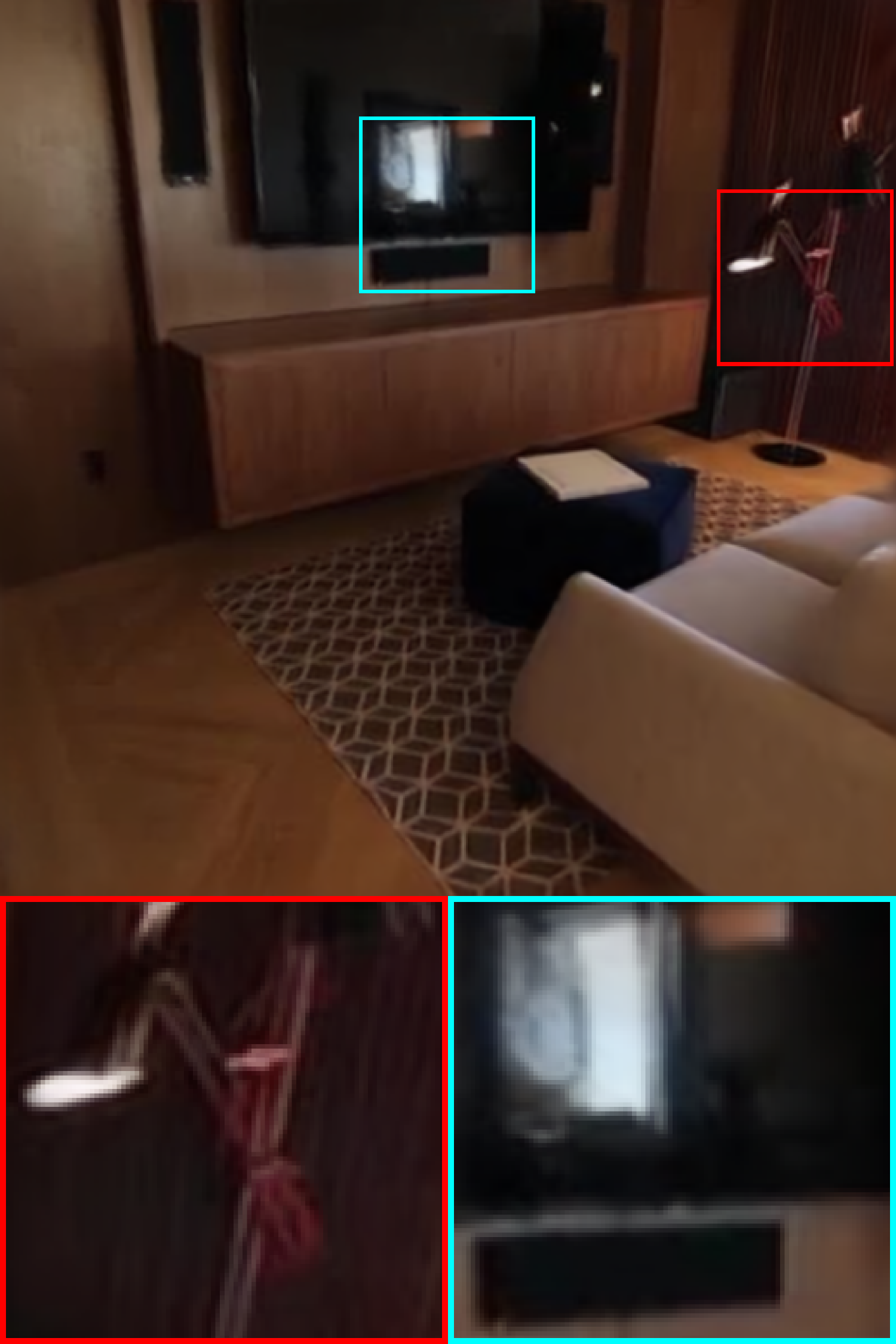} &
        \includegraphics[width=\imgsize]{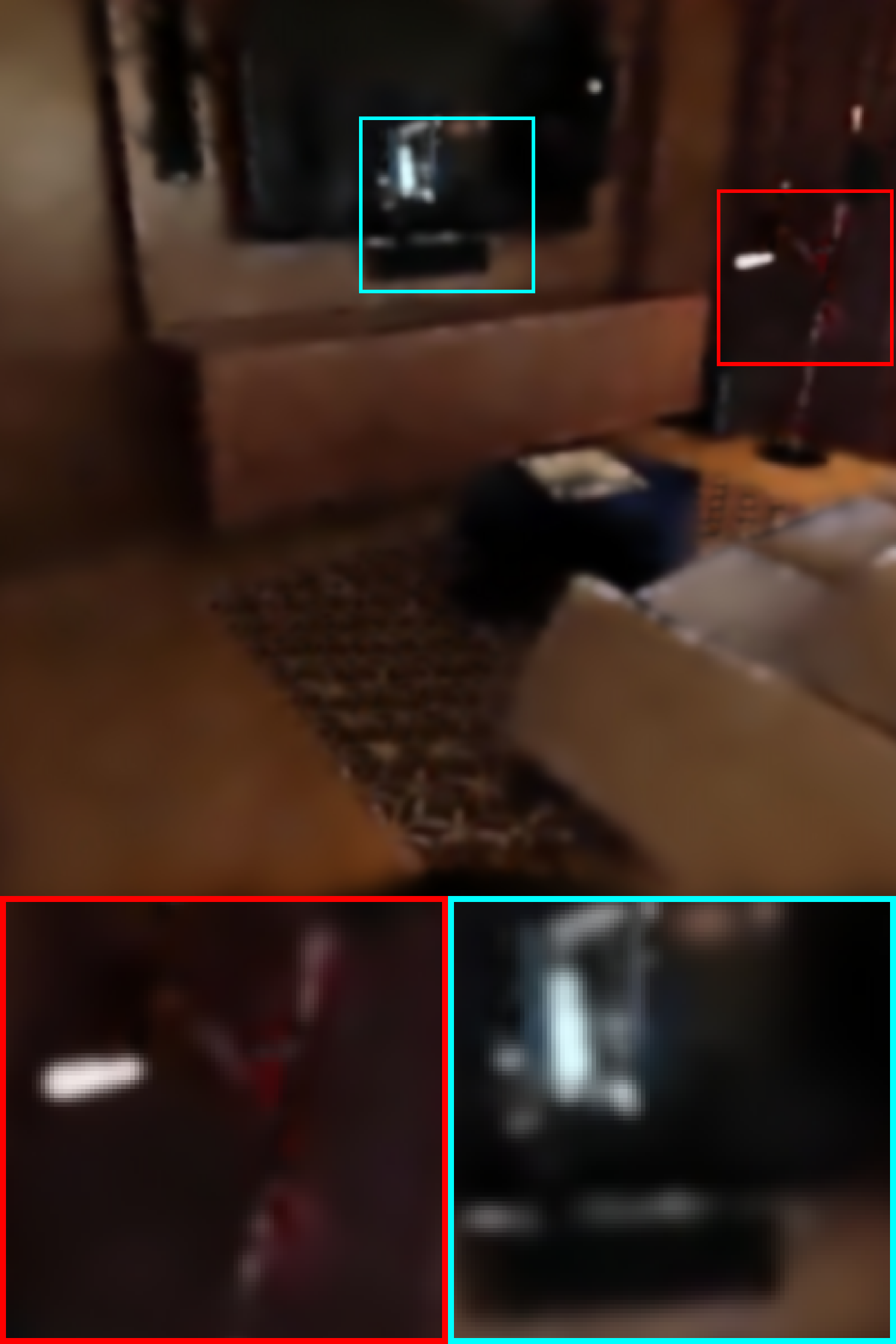} &
        \includegraphics[width=\imgsize]{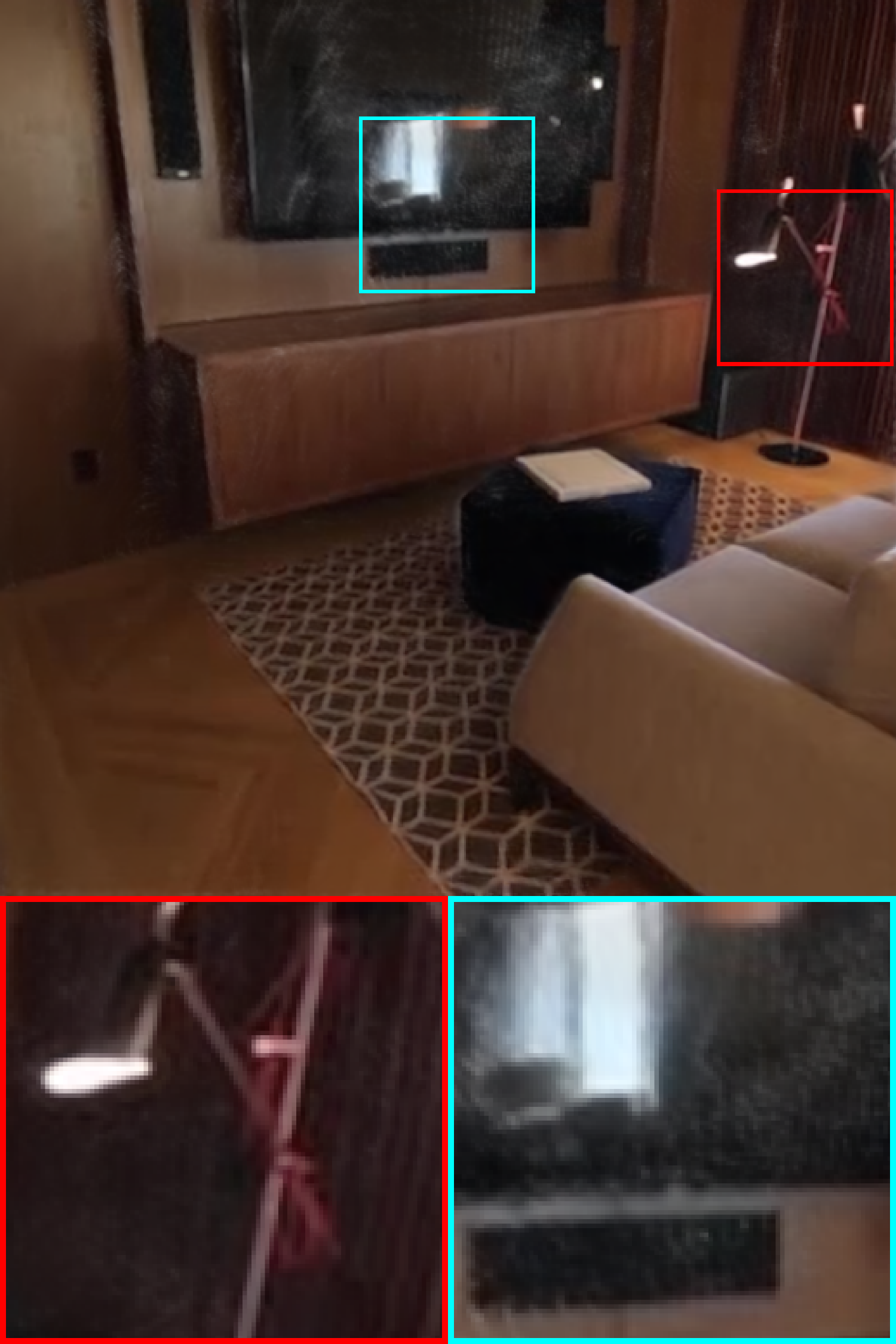} &
        \includegraphics[width=\imgsize]{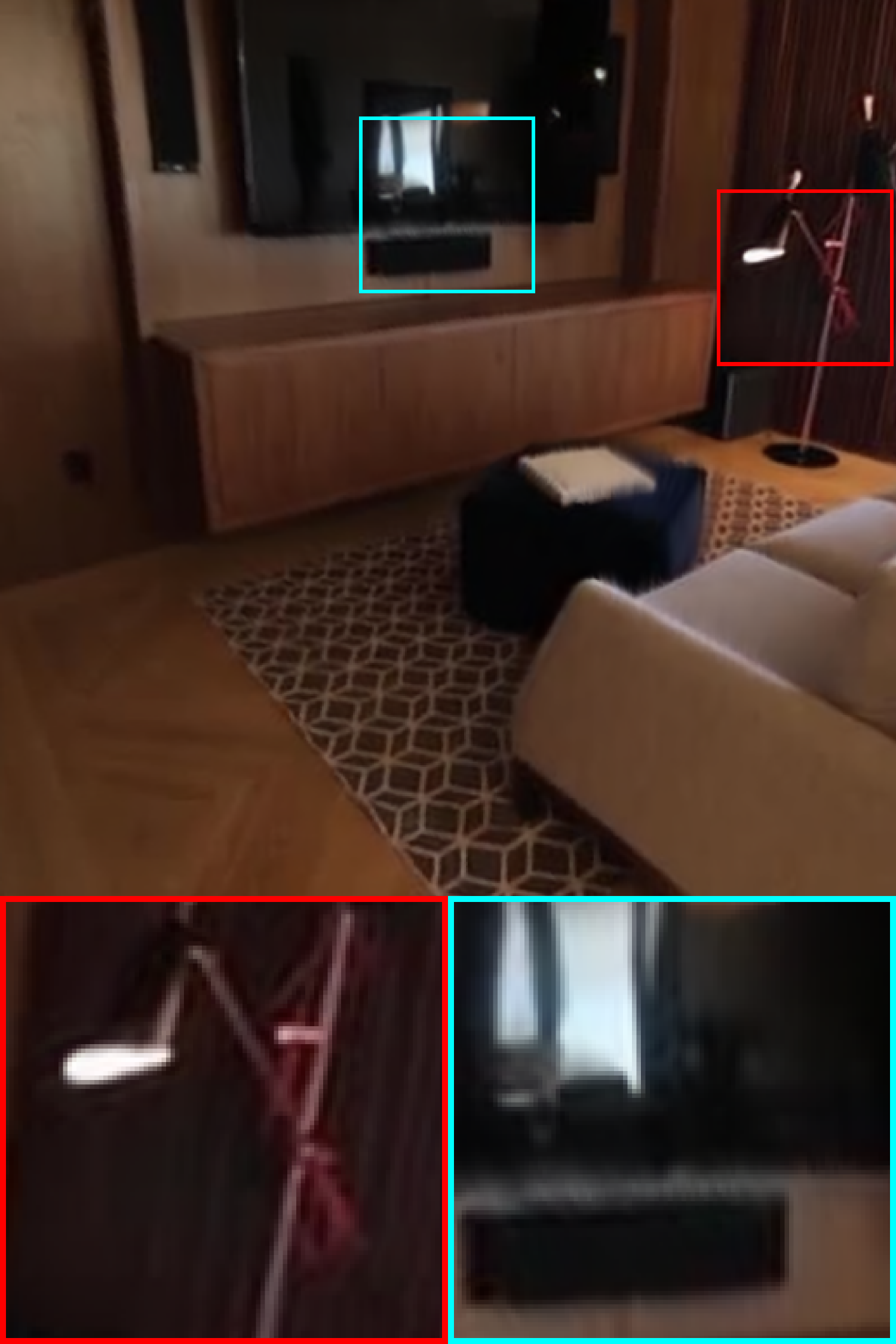} &
        \includegraphics[width=\imgsize]{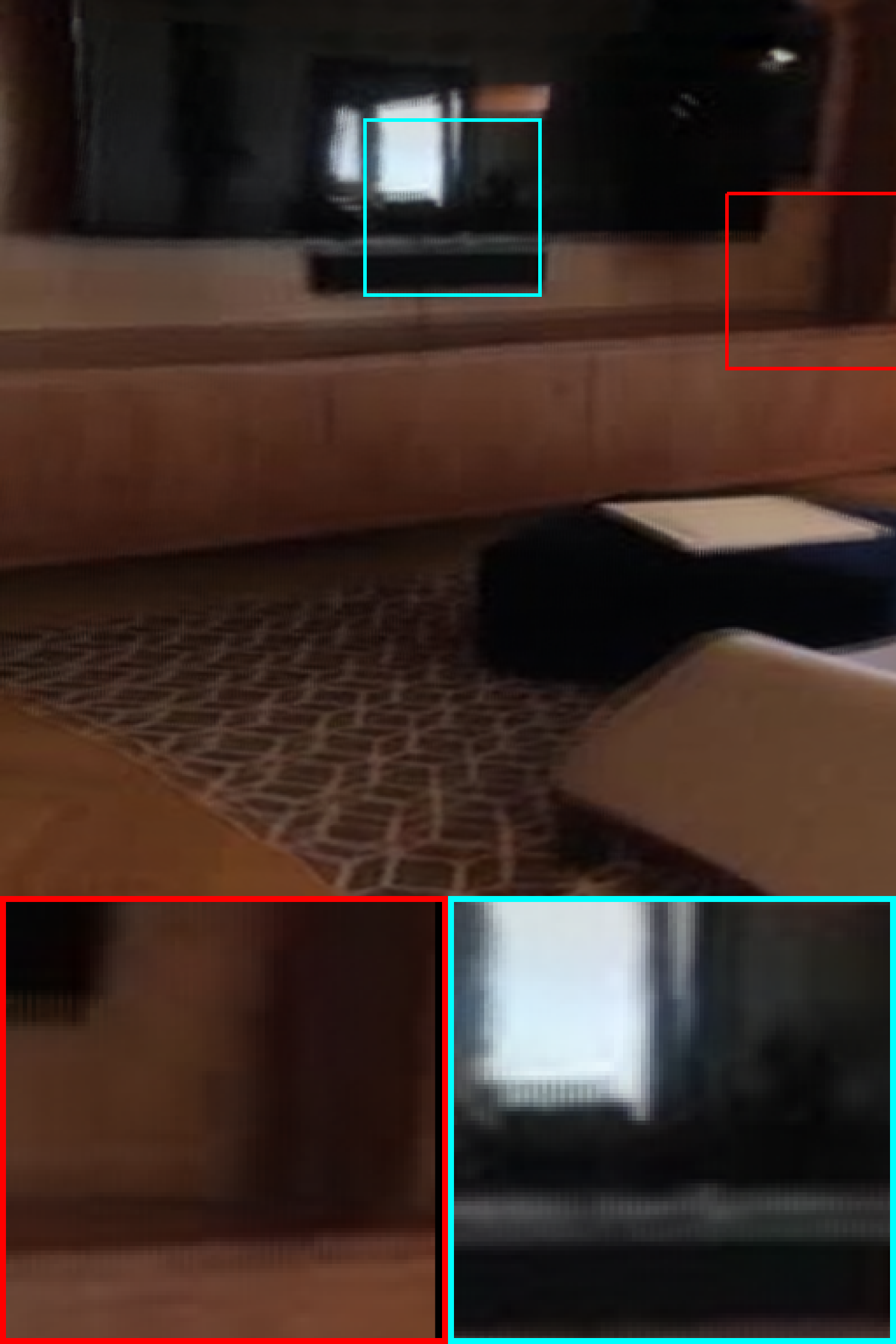} \\
        \includegraphics[width=\imgsize]{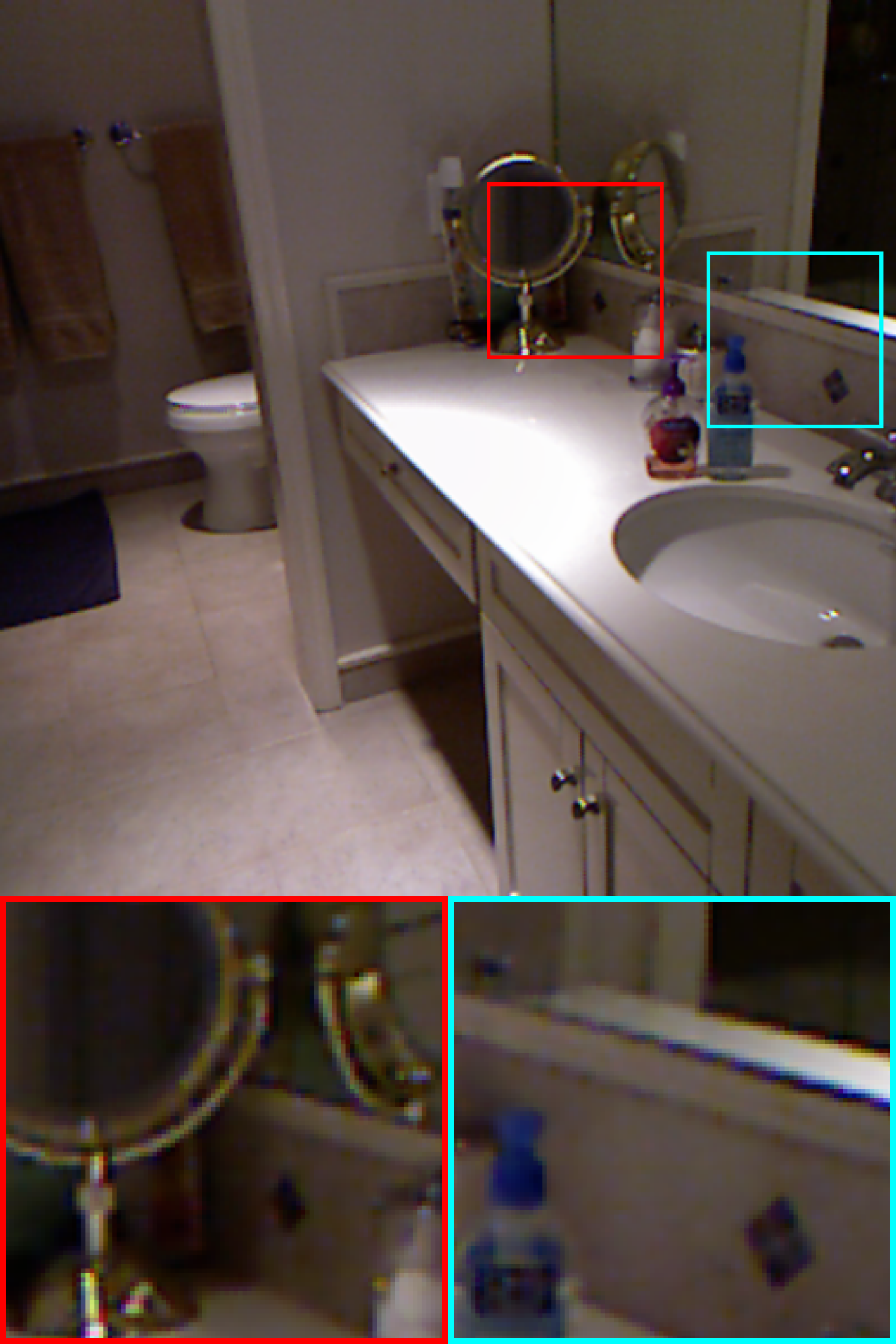} &
        \includegraphics[width=\imgsize]{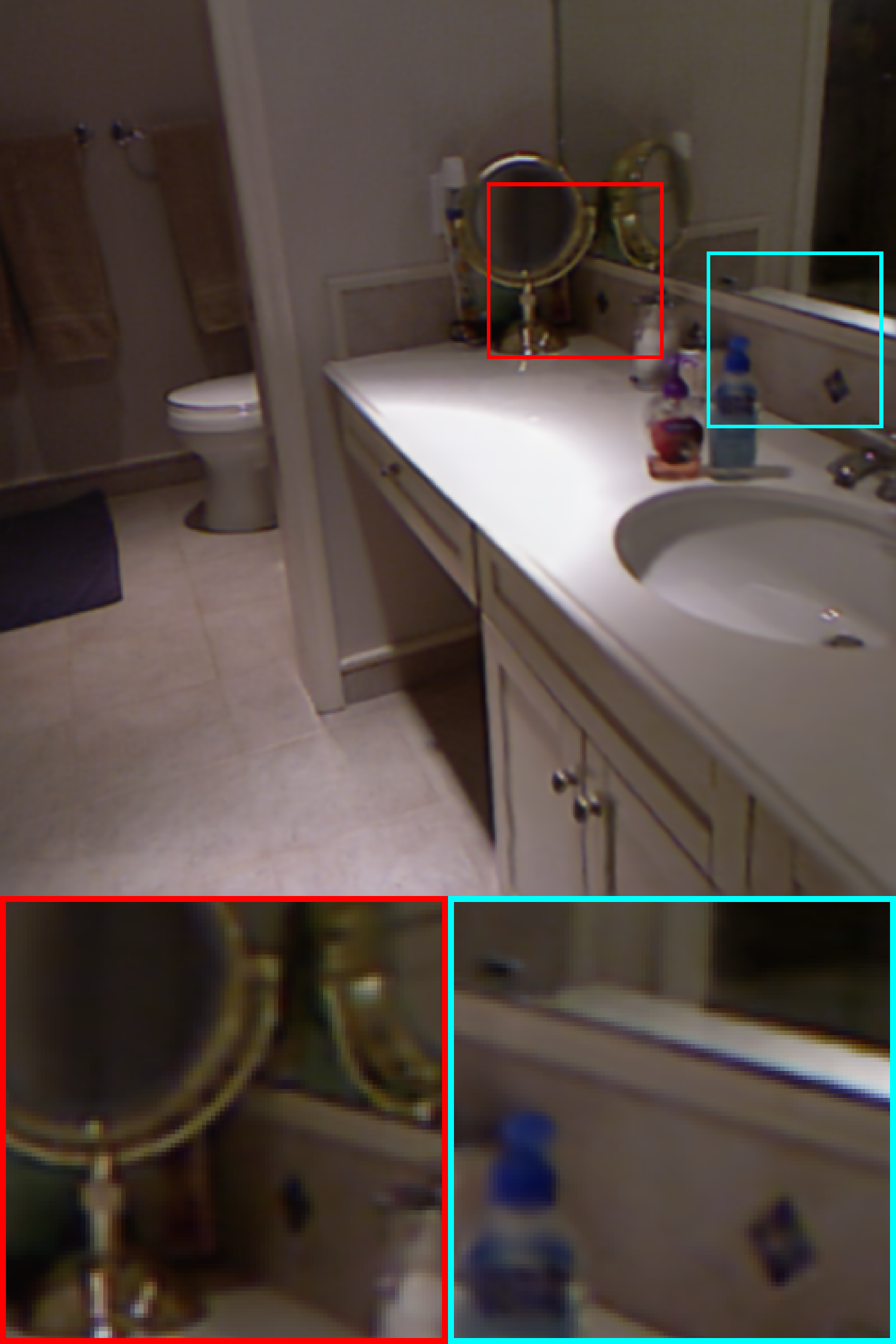} &
        \includegraphics[width=\imgsize]{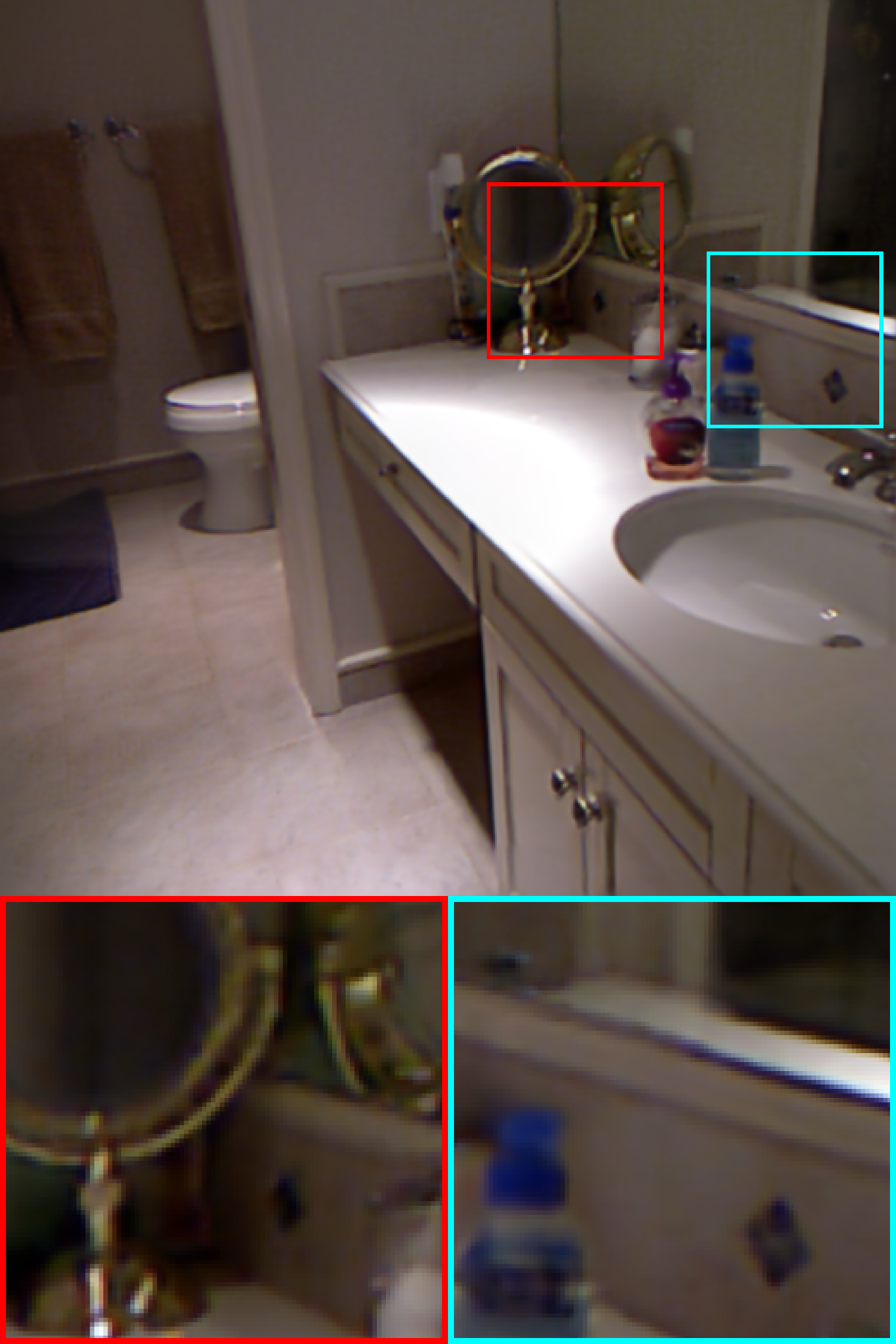} &
        \includegraphics[width=\imgsize]{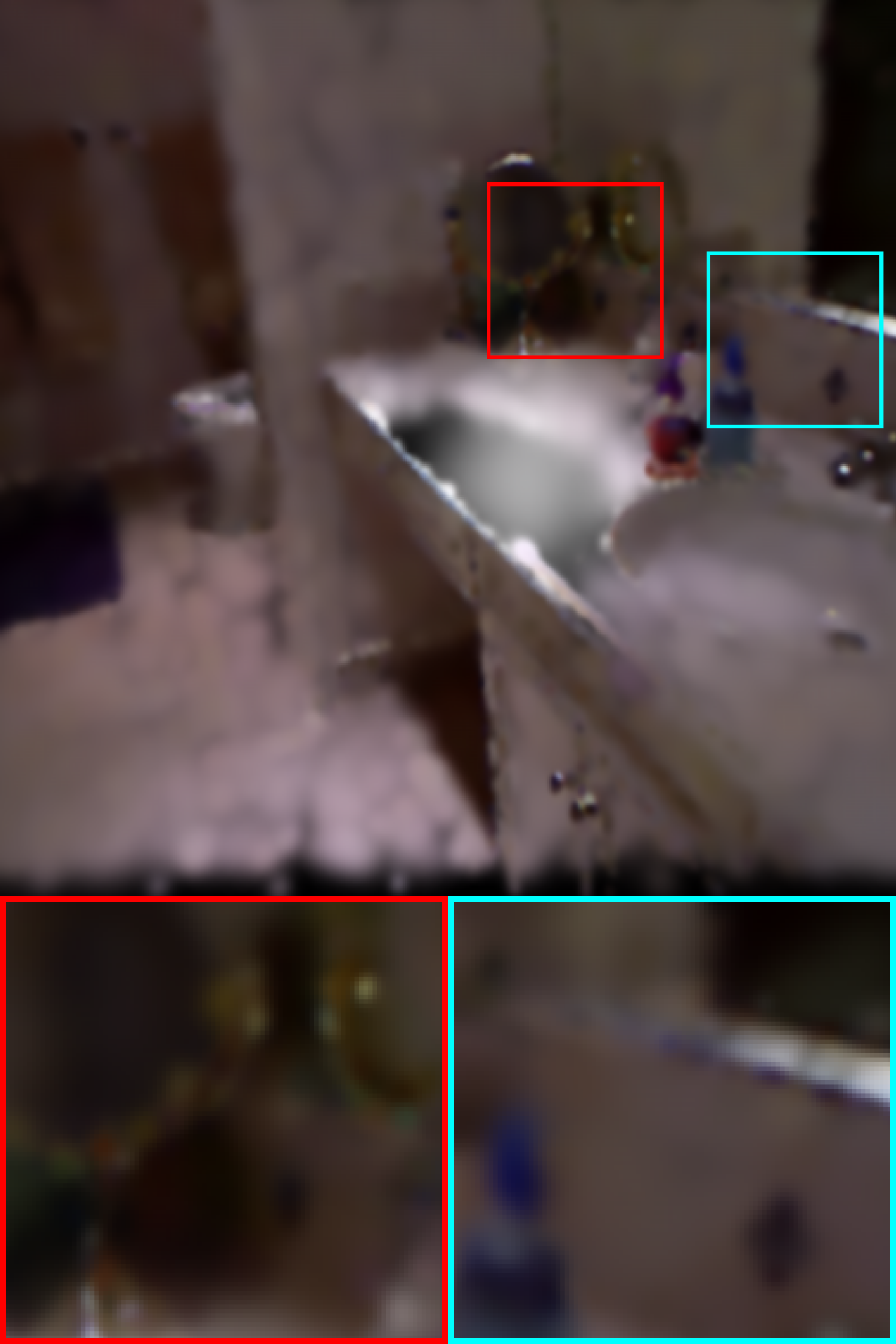} &
        \includegraphics[width=\imgsize]{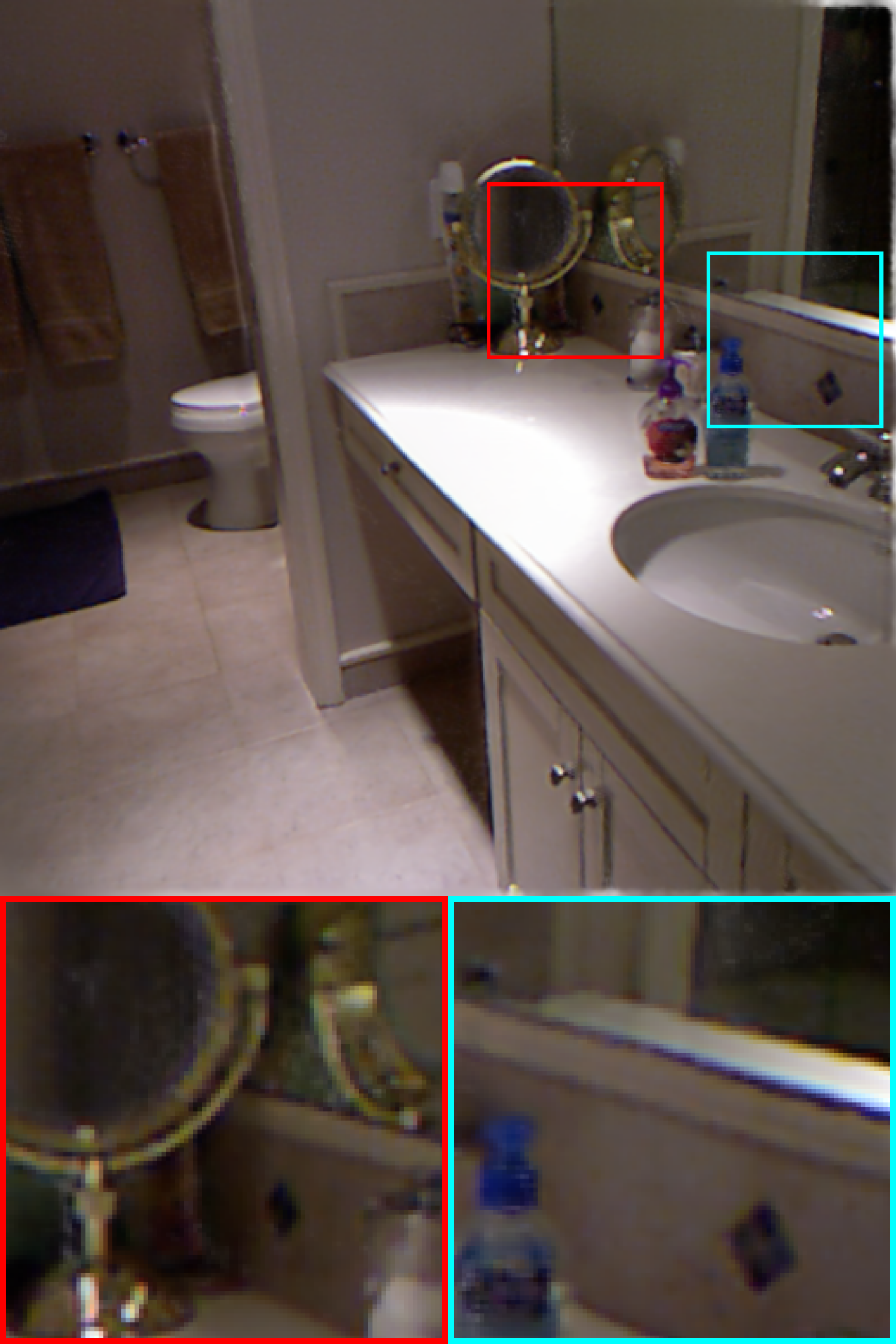} &
        \includegraphics[width=\imgsize]{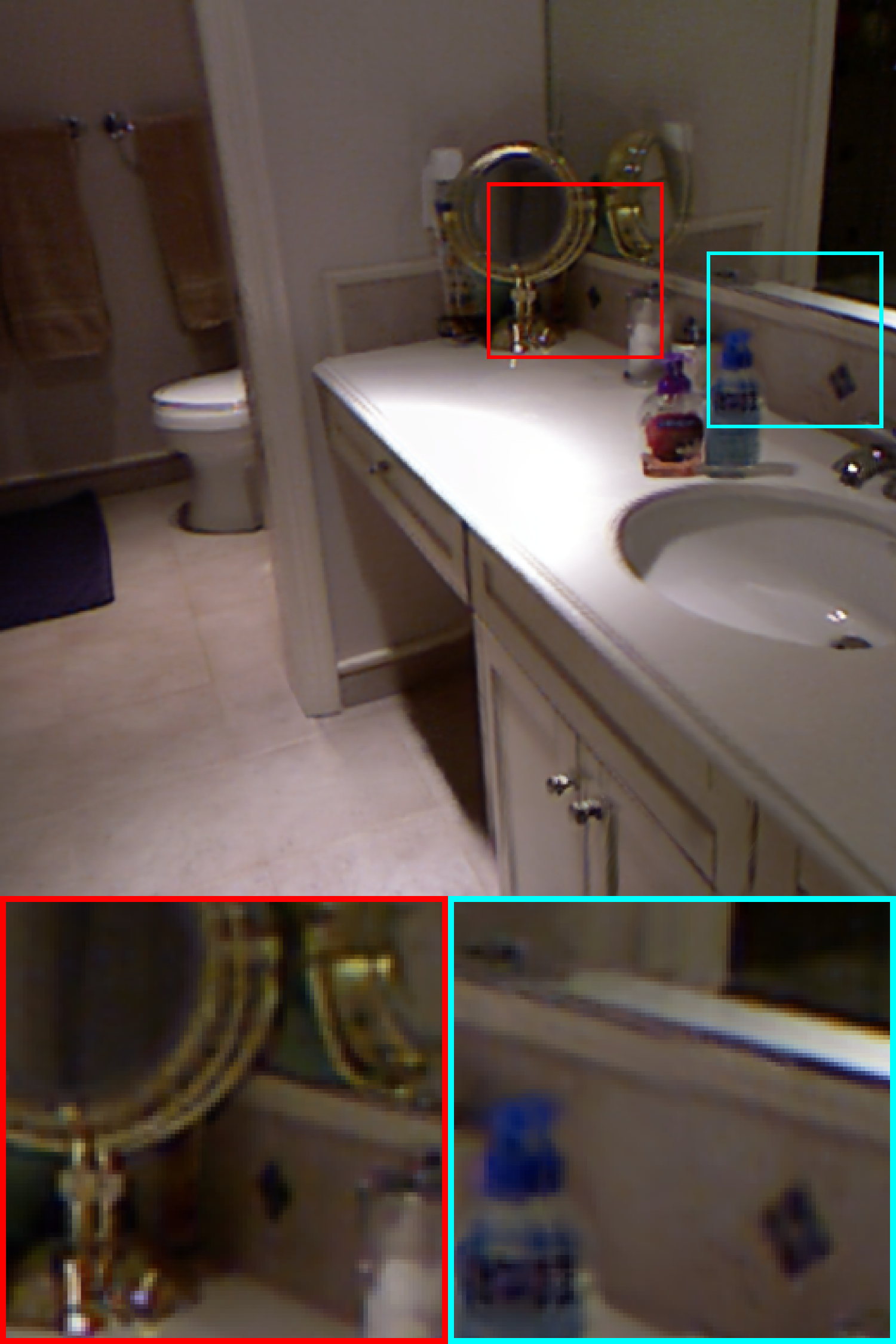} &
        \includegraphics[width=\imgsize]{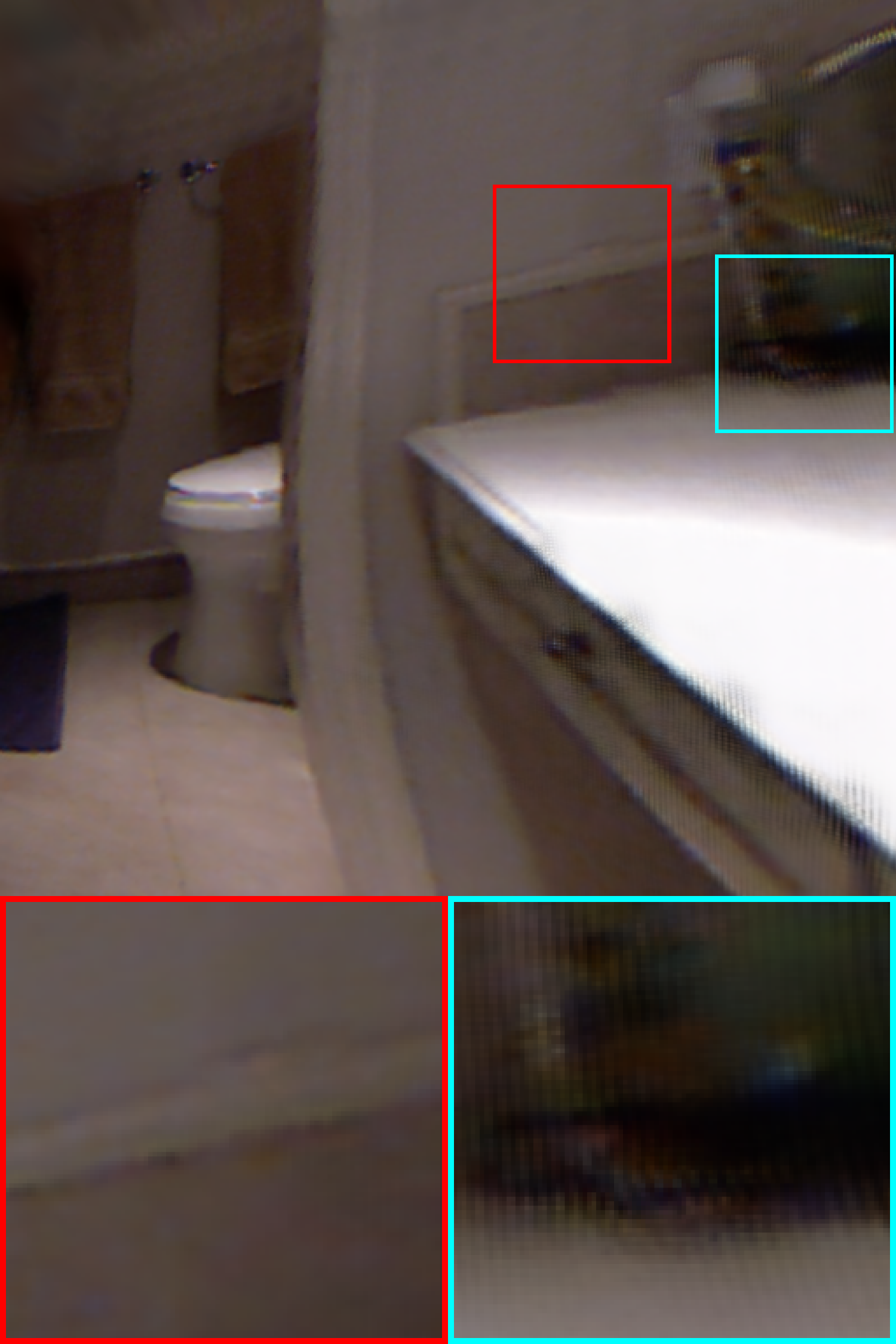} \\
        \scriptsize GT &
        \scriptsize \textbf{Ours} &
        \scriptsize SVGGT+GS &
        \scriptsize OnTheFly-NVS &
        \scriptsize WorldMirror &
        \scriptsize AnySplat &
        \scriptsize FLARE \\
    \end{NiceTabular}
    
    \caption{Qualitative comparison on \dlbench~(top row), \re~(middle row), and unseen \nyu~(bottom row) datasets. Our method produces high-quality novel view renderings while preserving fine structural details.}
    \label{fig:fig_nvs}
    \vspace{-13pt}
\end{figure*}

\begin{figure*}[!t]
    \centering
    \begin{tabular}{@{}c@{\hspace{3pt}}c@{}}
        \includegraphics[width=0.455\linewidth]{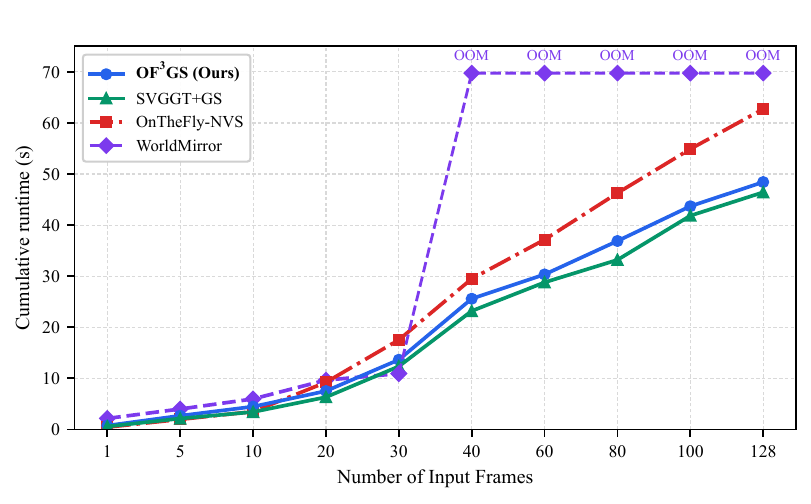} &
        \includegraphics[width=0.455\linewidth]{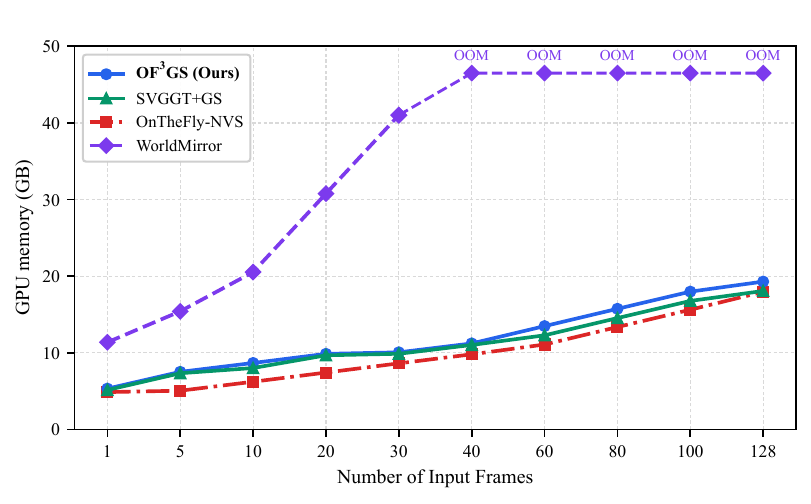}
    \end{tabular}
    \caption{End-to-end runtime and memory comparison on \dl under 128 input views, measured on a single NVIDIA RTX 5880 Ada GPU. Left: average cumulative wall-clock time after sequentially processing frames $1{:}t$. Right: GPU memory recorded after processing frame $t$.}
    \label{fig:runtime_memory}
    \vspace{-18pt}
\end{figure*}

\begin{table*}[t]
        \centering
        \resizebox{1.0\linewidth}{!}{
            \begin{tabular}{@{}l ccc | ccc | ccc@{}}
                \toprule
                \multirow{2}{*}{\textbf{Method}} & 
                \multicolumn{3}{c|}{\textbf{DL3DV}} & 
                \multicolumn{3}{c|}{\textbf{RE10K}} & 
                \multicolumn{3}{c}{\textbf{NYUv2 (Gen.)}} \\
                \cmidrule(lr){2-4} \cmidrule(lr){5-7} \cmidrule(l){8-10}
                 & PSNR $\uparrow$ & SSIM $\uparrow$ & LPIPS $\downarrow$ 
                 & PSNR $\uparrow$ & SSIM $\uparrow$ & LPIPS $\downarrow$ 
                 & PSNR $\uparrow$ & SSIM $\uparrow$ & LPIPS $\downarrow$ \\
                \midrule
                w/o NV-Sup & 21.188 & 0.626 & \second{0.298} & 25.210 & 0.810 & 0.232 & 24.645 & \second{0.708} & \second{0.298} \\
                w/o DIR-Head     & 20.546 & 0.629 & 0.331 & 23.677 & 0.774 & 0.241 & 22.964 & 0.675 & 0.321 \\
                w/o DPR-Offsets  & \second{21.348} & \second{0.653} & 0.302 & \second{25.412} & \second{0.816} & \second{0.210} & \second{24.698} & 0.701 & \second{0.298} \\
                \midrule
                \textbf{Ours (Full)} & \best{\textbf{21.884}} & \best{\textbf{0.688}} & \best{\textbf{0.273}} & \best{\textbf{25.797}} & \best{\textbf{0.833}} & \best{\textbf{0.200}} & \best{\textbf{24.875}} & \best{\textbf{0.717}} & \best{\textbf{0.291}} \\
                \bottomrule
            \end{tabular}
        }
        \caption{Ablation study of \ours, evaluated under $5$ input views.}
        \label{tab:tab_abl}
\end{table*}

\begin{figure*}[htbp]
    \centering
    \newlength{\figW}
    \setlength{\figW}{0.30\textwidth} 

    \newcommand{\sublabel}[2]{%
        \makebox[0.5\figW][c]{\scriptsize #1}%
        \makebox[0.5\figW][c]{\scriptsize #2}%
    }
    \begin{NiceTabular}{ccc}
        \includegraphics[width=\figW]{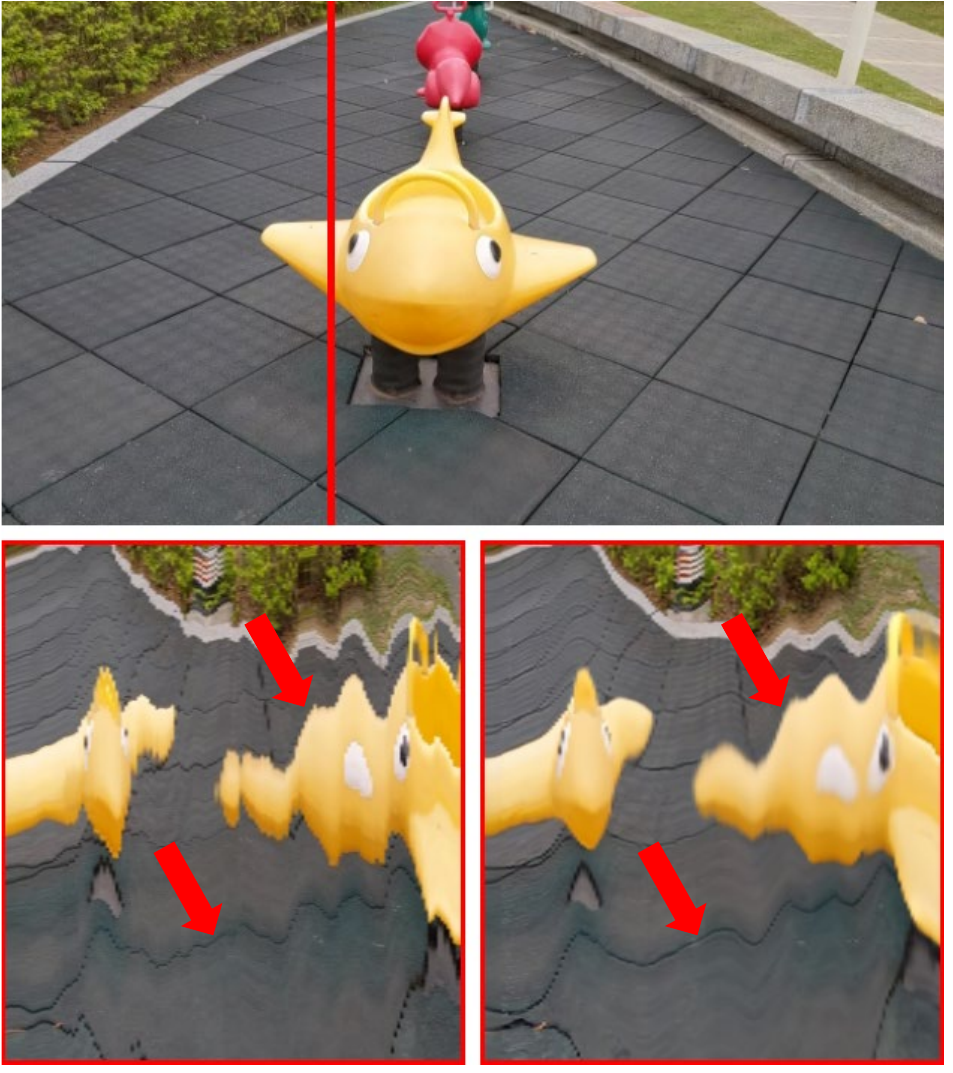} &
        \includegraphics[width=\figW]{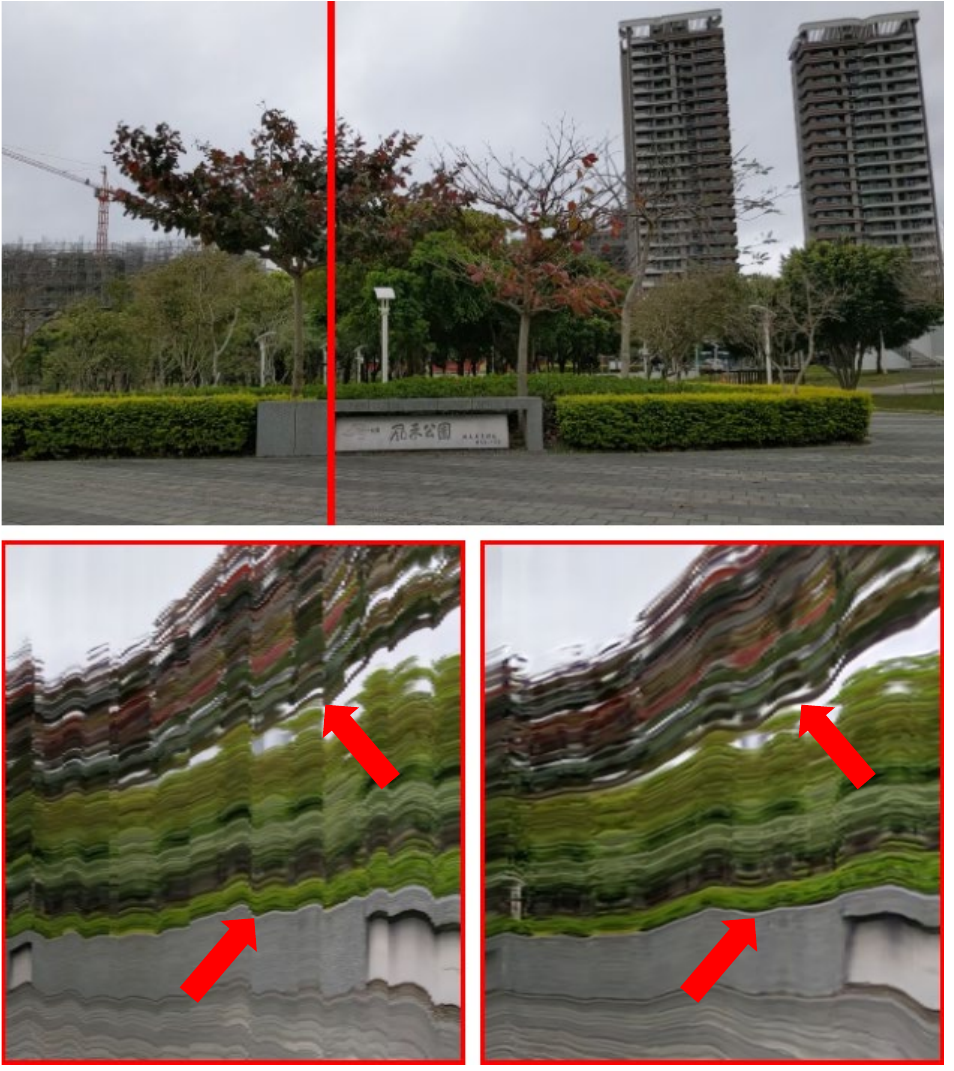} &
        \includegraphics[width=\figW]{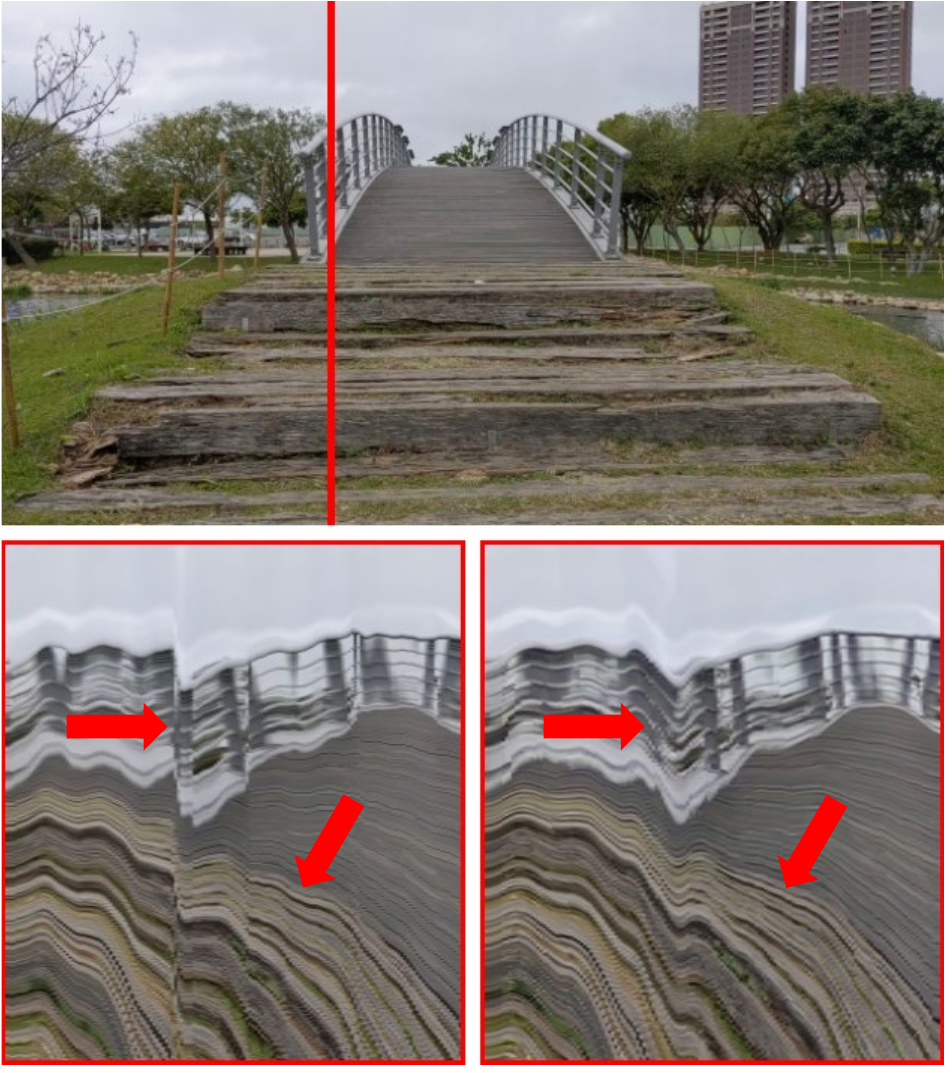} \\
        
        \sublabel{Unstable}{Stable (Ours)} & 
        \sublabel{Unstable}{Stable (Ours)} & 
        \sublabel{Unstable}{Stable (Ours)} \\
    \end{NiceTabular}

    \caption{Downstream demonstration of \ours on on-the-fly video stabilization~(tested on 150-frame hand-held sequences). \textbf{Top}: reference frames with red lines marking sampled columns. \textbf{Bottom}: temporal profiles (X-T slices), generated by horizontally concatenating the sampled column across all frames.}
    \vspace{-7pt}
    \label{fig:fig_sta}
\end{figure*}



\noindent\textbf{Novel View Synthesis and Efficiency.}
\Tref{tab:tab_nvs_main} evaluates NVS from $5$ to $128$ input views. The sparse settings ($5/10$ views) test core causal reconstruction under limited observations, where \ours consistently outperforms online baselines such as \onthefly and \sgs while remaining competitive with offline SOTA despite using only causal observations. Although \sgs runs faster than \ours in \fref{fig:runtime_memory}, its reconstruction quality is clearly weaker, especially under sparse inputs. As shown in \fref{fig:fig_nvs}, \ours also preserves sharper structures and details under sparse inputs.

The denser settings ($64/128$ views) evaluate resilience against catastrophic geometric collapse and error accumulation as redundant observations and state size increase. Under official all-input protocols, offline feed-forward baselines report OOM, whereas \ours maintains a recursive state and remains competitive with \onthefly, with lower cumulative runtime in \fref{fig:runtime_memory}. Compared with \sgs, \ours trades higher runtime for consistently better rendering quality, indicating that the gains come from causal geometric stabilization rather than the frozen backbone alone. Overall, \ours offers a practical speed-quality-memory trade-off in dense-view settings.

\noindent\textbf{Cross-dataset Generalization.}
We evaluate zero-shot generalization on unseen \nyu scenes without fine-tuning. As shown in \Tref{tab:tab_nvs_main} and the \nyu row in \fref{fig:fig_nvs}, \ours remains competitive under domain shift and surpasses the offline SOTA \wm in the sparse 5-view setting, indicating robust causal reconstruction beyond the training domains.

\noindent\textbf{Ablation Study.}
As shown in \Tref{tab:tab_abl}, we ablate three core components under $5$ input views on \dlbench, \re, and \nyu. Removing DIR-Head causes the largest degradation, showing that stable intrinsic recovery is critical for preventing recursive scale drift in causal unposed reconstruction. Removing DPR-Offsets also consistently weakens performance, as ray-constrained Gaussian updates cannot fully absorb coupled pose-depth errors during fusion. Without NV-Sup, the model loses part of its novel-view consistency because training becomes more biased toward input-view appearance. These results verify that the three designs jointly stabilize sparse-view causal reconstruction; additional qualitative evidence and analysis are provided in the supplementary material.

\noindent\textbf{Application: On-the-fly Video Stabilization.}
We further apply \ours to on-the-fly video stabilization for short hand-held clips. Specifically, we smooth the estimated camera extrinsic trajectory using a Savitzky--Golay filter~\cite{savitzky1964smoothing} and perform NVS along the smoothed trajectory based on the reconstructed 3DGS. As illustrated by the flattened temporal profiles (X--T slices) in \fref{fig:fig_sta}, high-frequency camera shakes are effectively suppressed over 150-frame sequences, yielding smoother temporal evolution and stable online video rendering. Additional dynamic results are provided in the supplementary video.

\section{Conclusion}
In this work, we presented \textit{\ours}, an on-the-fly feed-forward framework for high-fidelity 3DGS reconstruction and NVS from unposed images. By introducing the \textit{DIR-Head} and \textit{DPR-Offsets}, we effectively mitigate recursive intrinsic drift and rectify Gaussian placement distortions stemming from rigid viewing-direction constraints, thereby restoring the geometric consistency typically weakened in causal settings. Extensive experiments demonstrate strong sparse-view causal NVS quality, with additional evaluations showing memory-feasible inference in dense-view settings.

\noindent\textbf{Limitations.} 
Despite these advancements, \ours is not designed for ultra-long videos, where bounded memory and long-range consistency become critical. It also remains challenged by highly dynamic scenes and time-varying camera intrinsics, which require stronger temporal modeling and more flexible intrinsic recovery.

\clearpage
\appendix
\setcounter{secnumdepth}{1}
\section*{Supplementary Material}
\setcounter{figure}{0}
\setcounter{table}{0}
\setcounter{equation}{0}
\renewcommand{\thesection}{\Alph{section}}
\renewcommand{\theequation}{\Alph{section}.\arabic{equation}}
\renewcommand{\thefigure}{\Alph{section}.\arabic{figure}}
\renewcommand{\thetable}{\Alph{section}.\arabic{table}}

\section*{Overview} 
In this supplementary material, we provide additional details organized as follows:
\begin{itemize}
    \item \Sref{sec:supp_impl}: 
    Additional implementation details, covering data processing and sampling strategies.
    
    \item \Sref{sec:supp_loss}: 
    Elaboration on the loss formulation, specifically detailing the geometric loss.

    \item \Sref{sec:supp_abl}: 
    Extension of the ablation study, providing qualitative evidence and detailed analysis of the contribution of each key component.
    
    \item \Sref{sec:supp_fail}: 
    Qualitative analysis and discussion of the failure case, presenting representative artifacts in challenging scenarios.
    
    \item \Sref{sec:supp_qual}: 
    Extensive qualitative experiments, showcasing additional visual results.
\end{itemize}

\noindent\textbf{Supplementary Video.} 
We strongly recommend to watch the accompanying video, which visualizes the incremental reconstruction process of our online feed-forward 3DGS and demonstrates the dynamic results of video stabilization applications.

\section{Implementation Details}
\label{sec:supp_impl}

\noindent\textbf{Data Processing and Sampling.} 
During training, we simulate the online process by randomly sampling video clips with a sequence length $L \in [2, 16]$. All input images are resized to a fixed width of 518 pixels, with the height scaled proportionally to maintain the original aspect ratio. For each clip, 50\% of the frames are randomly selected as target views for photometric supervision.

\noindent\textbf{Baseline and Training Protocol.}
Following the main-paper baseline protocol, all methods are evaluated at a consistent resolution of 518 pixels in width. We train \ours with AdamW~\cite{loshchilov2017decoupled} using a global batch size of 4, with one sample per GPU, for approximately one day on four NVIDIA RTX 5880 Ada GPUs (48GB). The loss weights are set to $\lambda_1=0.05$ for SSIM, $\lambda_2=0.05$ for LPIPS, $\lambda_{\text{novel}}=1.5$, $\lambda_{\text{depth}}=1.0$, and $\lambda_{\text{int}}=10.0$.

For the reproducible online feed-forward reference \sgs, we keep the same data processing, training schedule, optimizer, resolution, and hyperparameters as \ours. It uses the same frozen \svggt feature extractor and causal extrinsic head, but removes the three proposed components studied in the ablation: DIR-Head, DPR-Offsets, and NV-Sup.

For network initialization, the full-history causal feature extractor and causal extrinsic head are initialized from \svggt and kept frozen during training. This use of pretrained priors follows a common and practical paradigm in recent feed-forward 3D reconstruction and 3DGS methods, including \anysplat and \svggt. Our contribution is not to train a new foundation backbone, but to stabilize online feed-forward 3DGS under the same frozen backbone. Since training is performed in an unposed setting without ground-truth camera labels, these priors help avoid convergence instability or degenerate solutions. Our DPR-Offsets further compensate for residual pose drift at the Gaussian primitive level, making additional extrinsic optimization unnecessary in practice.

\noindent\textbf{Unposed NVS Evaluation Protocol.}
For feed-forward methods, we follow the two-pass evaluation protocol of \any. Specifically, this protocol is applied to \any, \wm, \sgs, and \ours. In the first pass, the context and target images are jointly fed into the camera estimation network to obtain relative camera poses for both context and target views, denoted as $\mathcal{P}^{\mathrm{joint}}_{\mathrm{ctx}}$ and $\mathcal{P}^{\mathrm{joint}}_{\mathrm{tgt}}$, respectively. Only these camera estimates are retained from the first pass.

In the second pass, each method is restarted and receives only the context images for actual 3DGS reconstruction. \sgs and \ours process the context images sequentially, whereas \any and \wm process the same context set jointly according to their official inference protocols. This pass produces the reconstructed Gaussian representation together with another set of estimated context-camera poses, denoted as $\mathcal{P}^{\mathrm{rec}}_{\mathrm{ctx}}$. After both passes are completed, we use the context cameras shared by the two pose sets to estimate the coordinate transformation that aligns $\mathcal{P}^{\mathrm{joint}}_{\mathrm{ctx}}$ with $\mathcal{P}^{\mathrm{rec}}_{\mathrm{ctx}}$. We then apply the same transformation to $\mathcal{P}^{\mathrm{joint}}_{\mathrm{tgt}}$ and render the reconstructed Gaussians from the aligned target viewpoints. Therefore, target images are used only in the first pass to recover their query-camera poses and are never used for Gaussian prediction, fusion, or scene reconstruction in the second pass.

For each scene, PSNR, SSIM, and LPIPS are first averaged over all target views. The reported dataset-level scores are then computed by averaging the scene-level results across all test scenes, giving equal weight to each scene.

\section{Details of Loss}
\label{sec:supp_loss}

\noindent\textbf{Geometric Loss.} 
In our geometric loss (scale-shift invariant geometric supervision), we align the predicted depth $\mathbf{D}_t$ to the pseudo-ground-truth $\mathbf{D}^{\text{prior}}_t$ using a least-squares approach. 
Specifically, we flatten the depth maps into vectors $\mathbf{p}_t$ and $\mathbf{q}_t$, respectively. To stabilize the alignment, we center them to zero mean: $\mathbf{p}^c_t = \mathbf{p}_t - \mu^p_t$ and $\mathbf{q}^c_t = \mathbf{q}_t - \mu^q_t$, where $\mu^p_t$ and $\mu^q_t$ are the mean values of the predictions and targets. 
The optimal per-frame scale $\gamma_t$ and shift $\beta_t$ are then derived as:
\begin{equation}
    \gamma_t = \frac{{\mathbf{p}^c_t}^\top \mathbf{q}^c_t}{{\mathbf{p}^c_t}^\top \mathbf{p}^c_t + \epsilon}, \quad \beta_t = \mu^q_t - \gamma_t \cdot \mu^p_t,
\end{equation}
where $\epsilon = 10^{-6}$ is a small constant for numerical stability. The final aligned prediction $\mathbf{D}^{\text{align}}_t = \gamma_t \cdot \mathbf{D}_t + \beta_t$ is used for the geometric loss calculation in the main paper.

\section{More Qualitative Ablations}
\label{sec:supp_abl}

\begin{figure*}[t!]
    \makeatletter\setlength{\@fptop}{0pt}\makeatother
    \centering
    \newcommand{\imgsize}{0.19\textwidth} 
    \setlength{\tabcolsep}{1pt} 
    \renewcommand{\arraystretch}{0.5} 
    
    \begin{NiceTabular}{ccccc}
        \includegraphics[width=\imgsize]{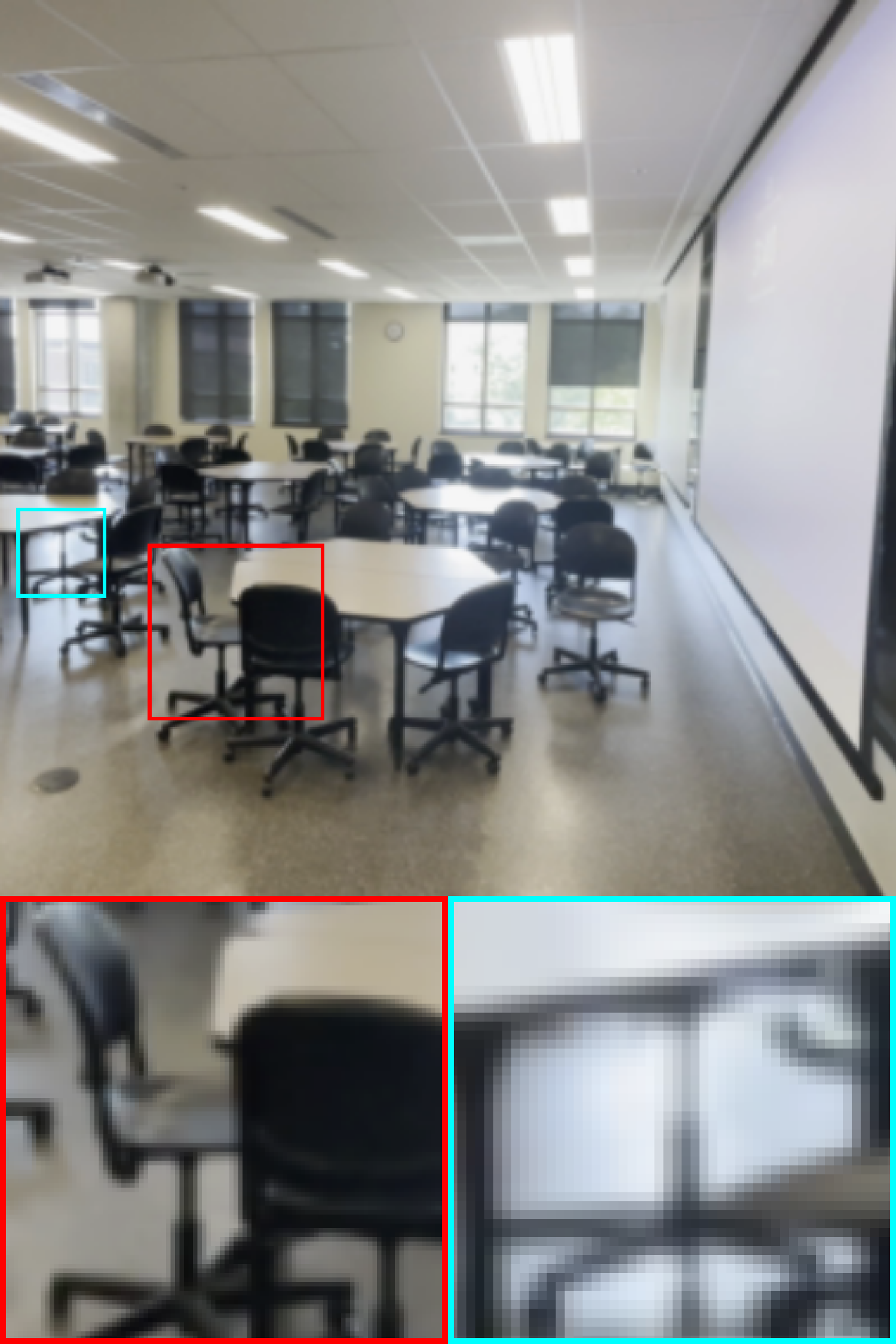} &
        \includegraphics[width=\imgsize]{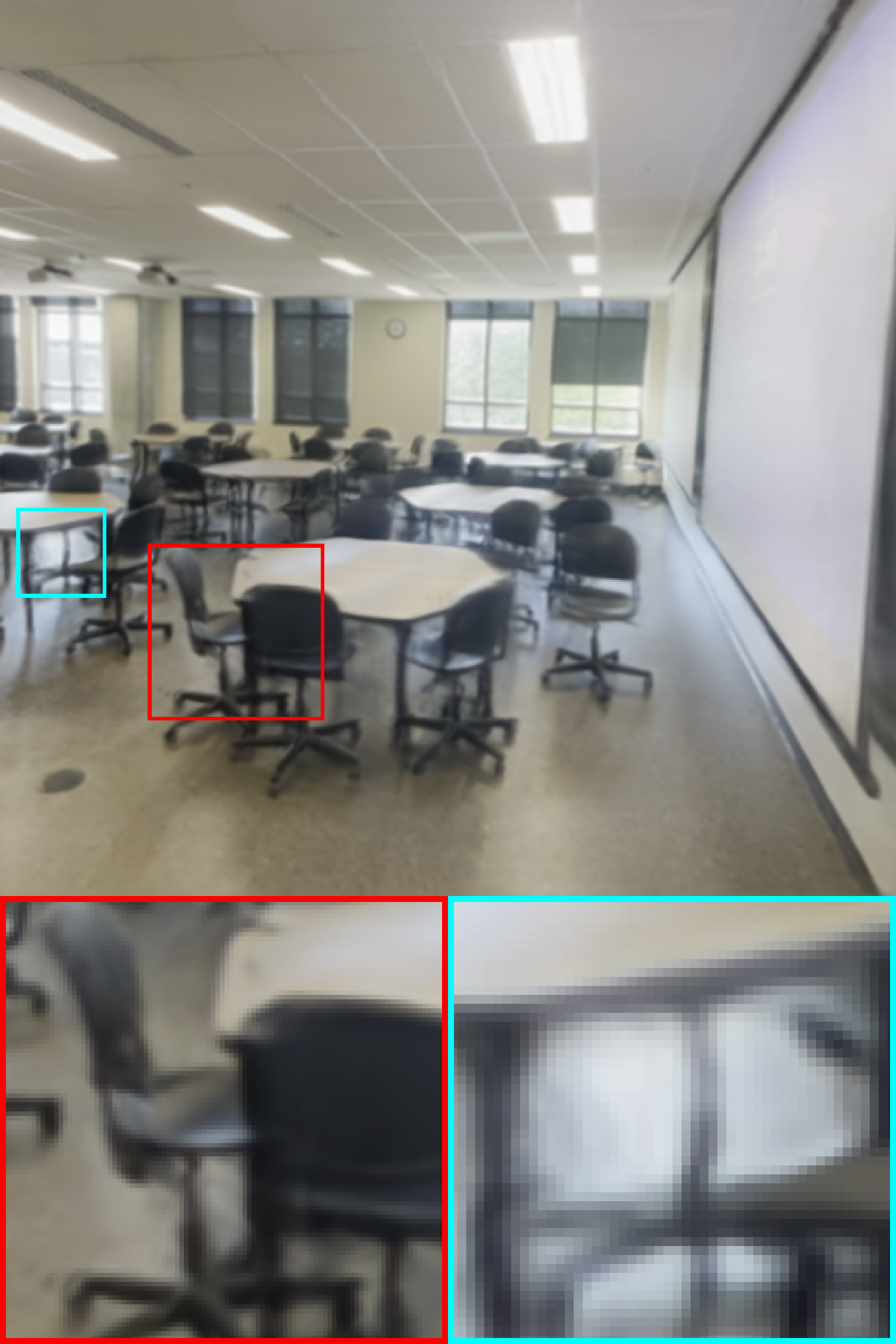} &
        \includegraphics[width=\imgsize]{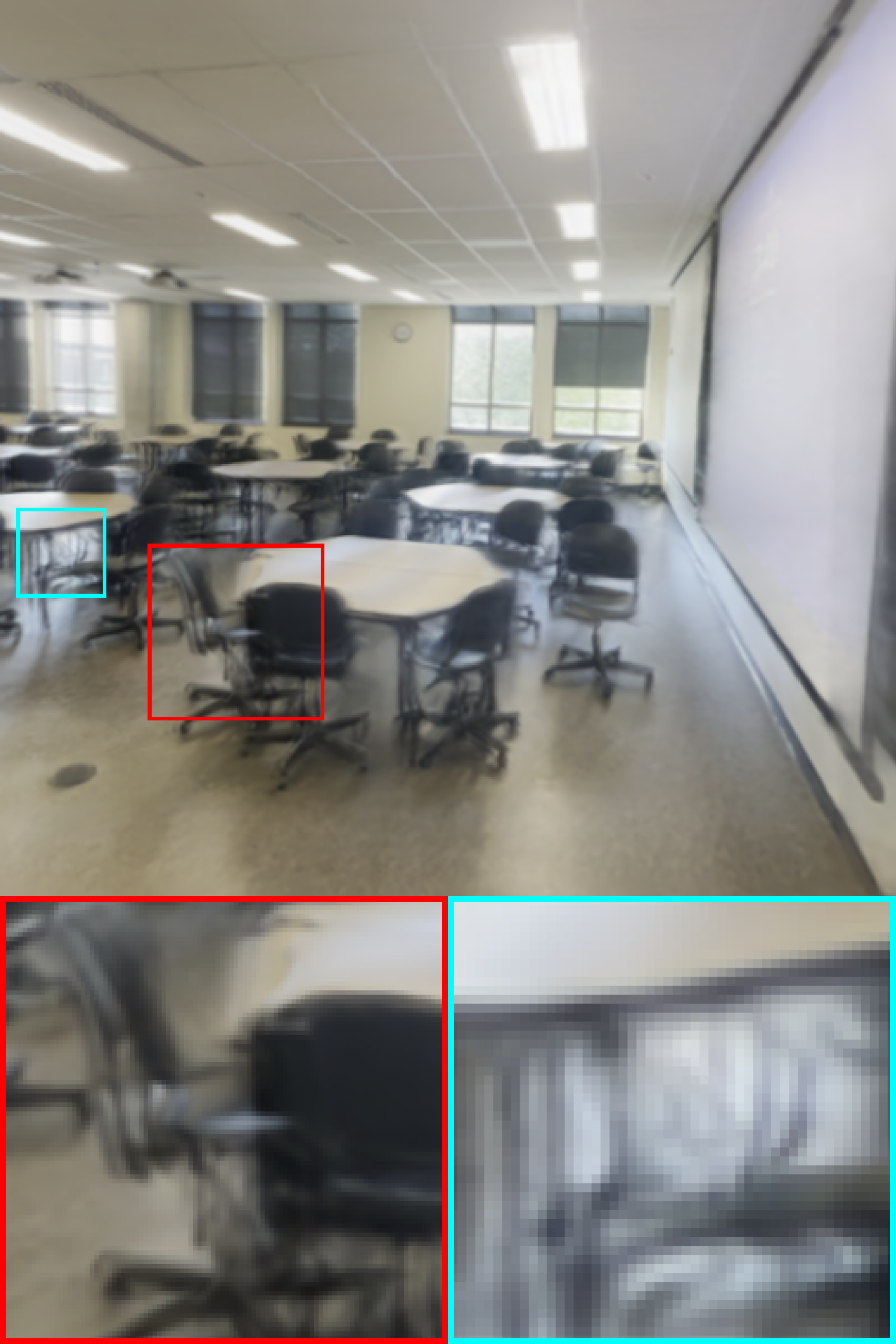} &
        \includegraphics[width=\imgsize]{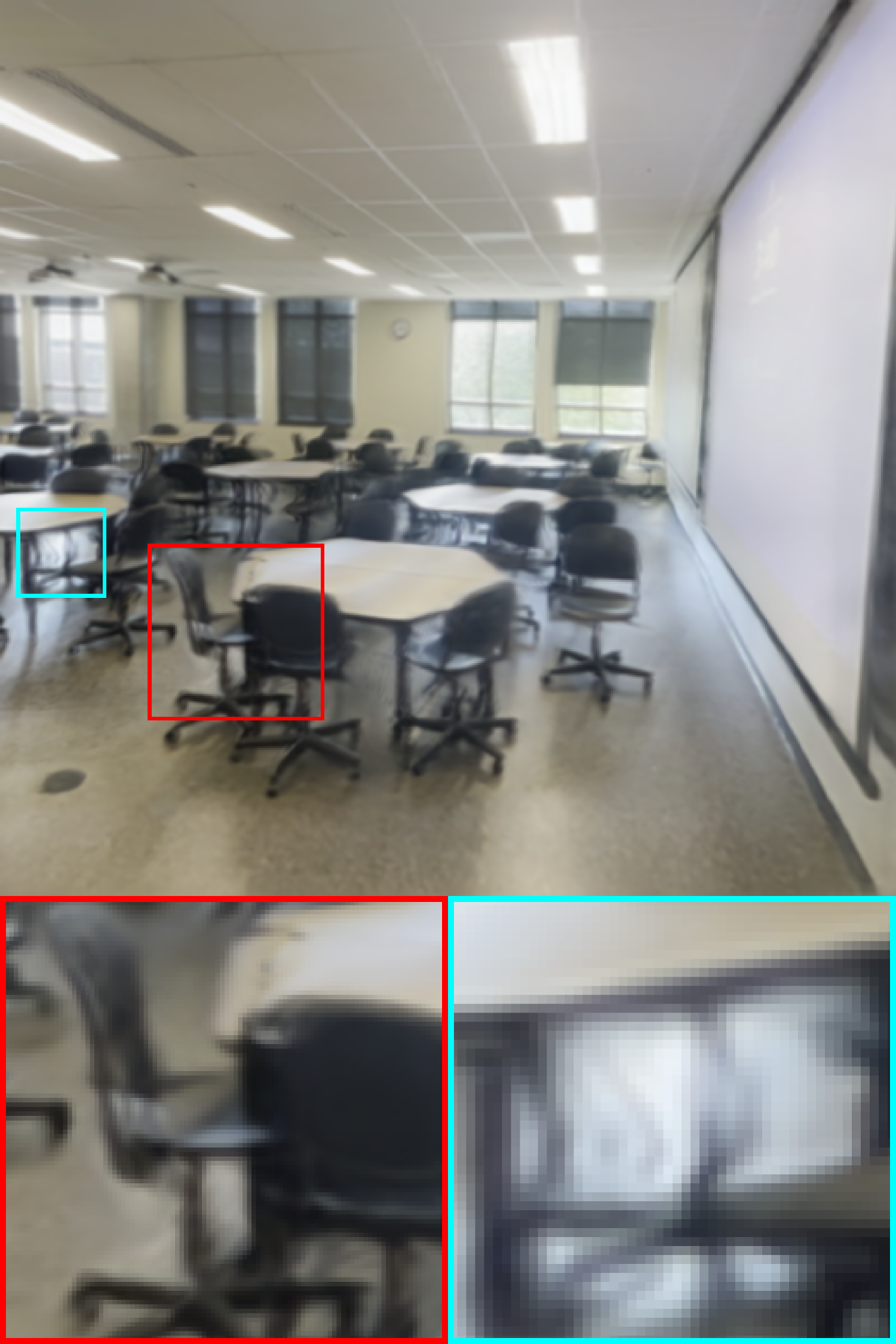} &
        \includegraphics[width=\imgsize]{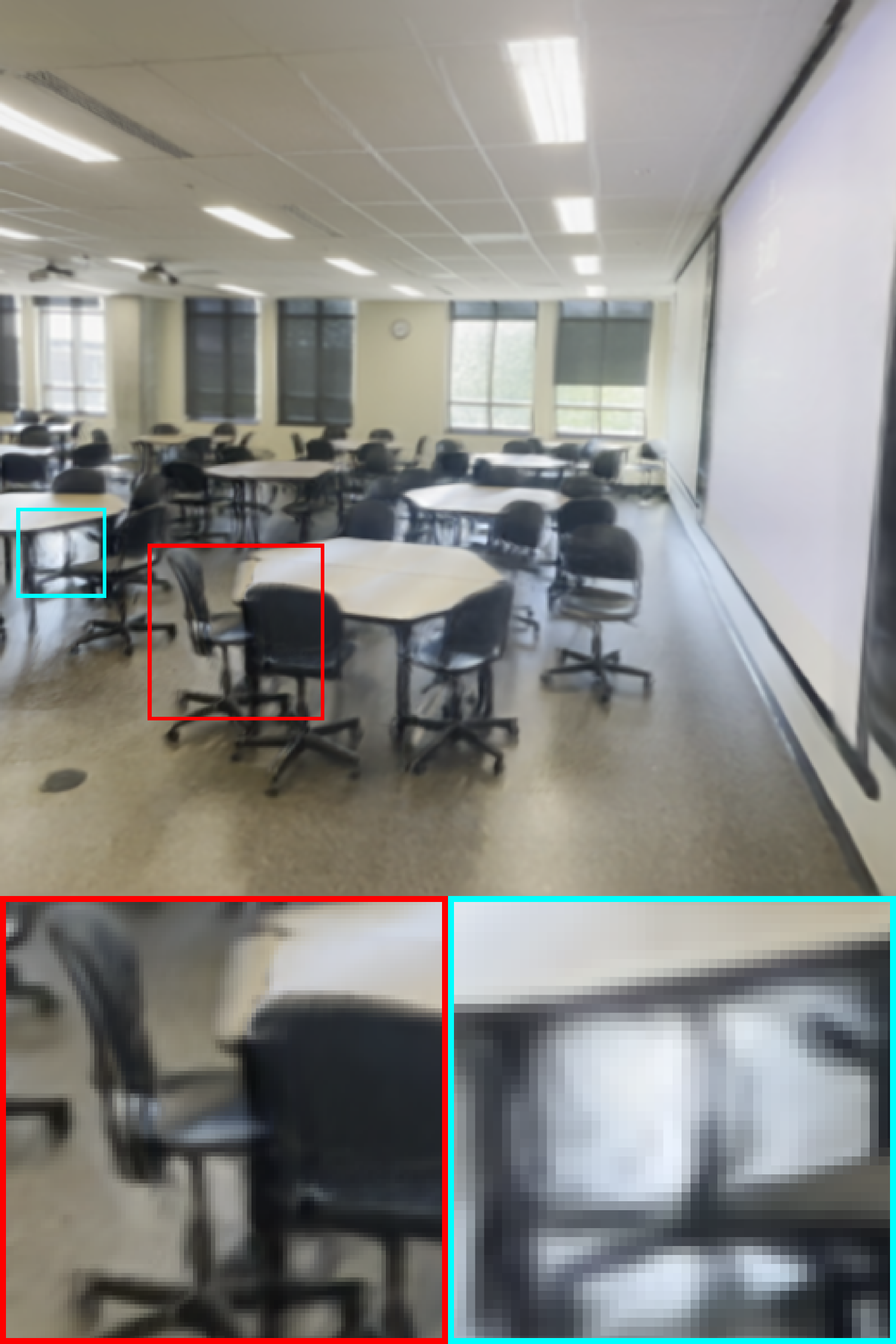} \\
        \includegraphics[width=\imgsize]{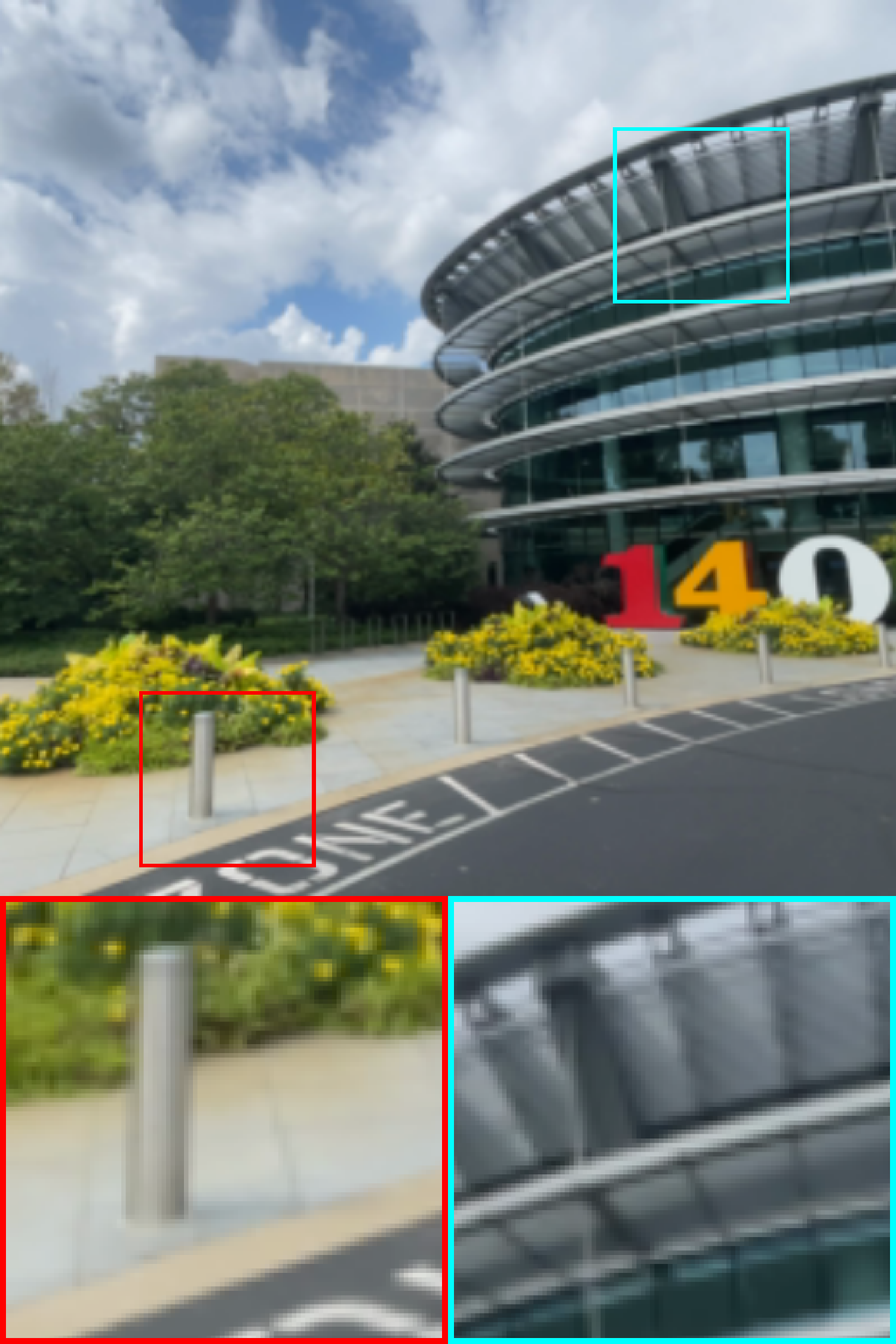} &
        \includegraphics[width=\imgsize]{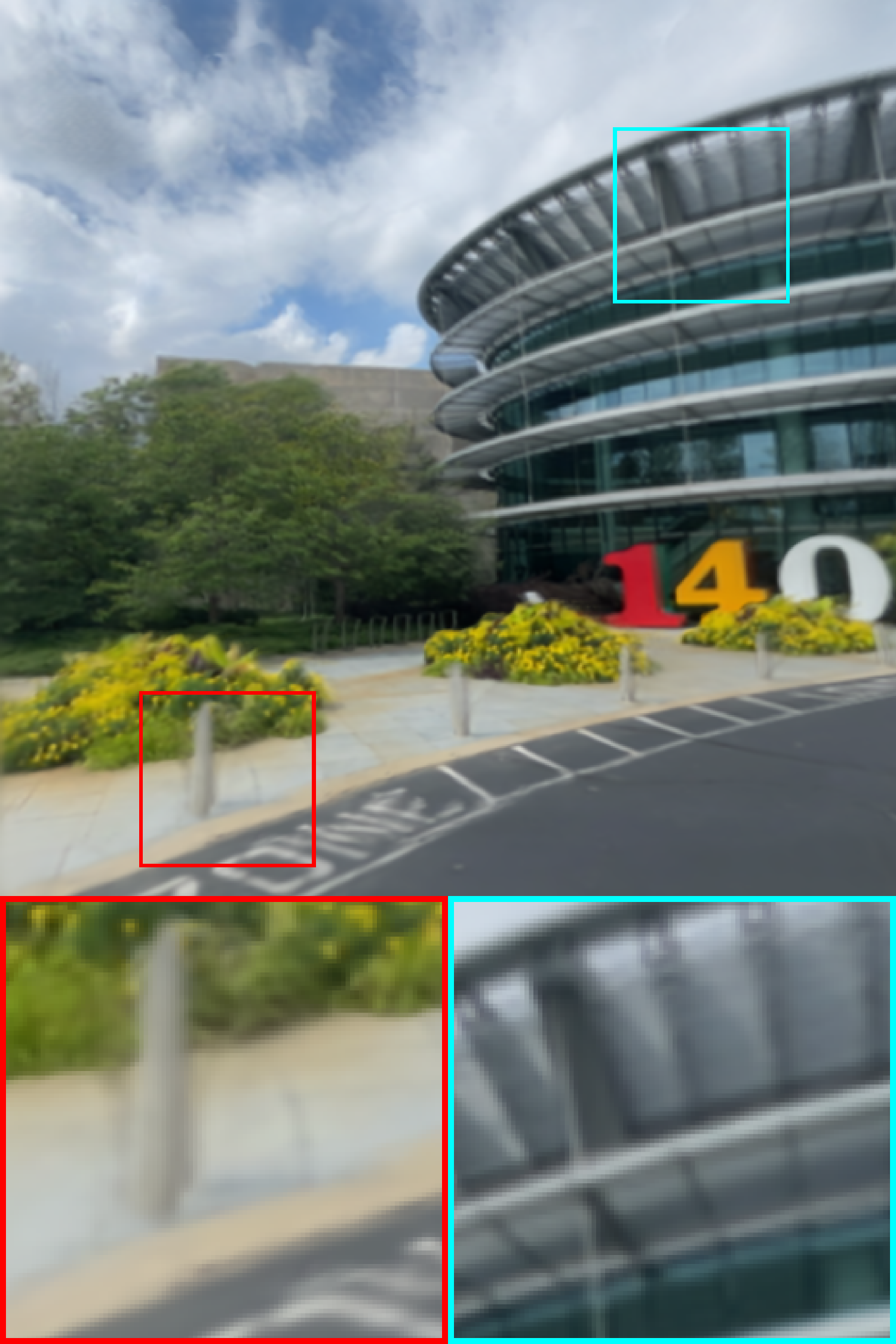} &
        \includegraphics[width=\imgsize]{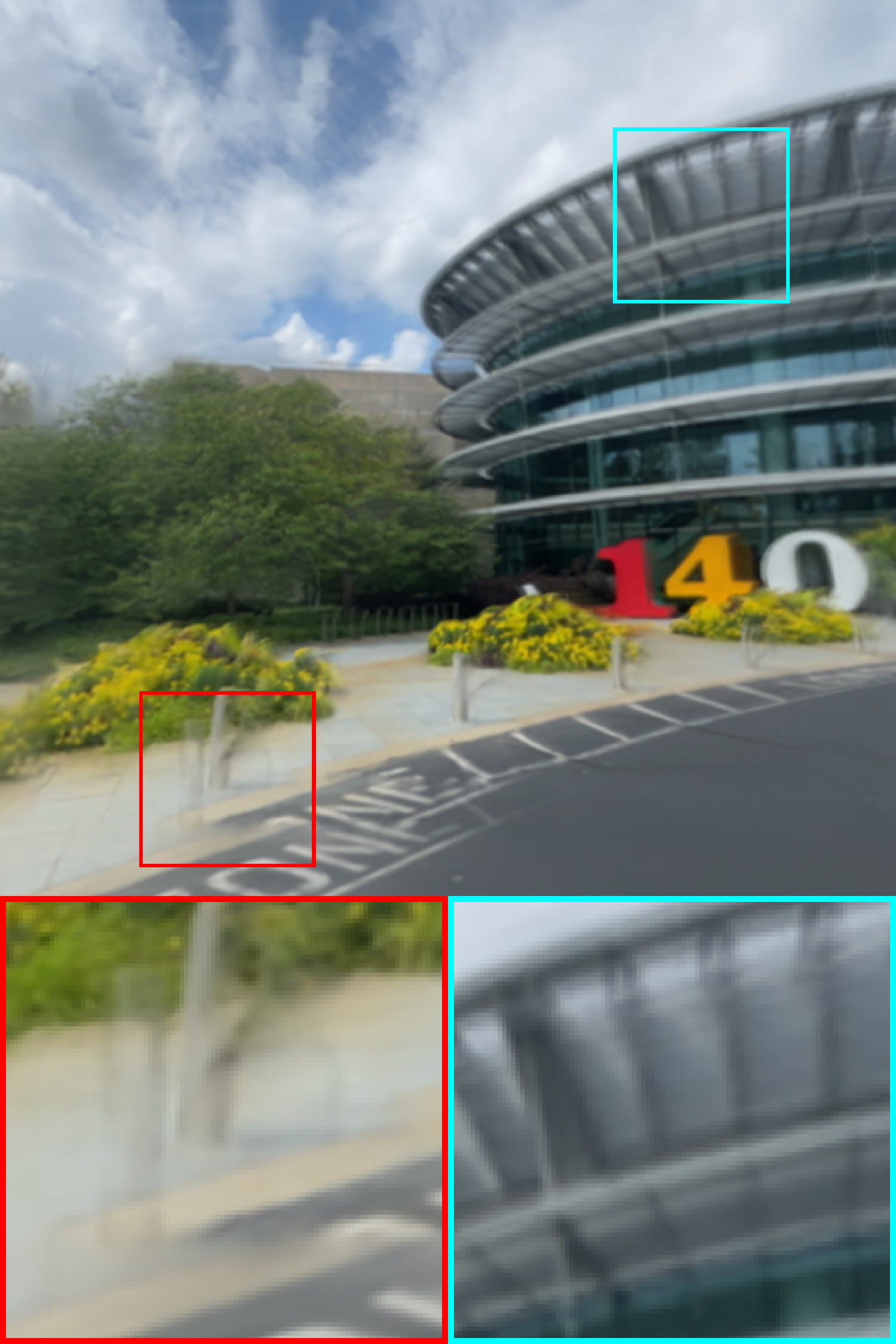} &
        \includegraphics[width=\imgsize]{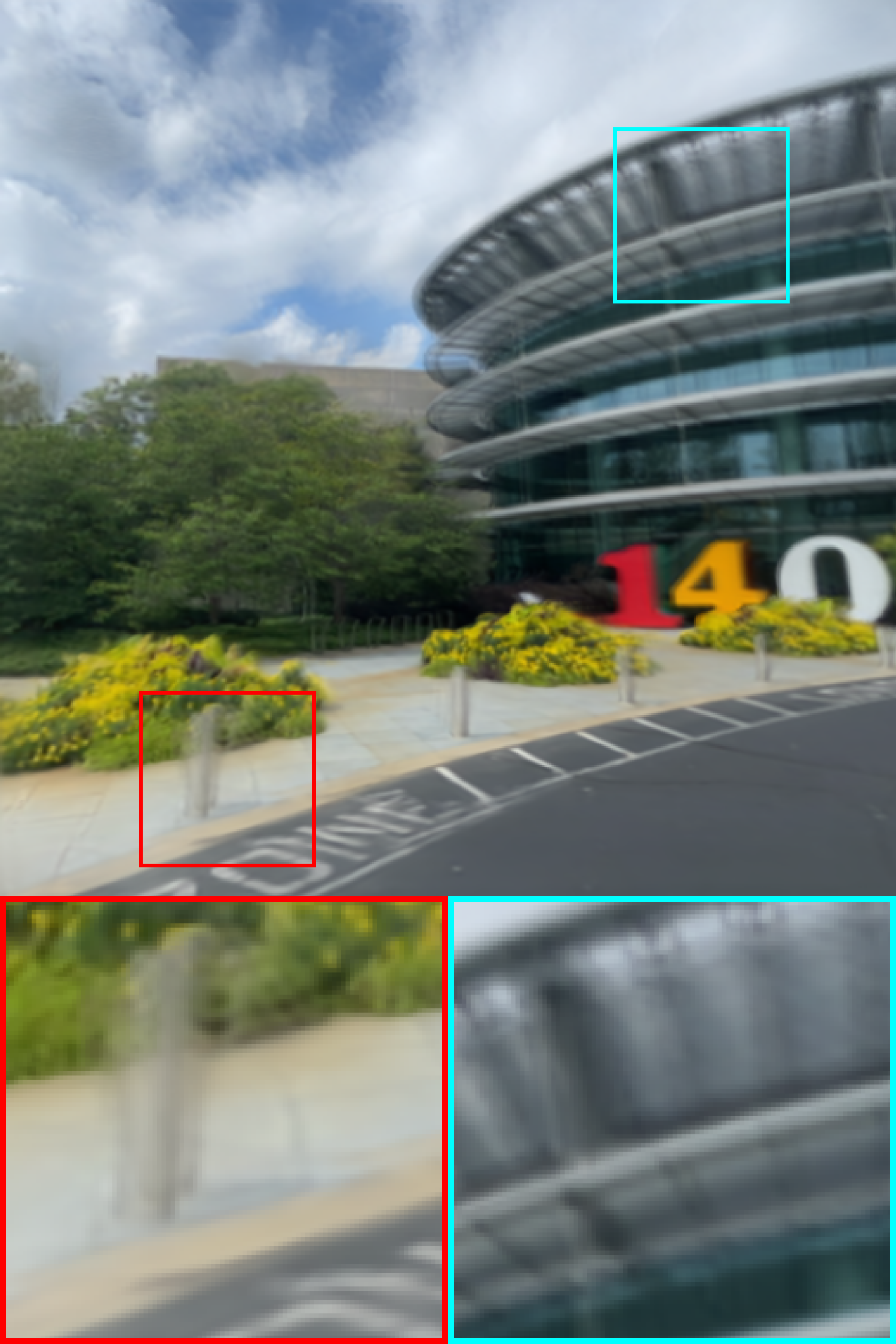} &
        \includegraphics[width=\imgsize]{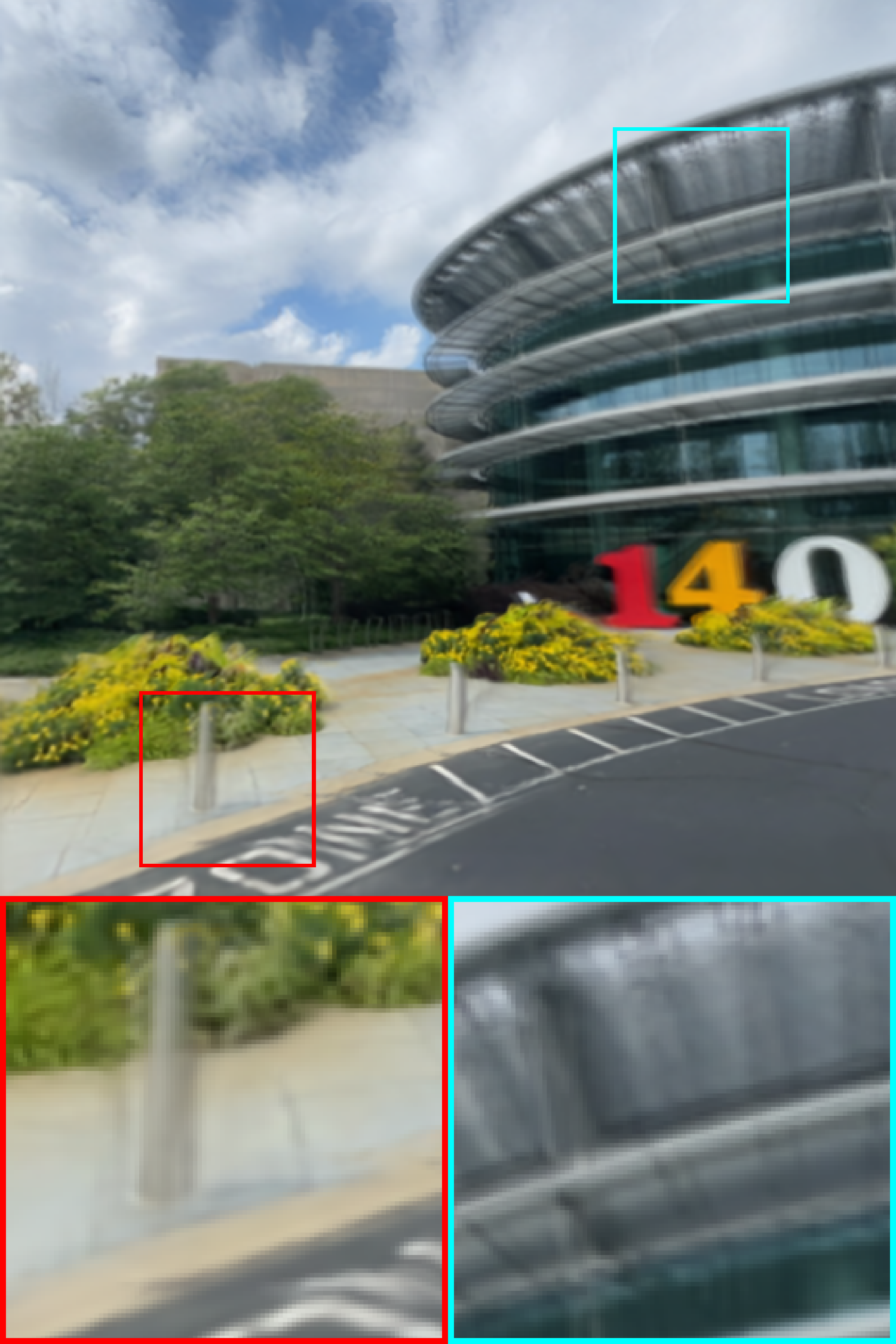} \\

        \includegraphics[width=\imgsize]{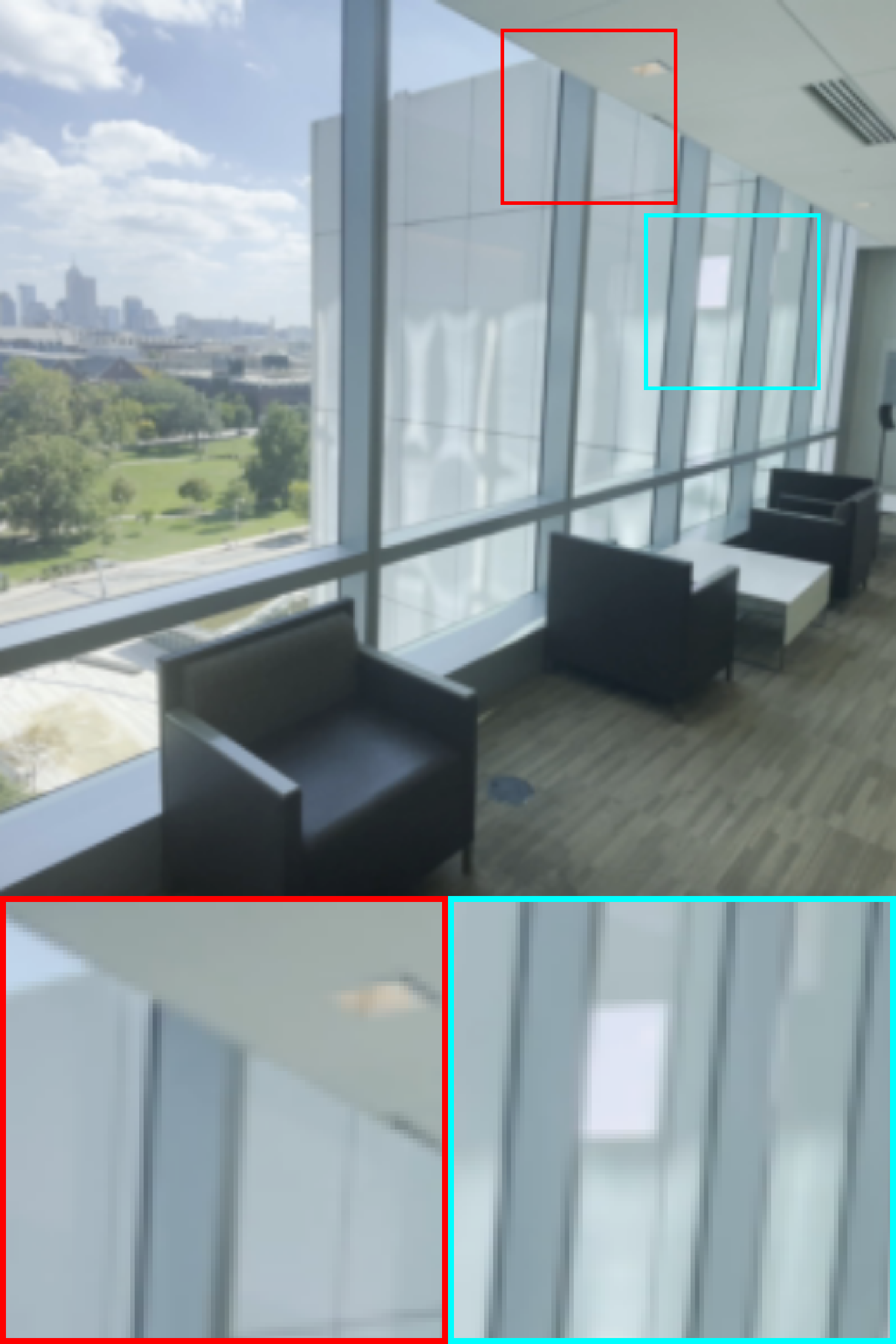} &
        \includegraphics[width=\imgsize]{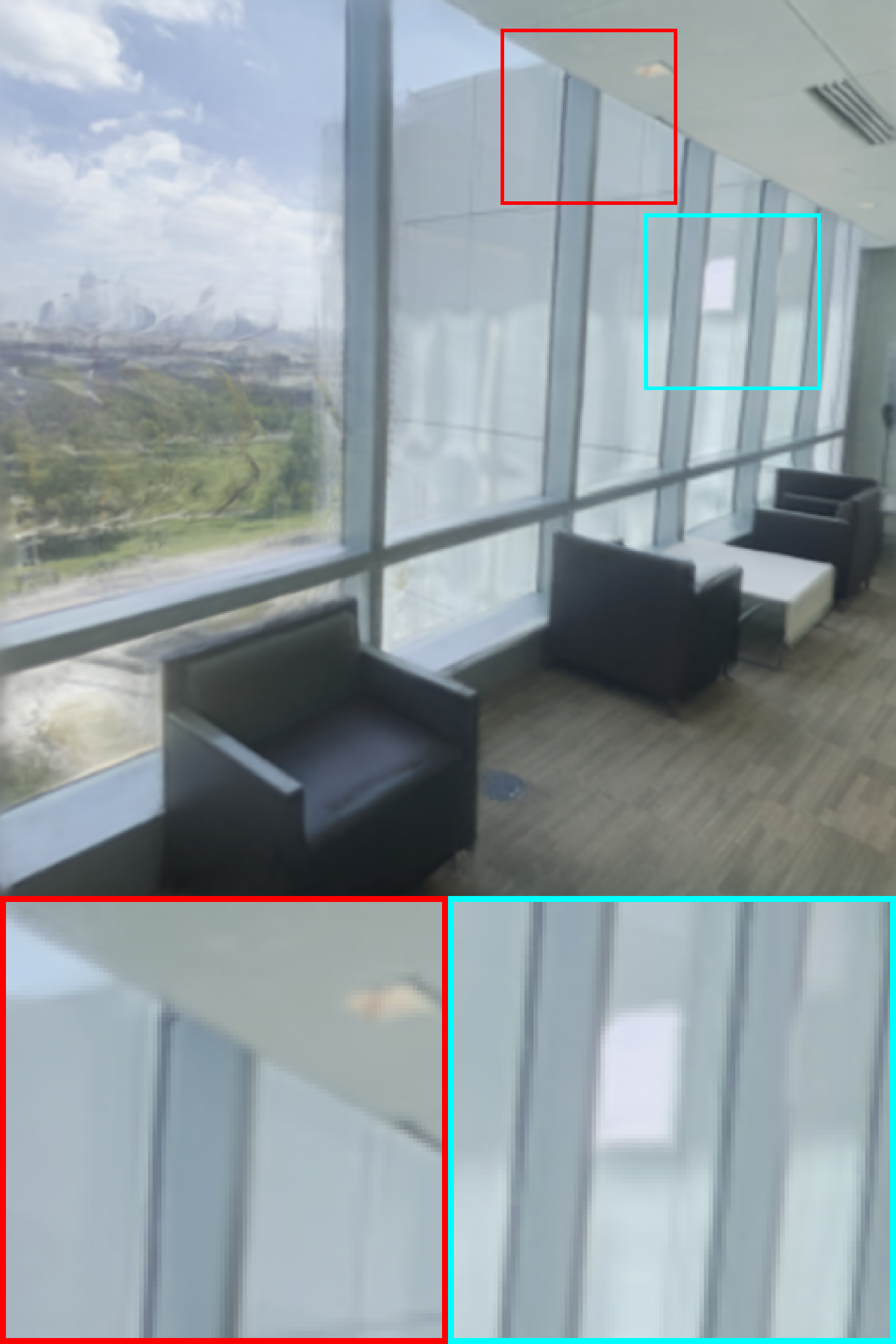} &
        \includegraphics[width=\imgsize]{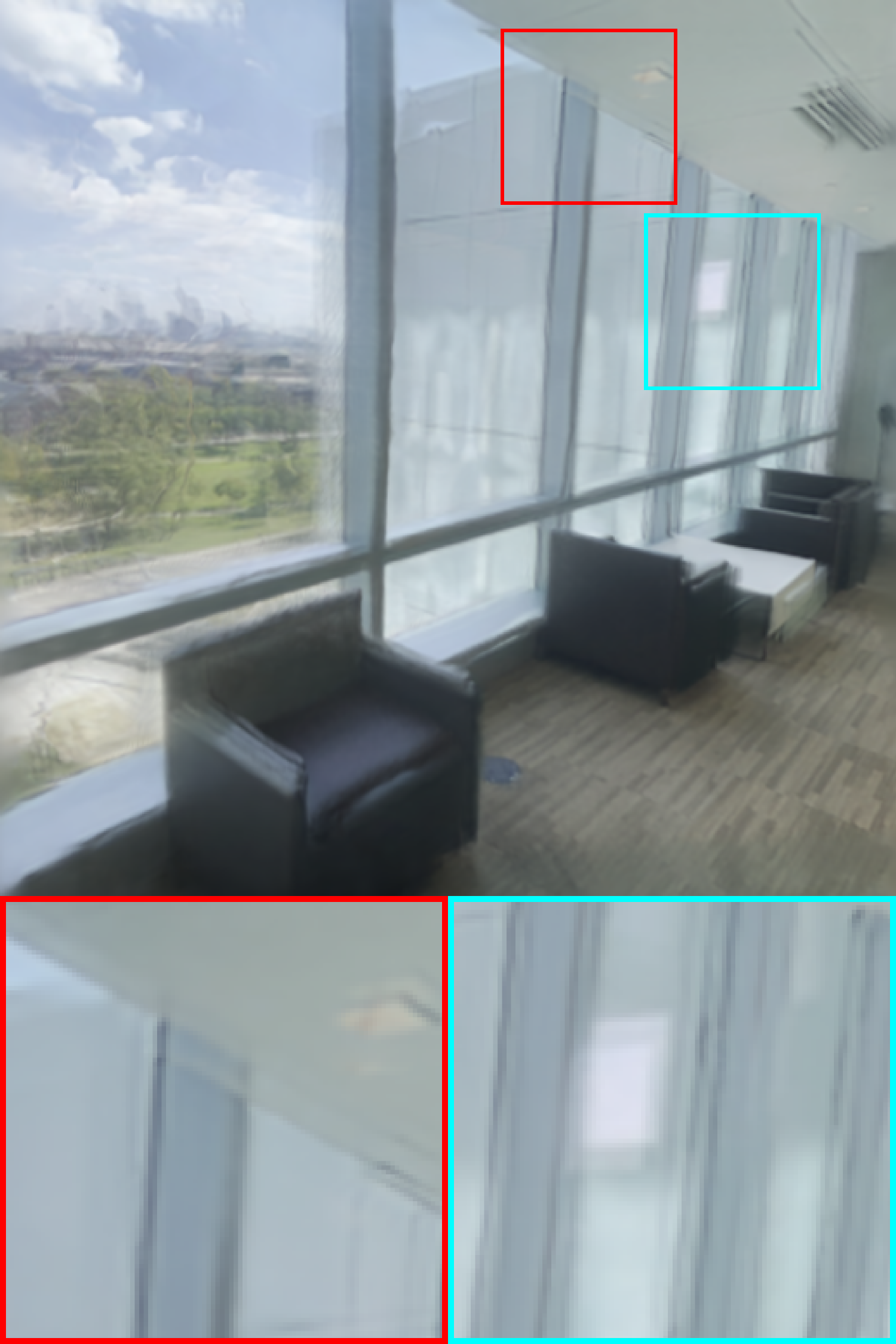} &
        \includegraphics[width=\imgsize]{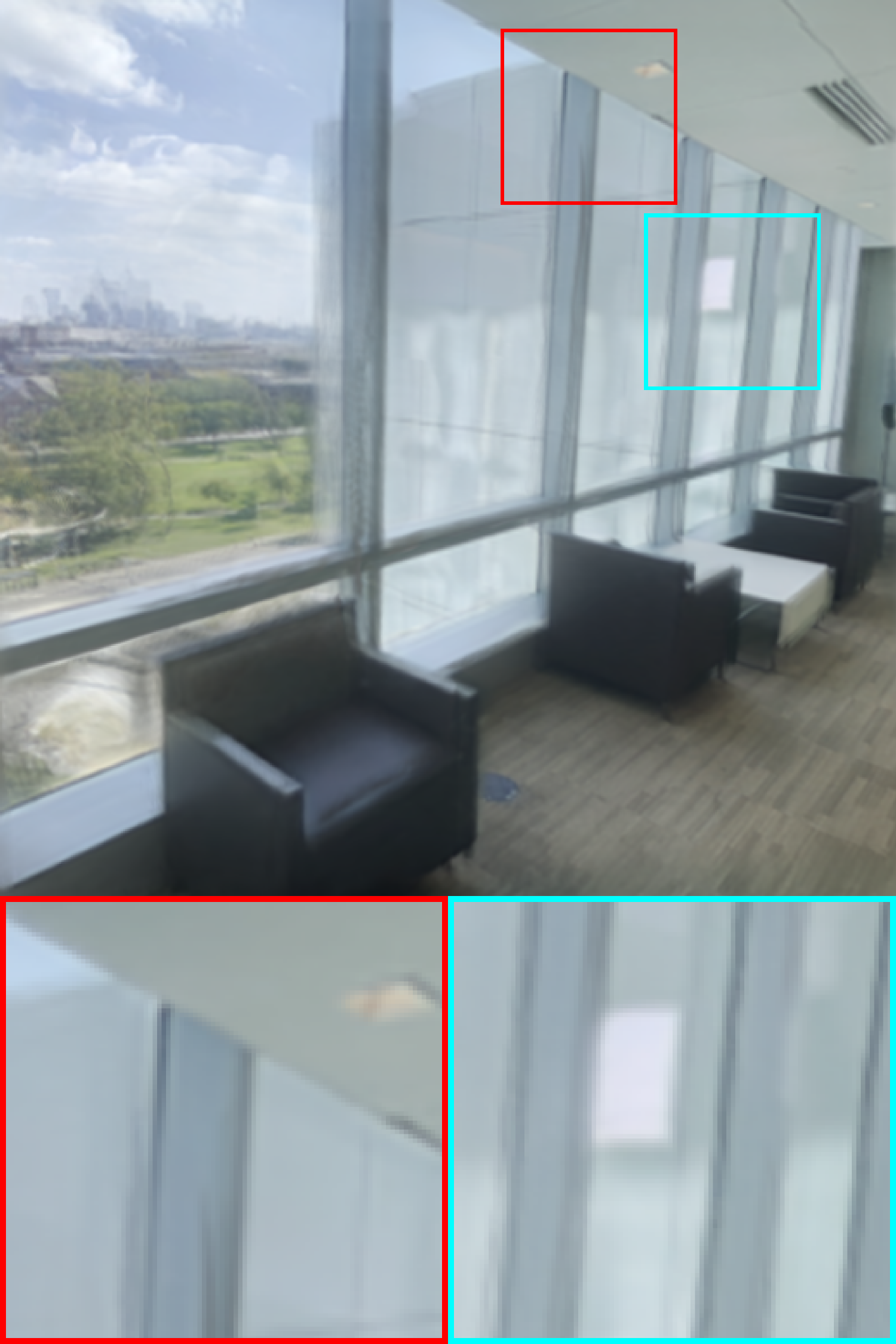} &
        \includegraphics[width=\imgsize]{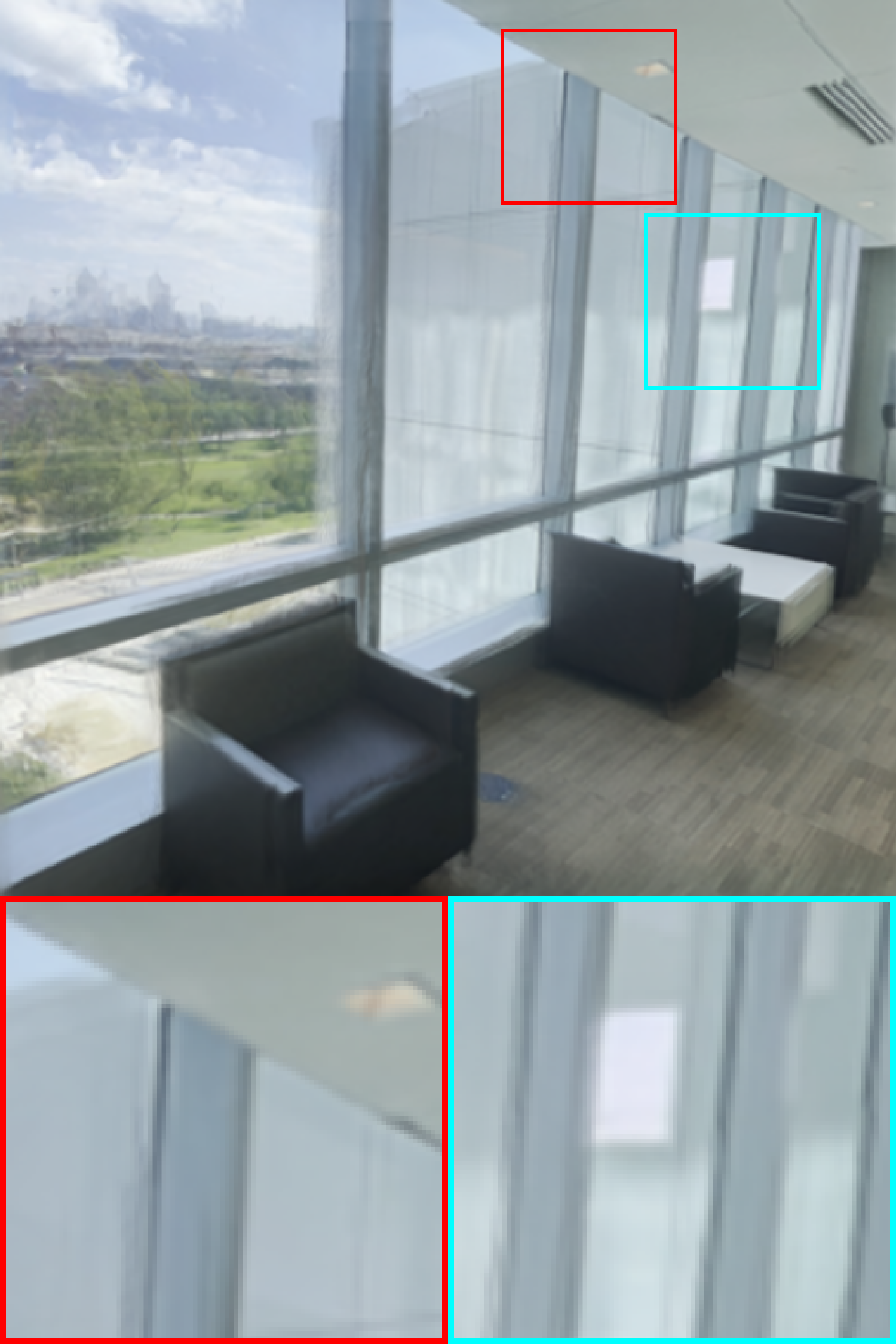} \\
        
        \scriptsize GT & 
        \scriptsize Ours (Full) & 
        \scriptsize w/o DIR-Head &
        \scriptsize w/o DPR-Offsets &
        \scriptsize w/o NV-Sup 
         \\
    \end{NiceTabular}
    
    \caption{Qualitative ablation study on \dlbench datasets under 5 input views.}

    \label{fig:fig_abl_dl3dv}
\end{figure*}

\begin{figure*}[t!]
    \makeatletter\setlength{\@fptop}{0pt}\makeatother
    \centering
    \newcommand{\imgsize}{0.19\textwidth} 
    \setlength{\tabcolsep}{1pt} 
    \renewcommand{\arraystretch}{0.5} 
    
    \begin{NiceTabular}{cccccc}
        \includegraphics[width=\imgsize]{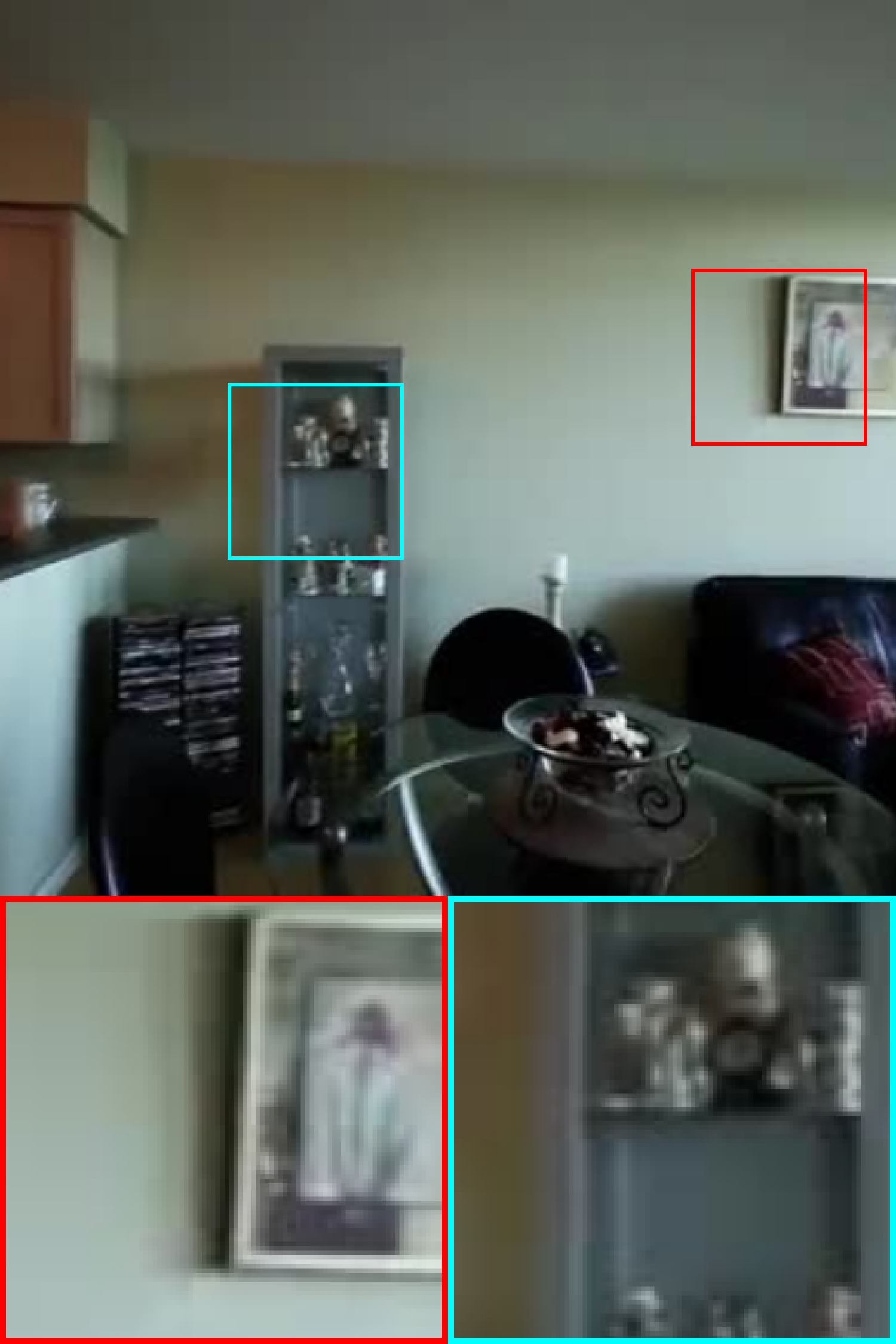} &
        \includegraphics[width=\imgsize]{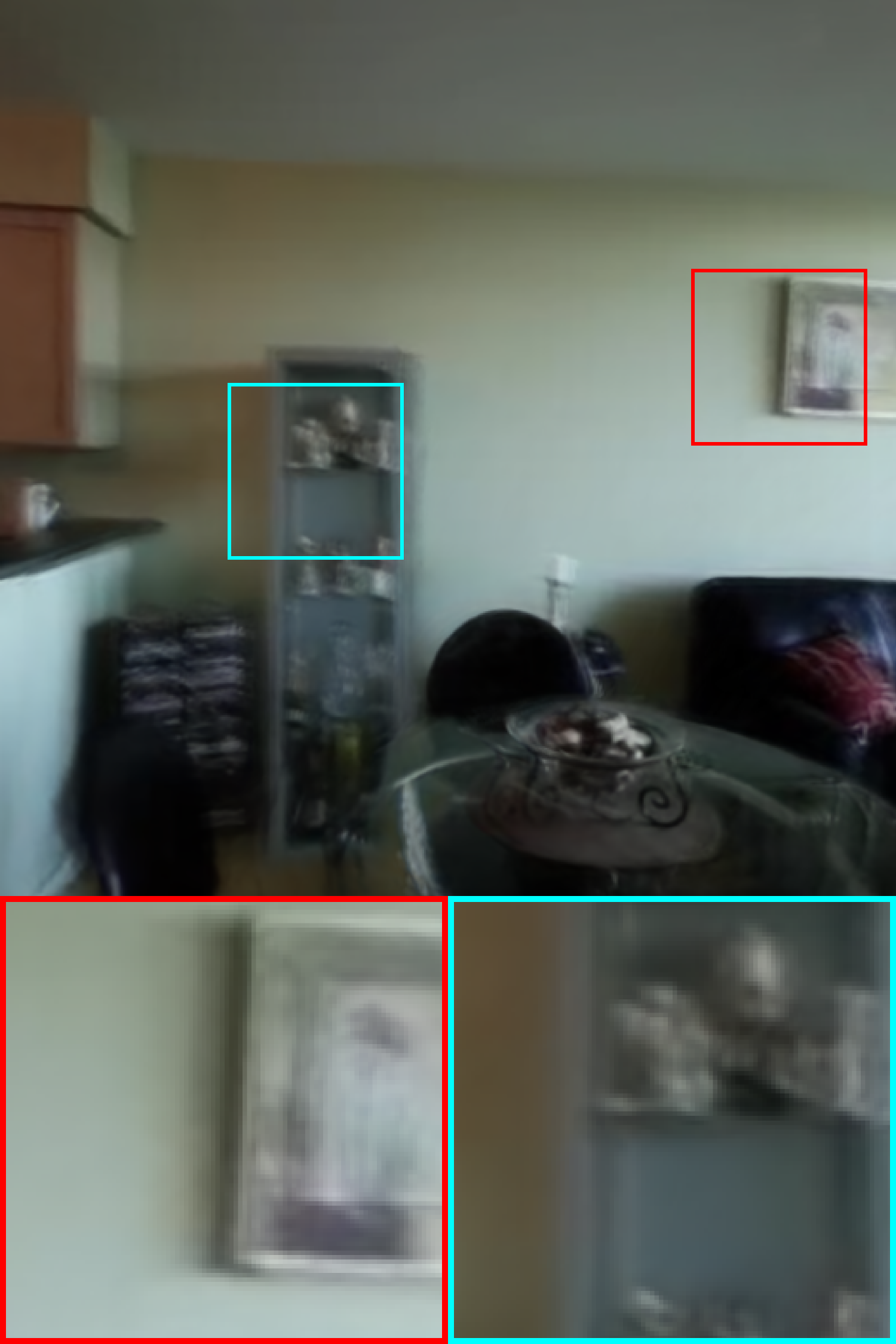} &
        \includegraphics[width=\imgsize]{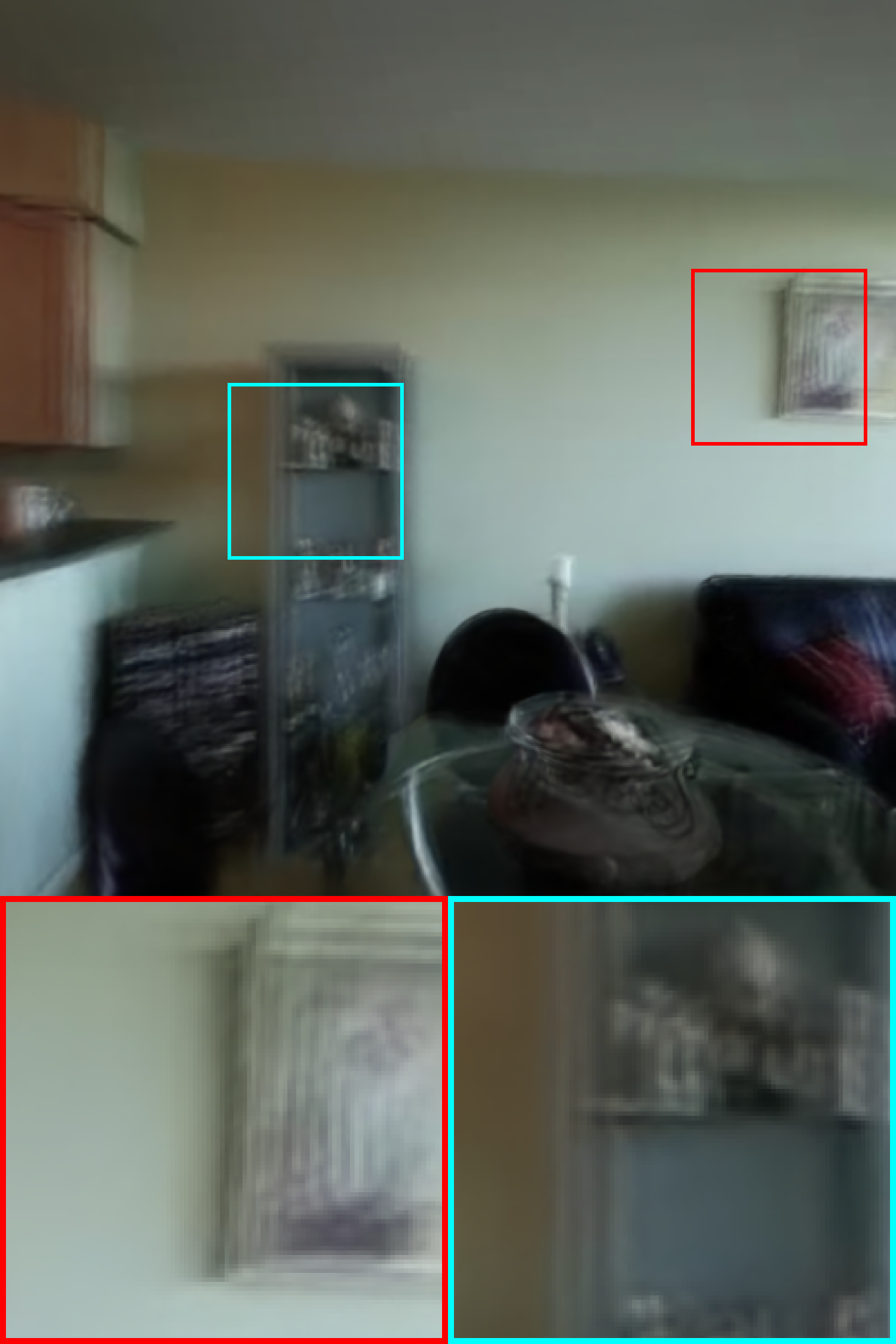} &
        \includegraphics[width=\imgsize]{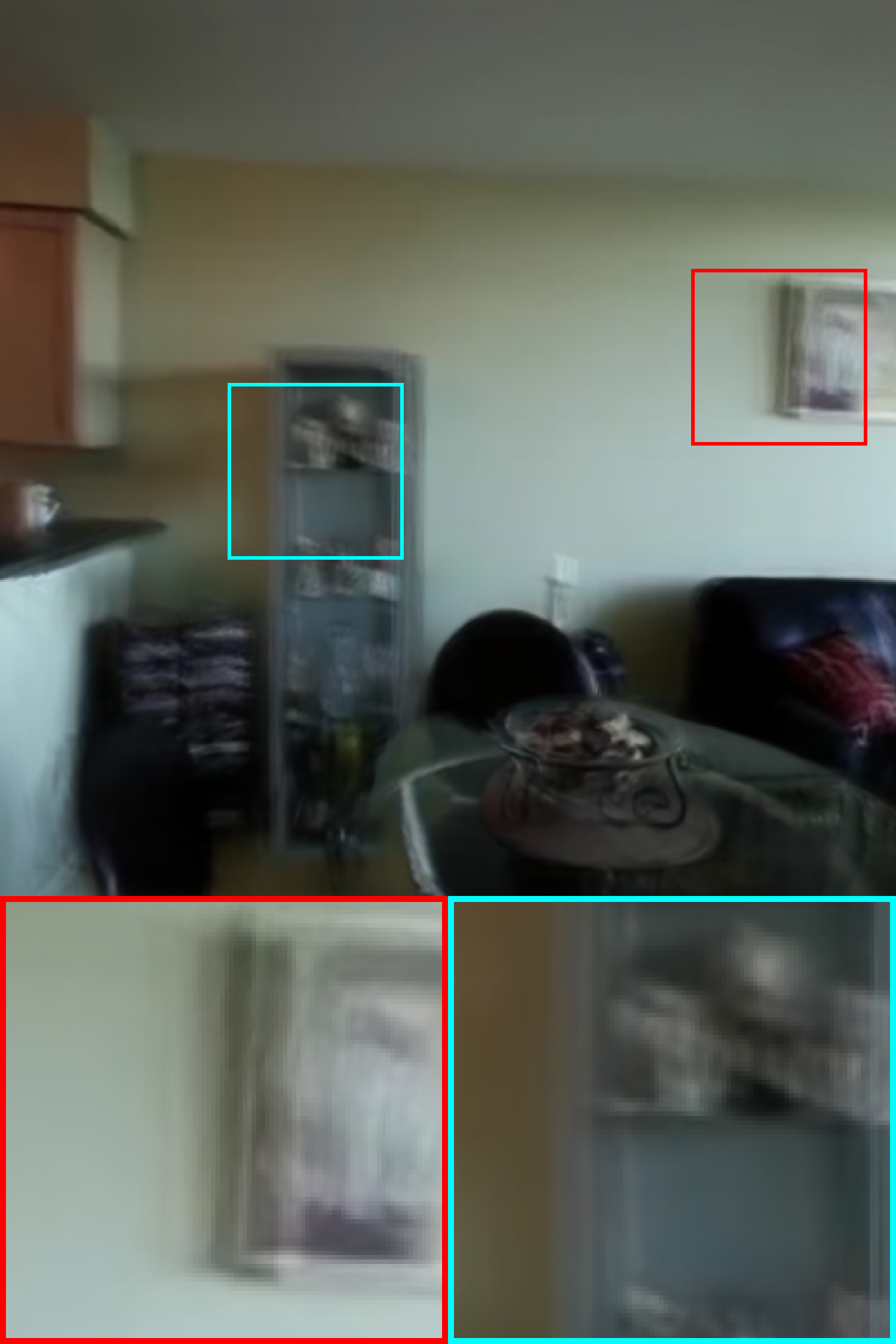} &
        \includegraphics[width=\imgsize]{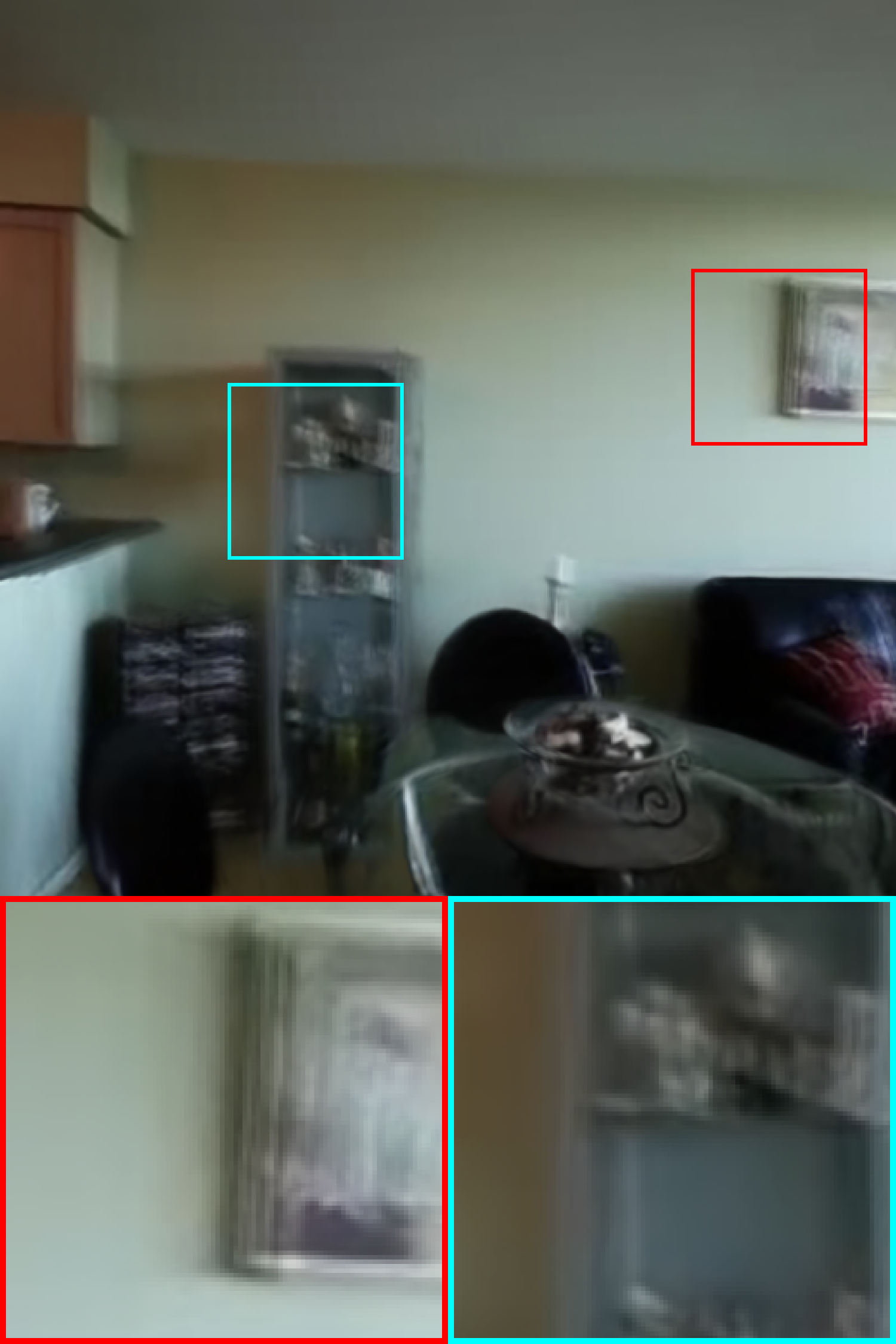} \\

        \includegraphics[width=\imgsize]{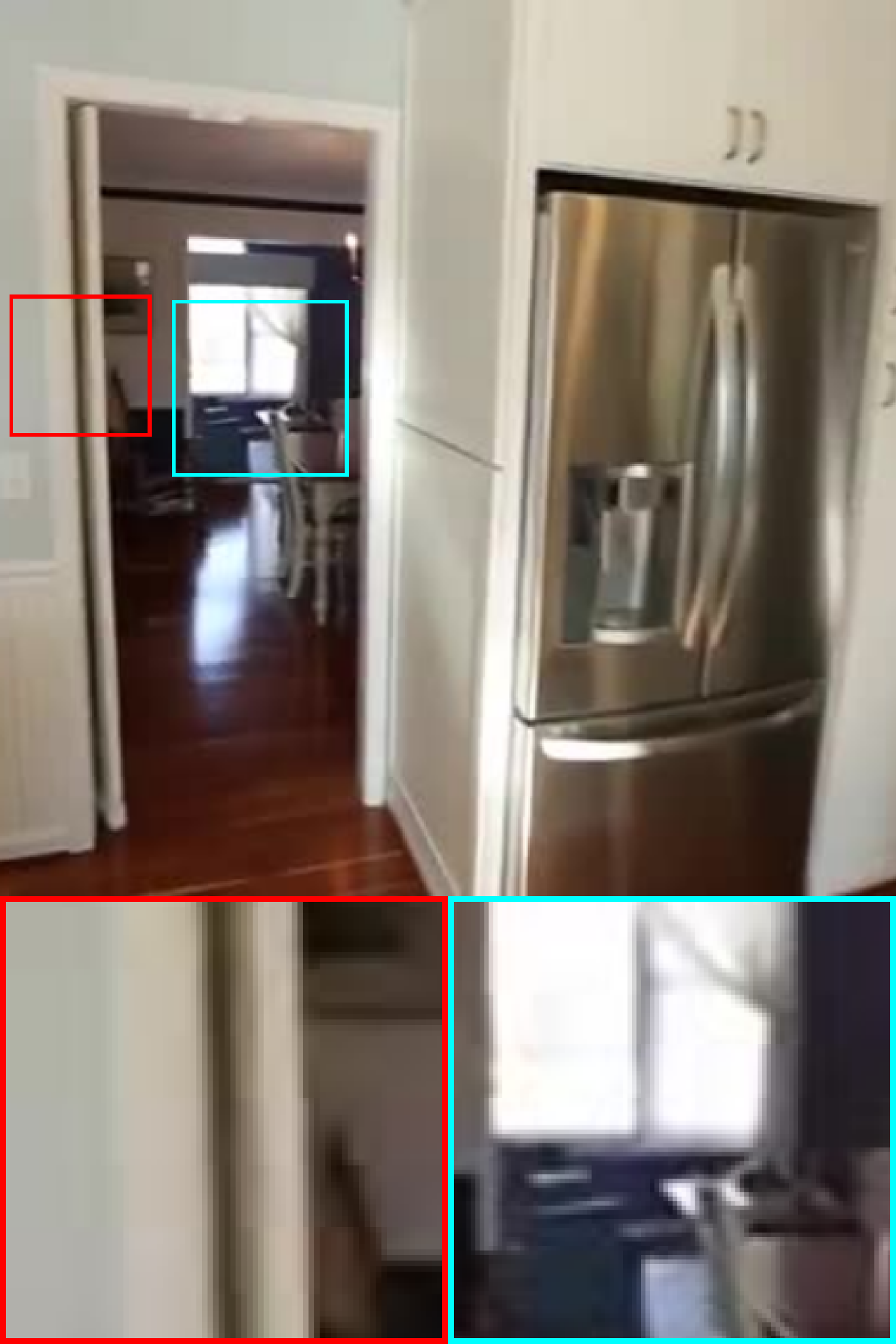} &
        \includegraphics[width=\imgsize]{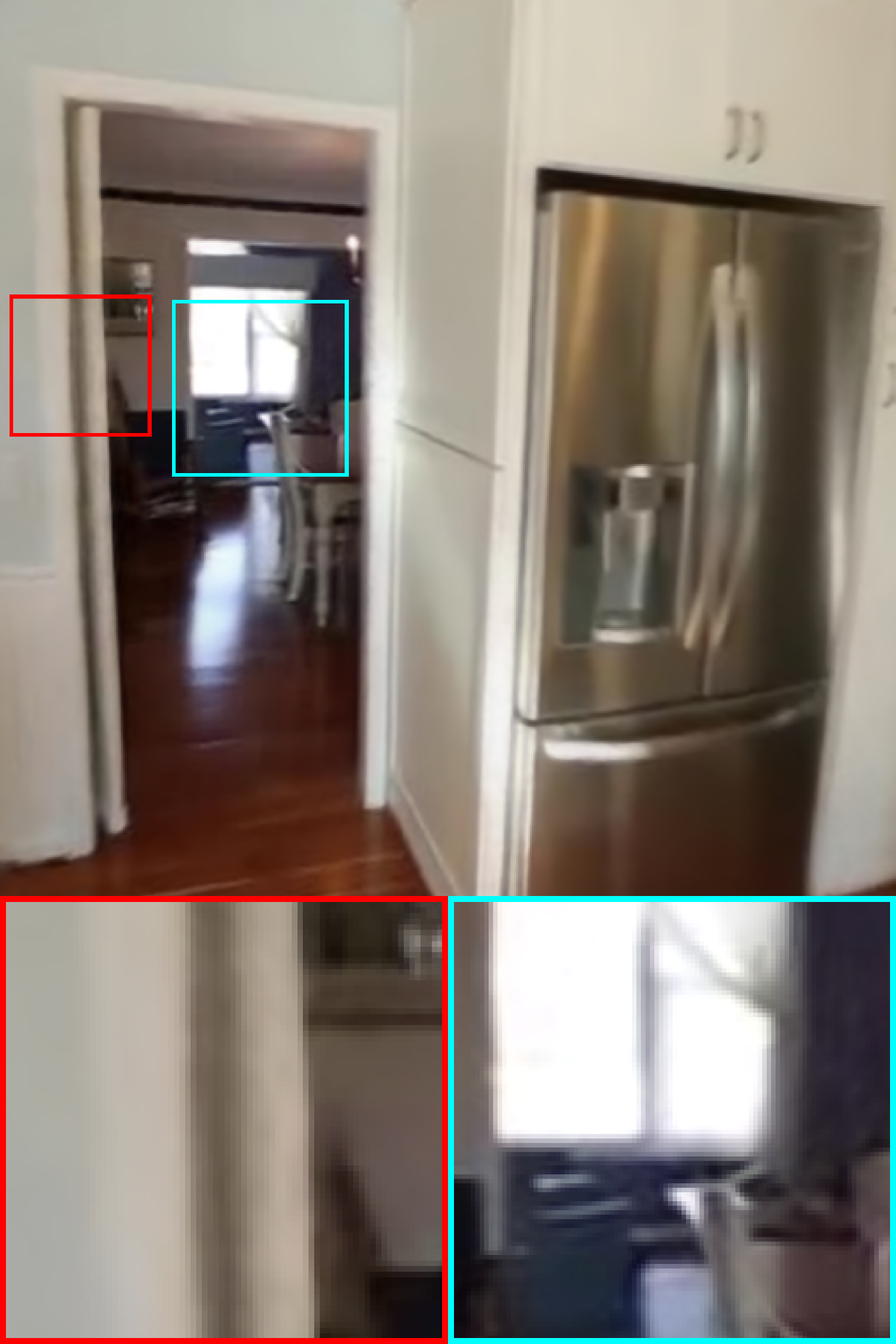} &
        \includegraphics[width=\imgsize]{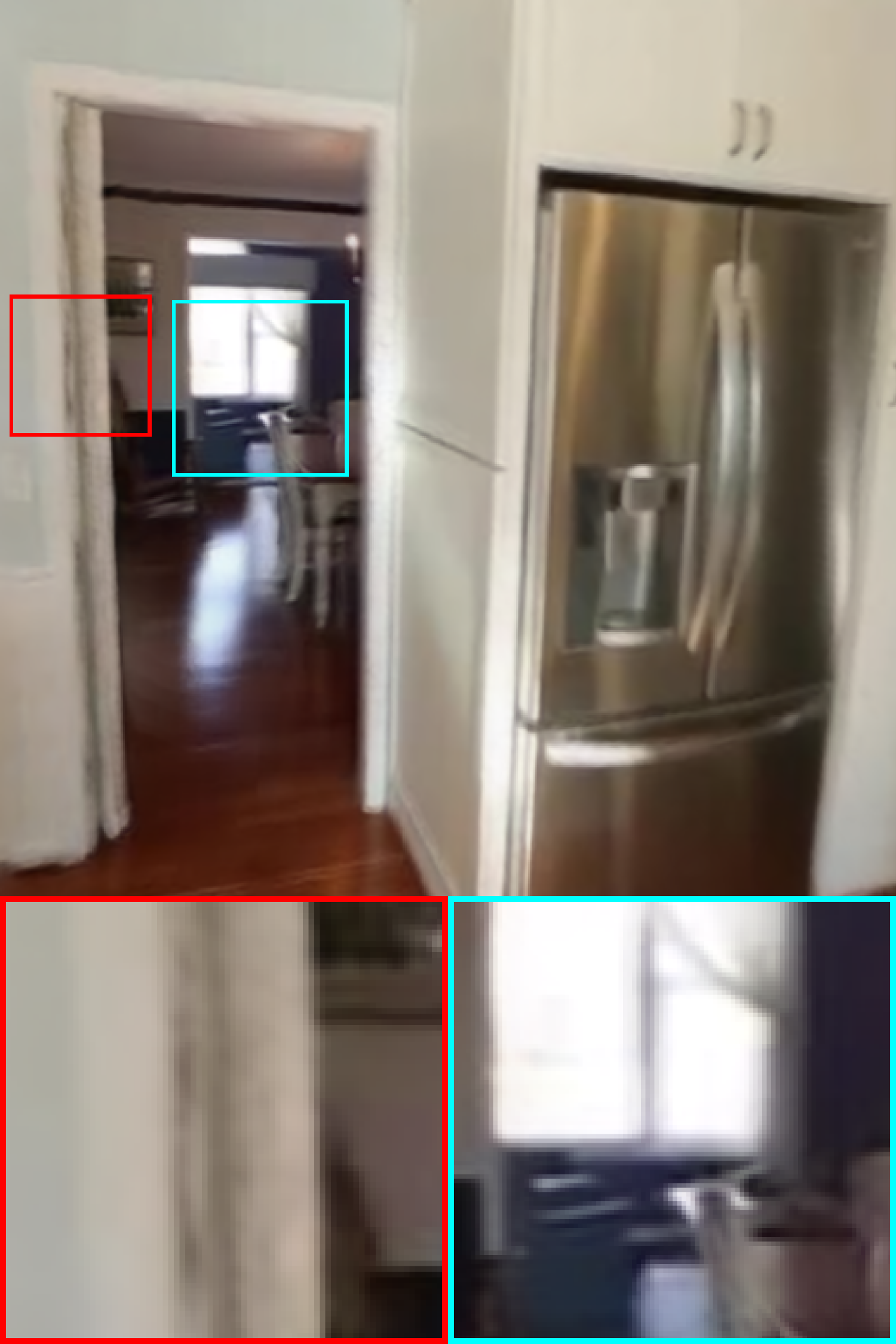} &
        \includegraphics[width=\imgsize]{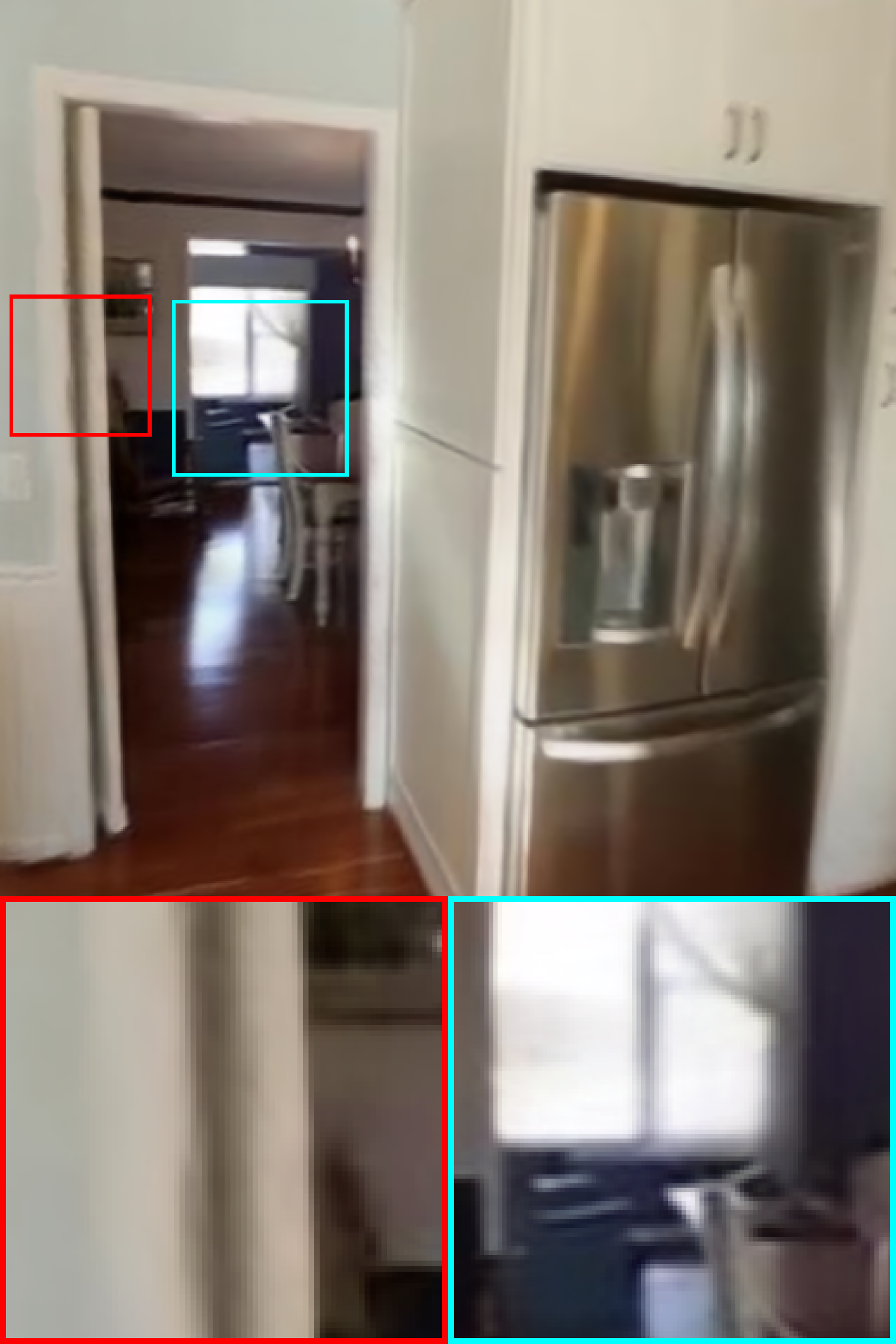} &
        \includegraphics[width=\imgsize]{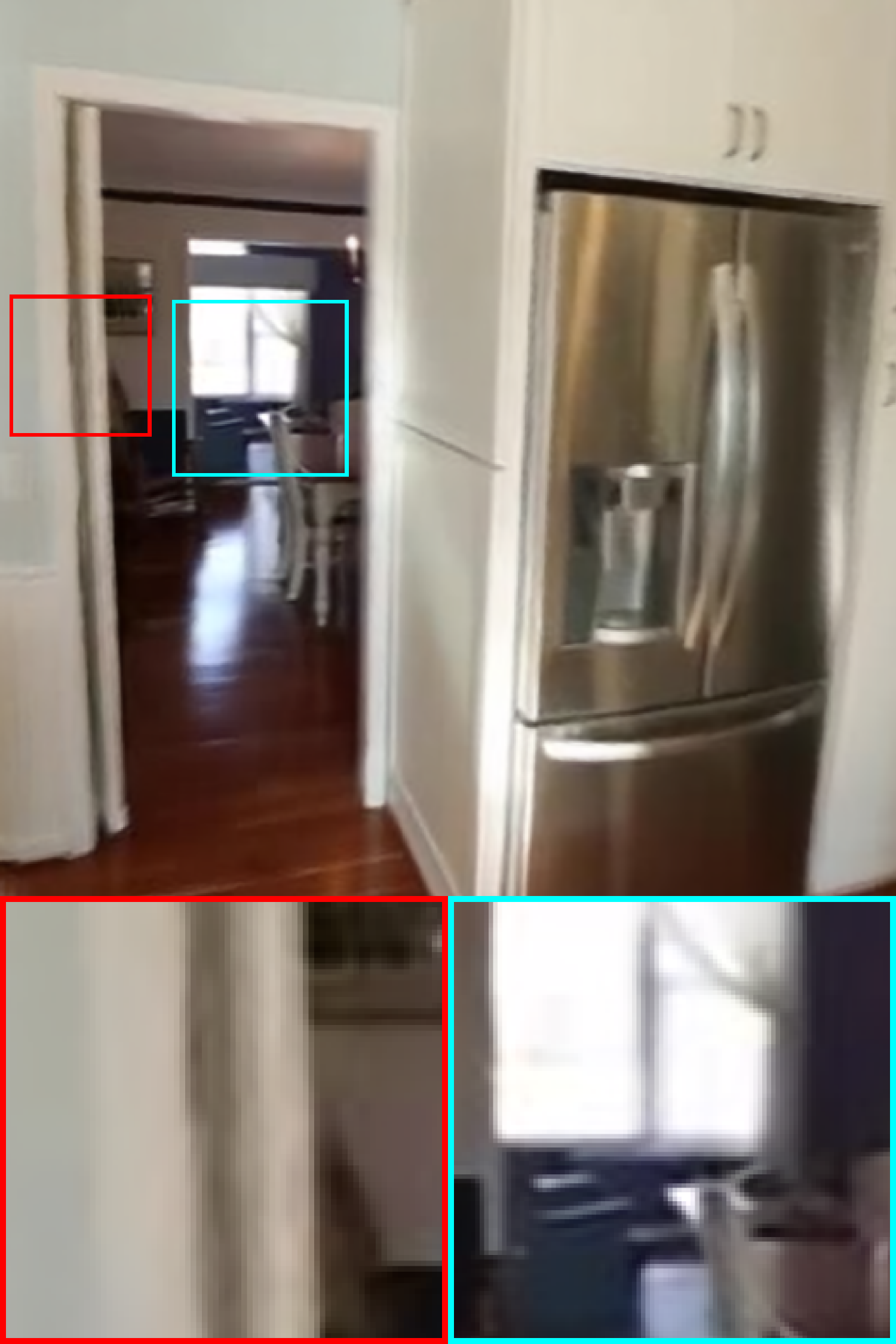} \\

        \includegraphics[width=\imgsize]{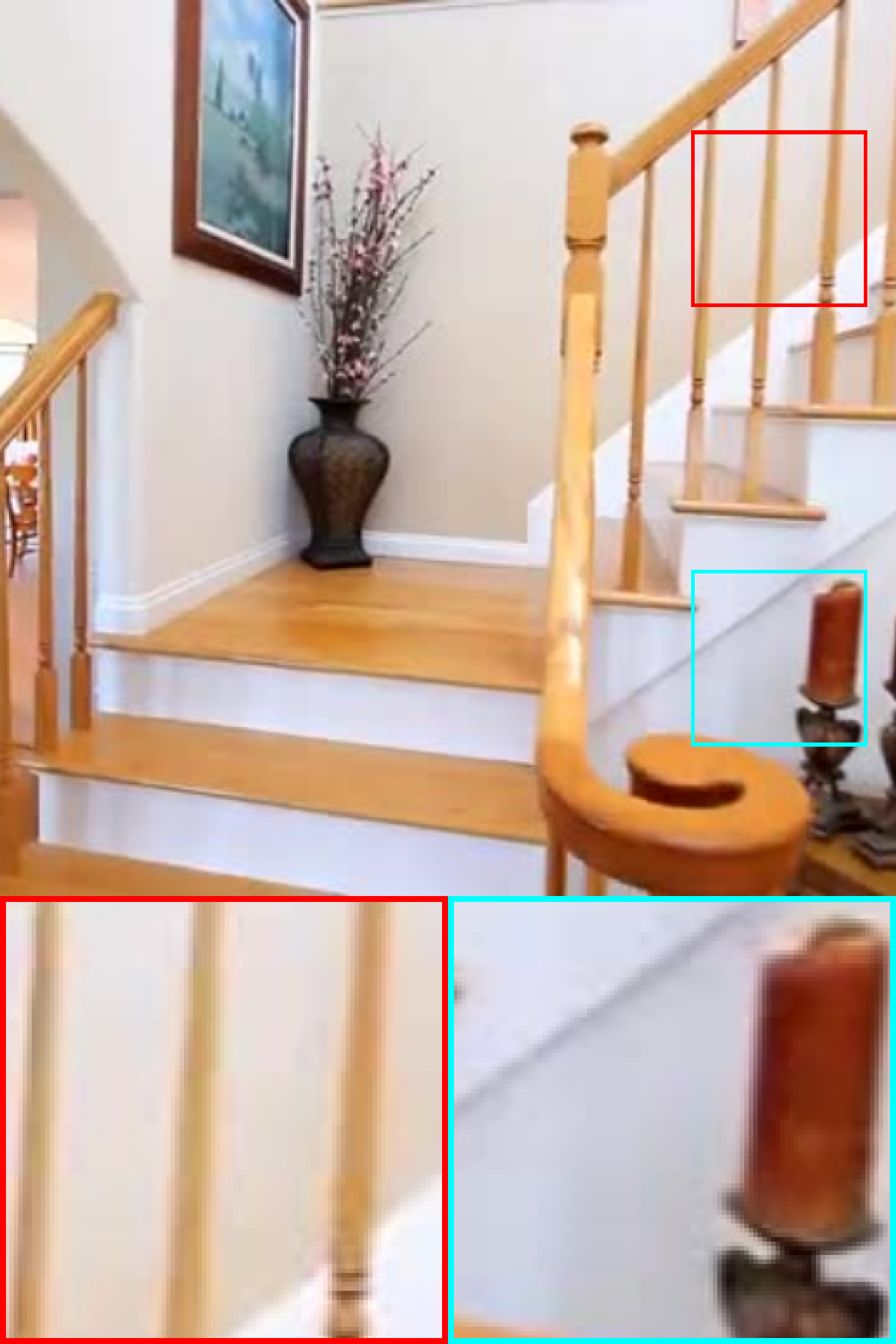} &
        \includegraphics[width=\imgsize]{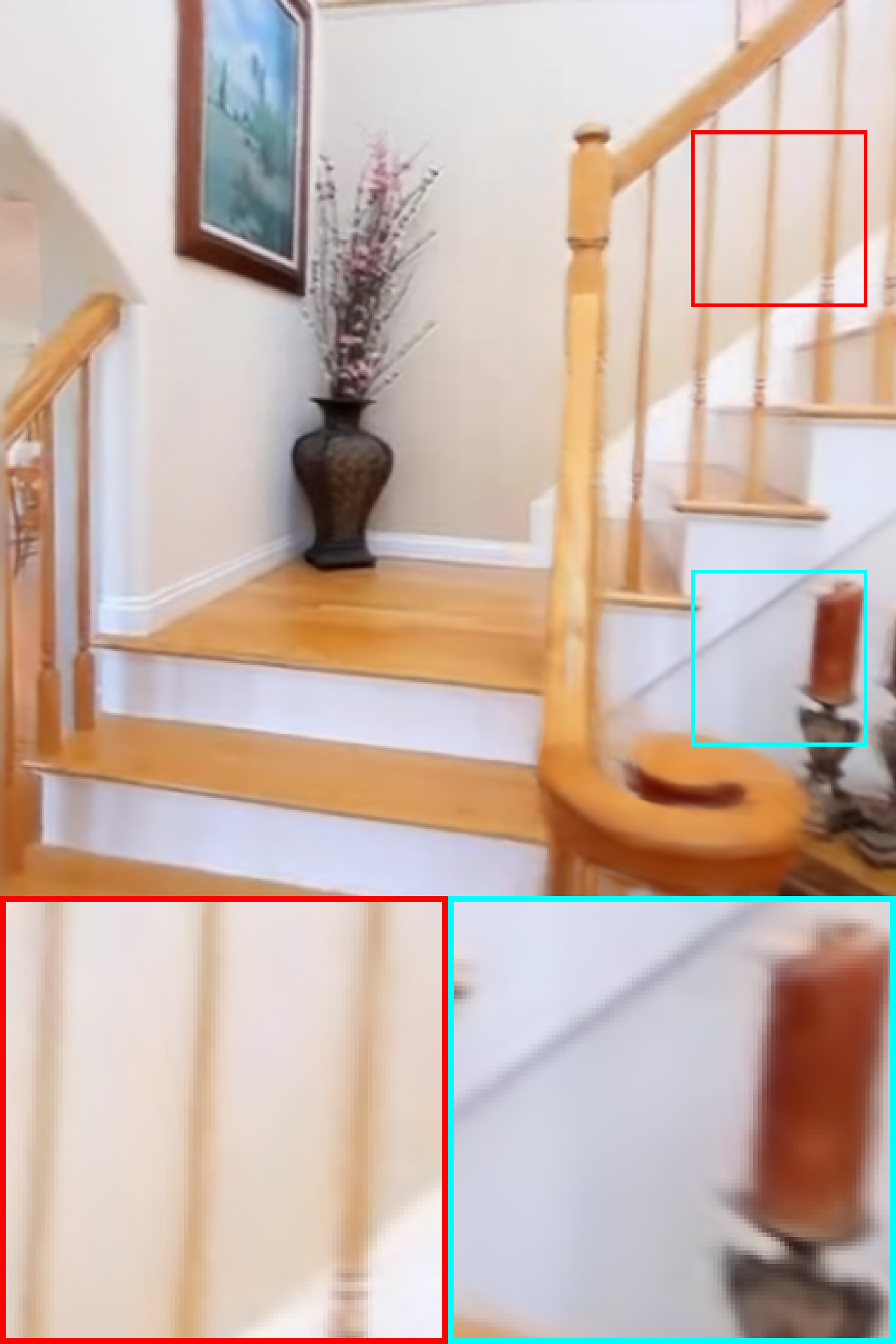} &
        \includegraphics[width=\imgsize]{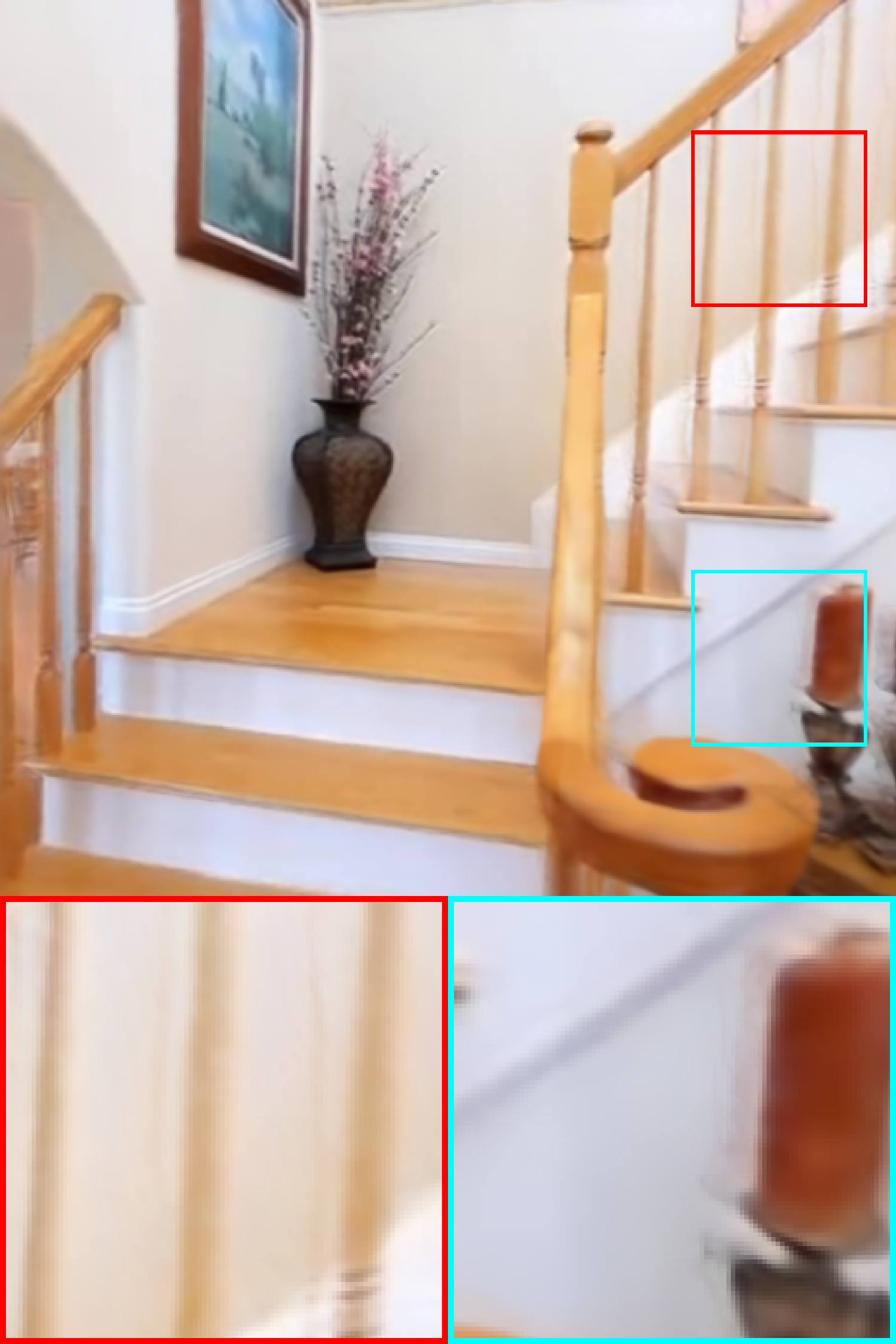} &
        \includegraphics[width=\imgsize]{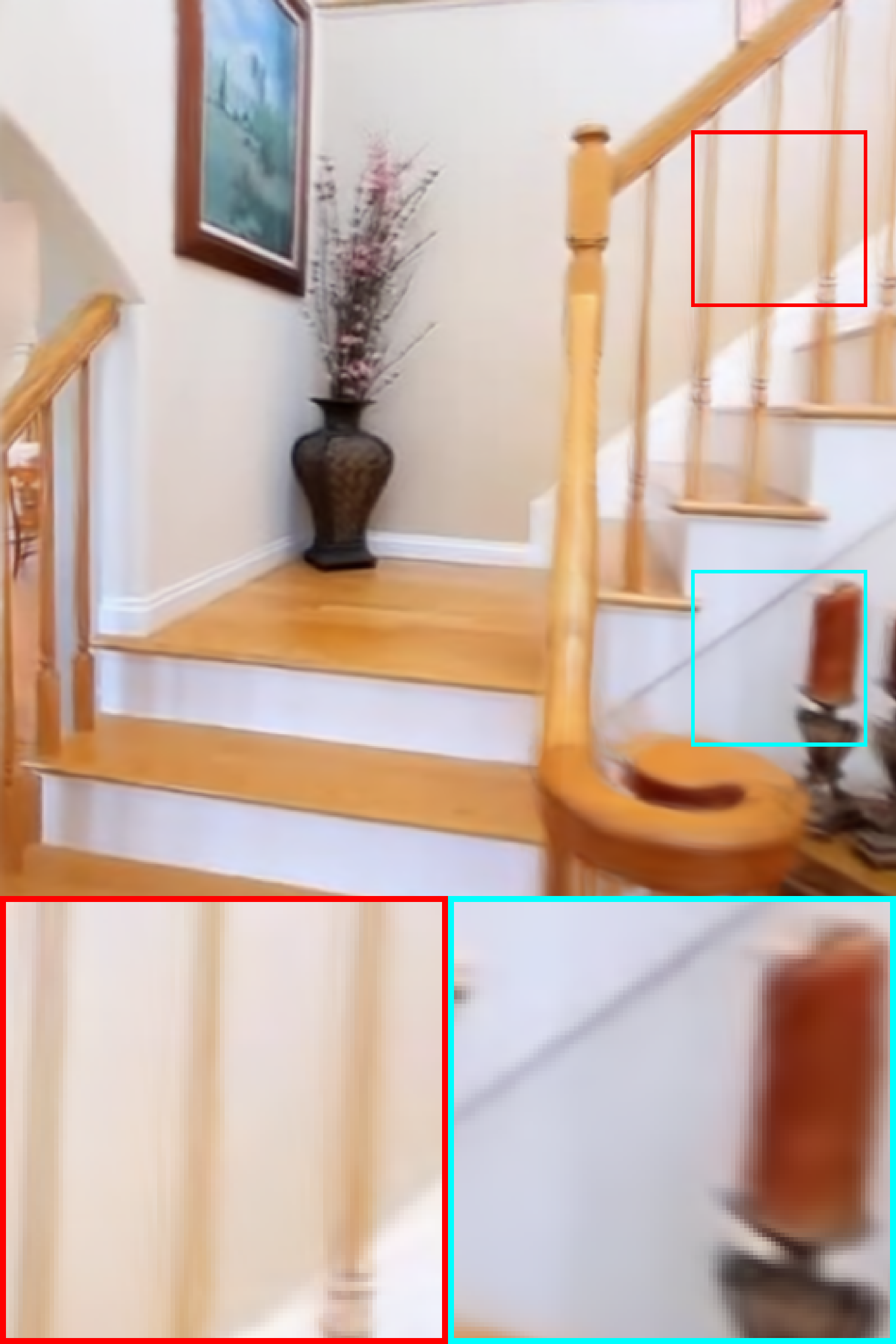} &
        \includegraphics[width=\imgsize]{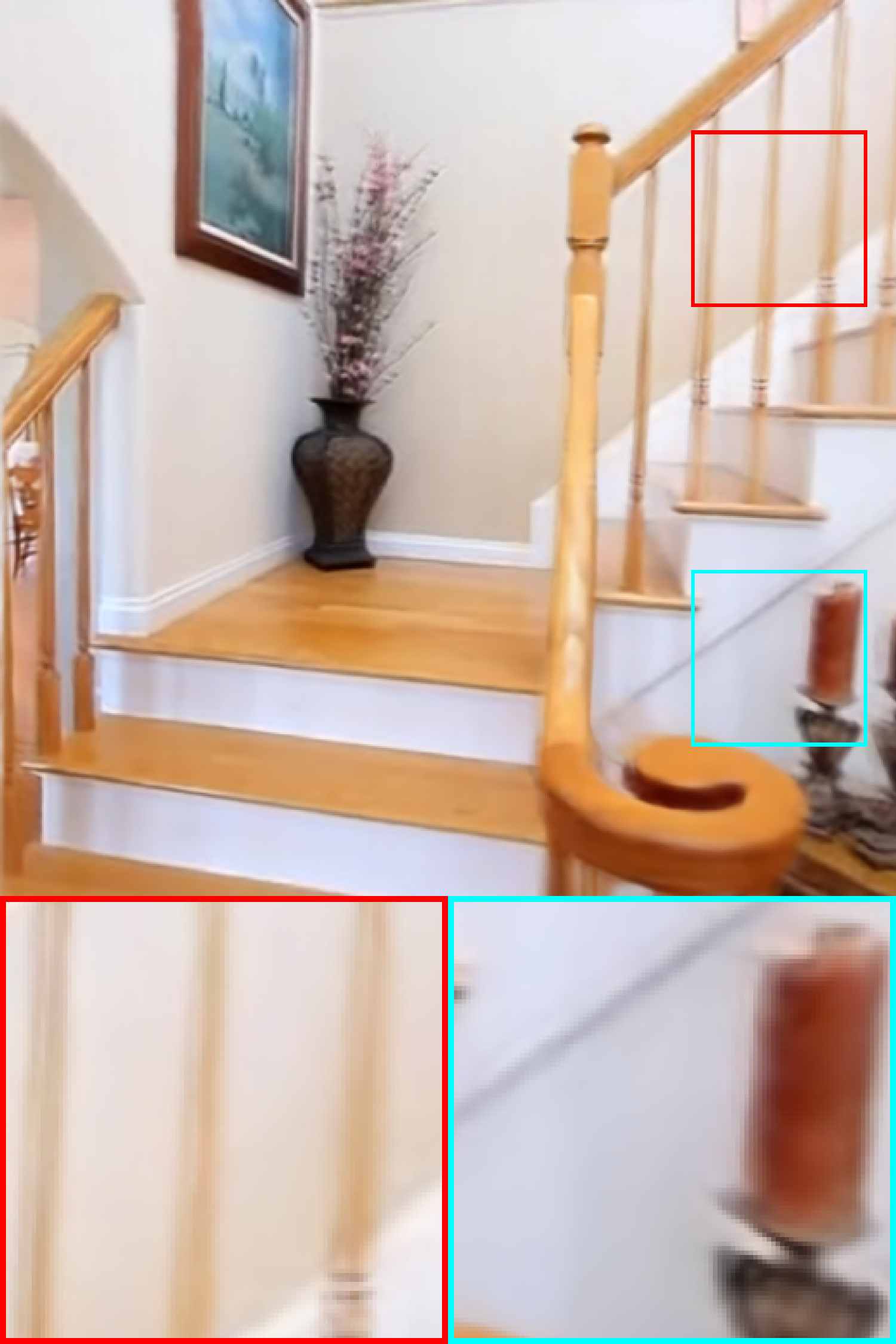} \\
        
        \scriptsize GT & 
        \scriptsize Ours (Full) & 
        \scriptsize w/o DIR-Head &
        \scriptsize w/o DPR-Offsets &
        \scriptsize w/o NV-Sup 
         \\
    \end{NiceTabular}
    
    \caption{Qualitative ablation study on \re datasets under 5 input views.}

    \label{fig:fig_abl_re10k}
\end{figure*}

\begin{figure*}[t!]
    \makeatletter\setlength{\@fptop}{0pt}\makeatother
    \centering
    \newcommand{\imgsize}{0.19\textwidth} 
    \setlength{\tabcolsep}{1pt} 
    \renewcommand{\arraystretch}{0.5} 
    
    \begin{NiceTabular}{ccccc}
        \includegraphics[width=\imgsize]{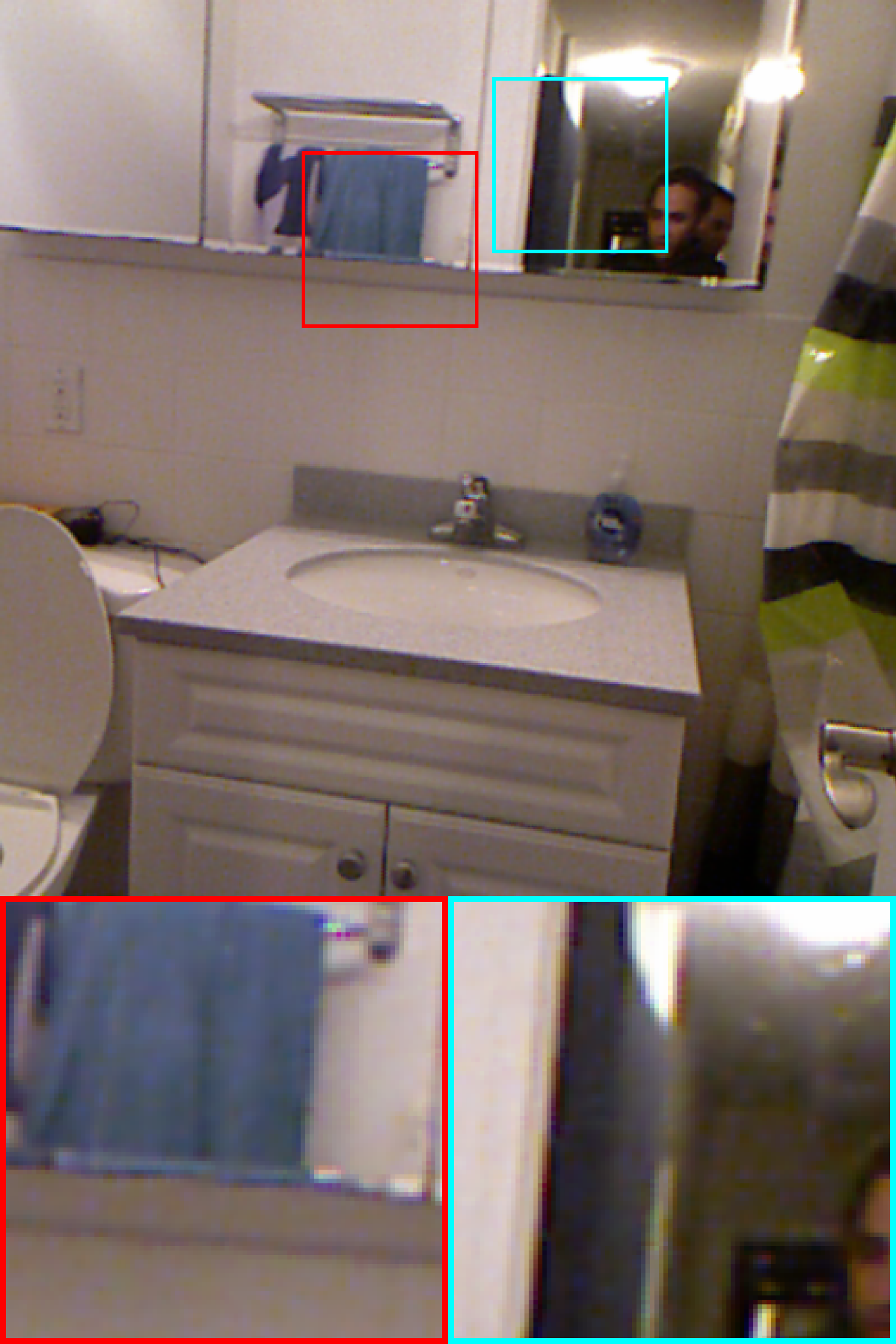} &
        \includegraphics[width=\imgsize]{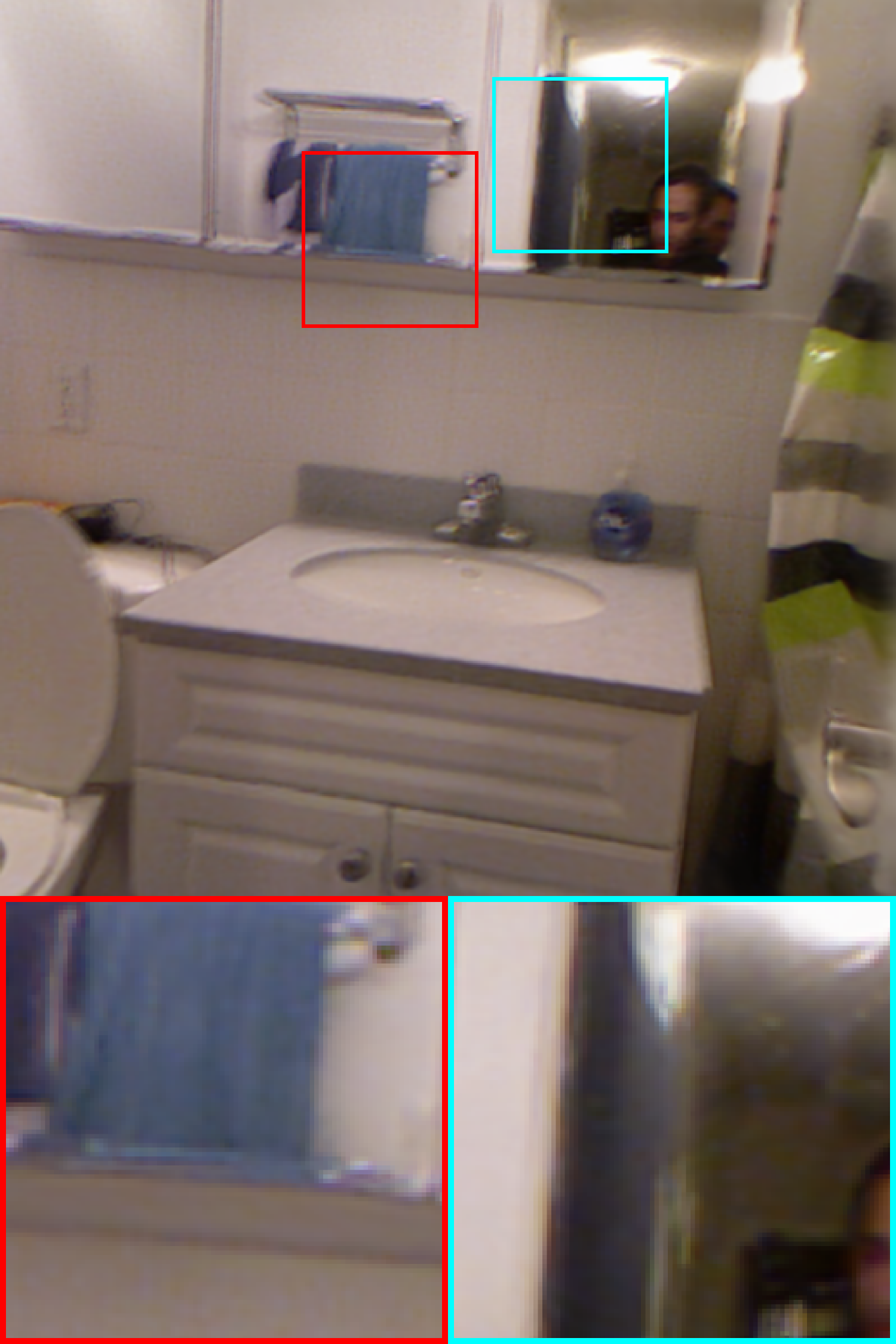} &
        \includegraphics[width=\imgsize]{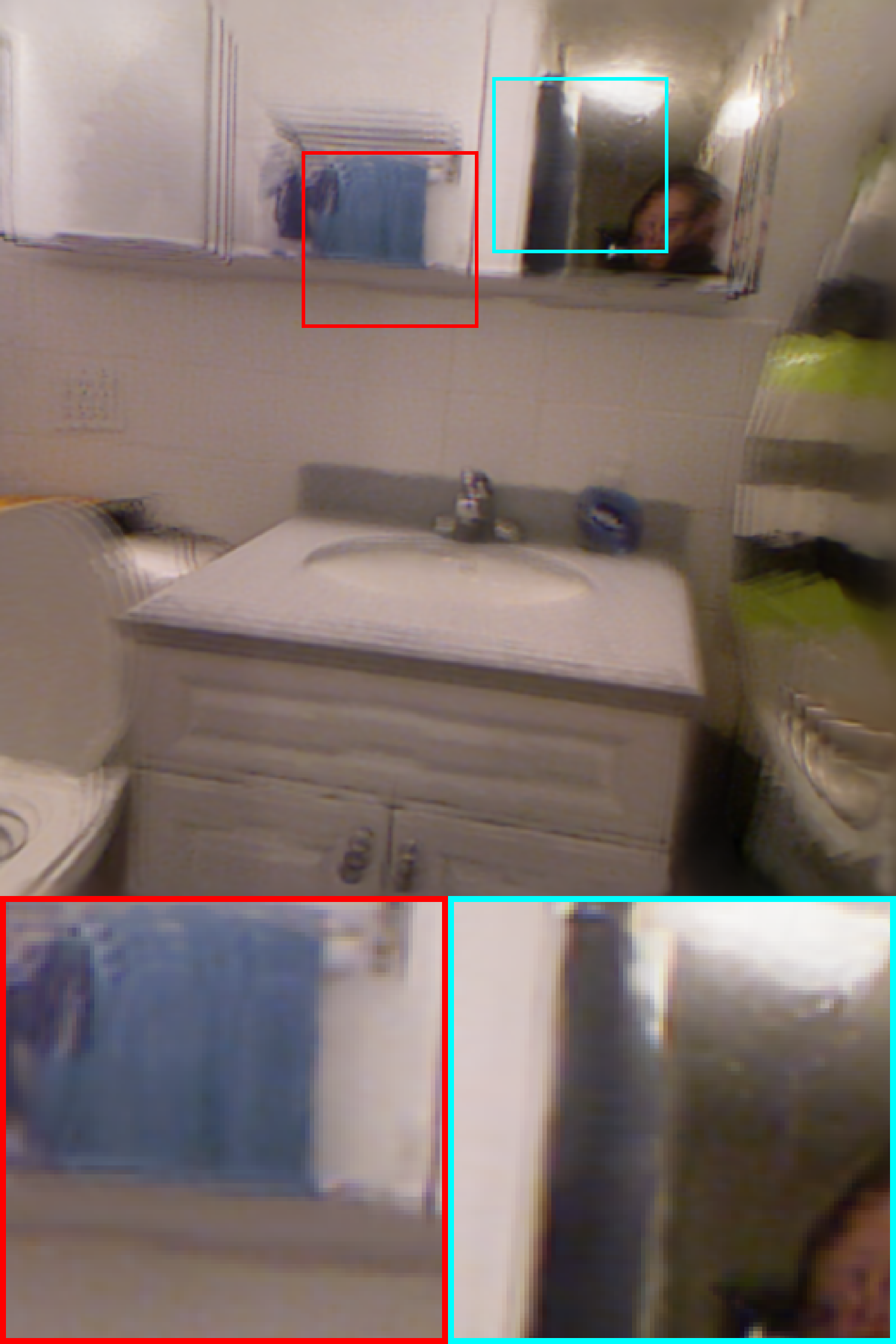} &
        \includegraphics[width=\imgsize]{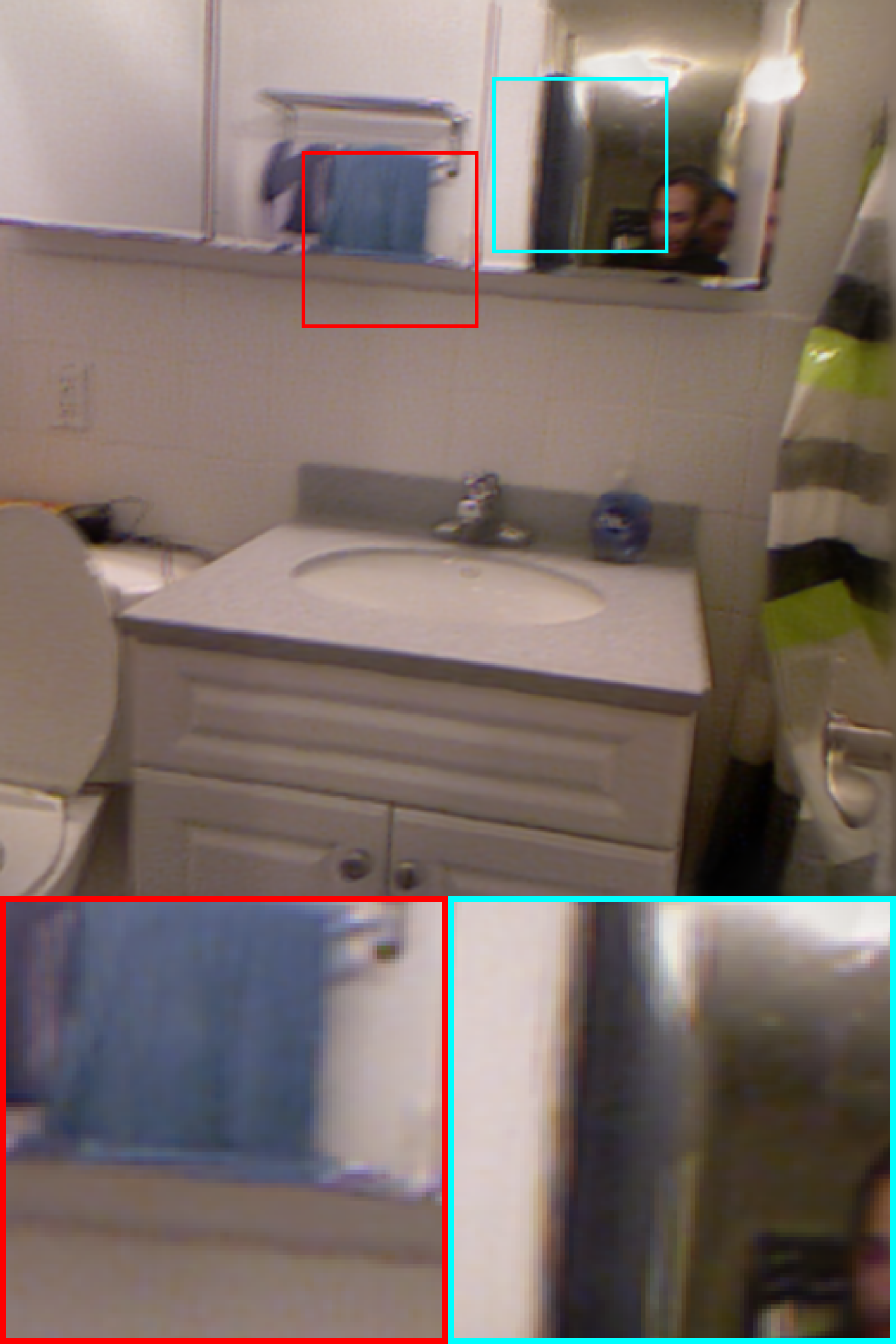} &
        \includegraphics[width=\imgsize]{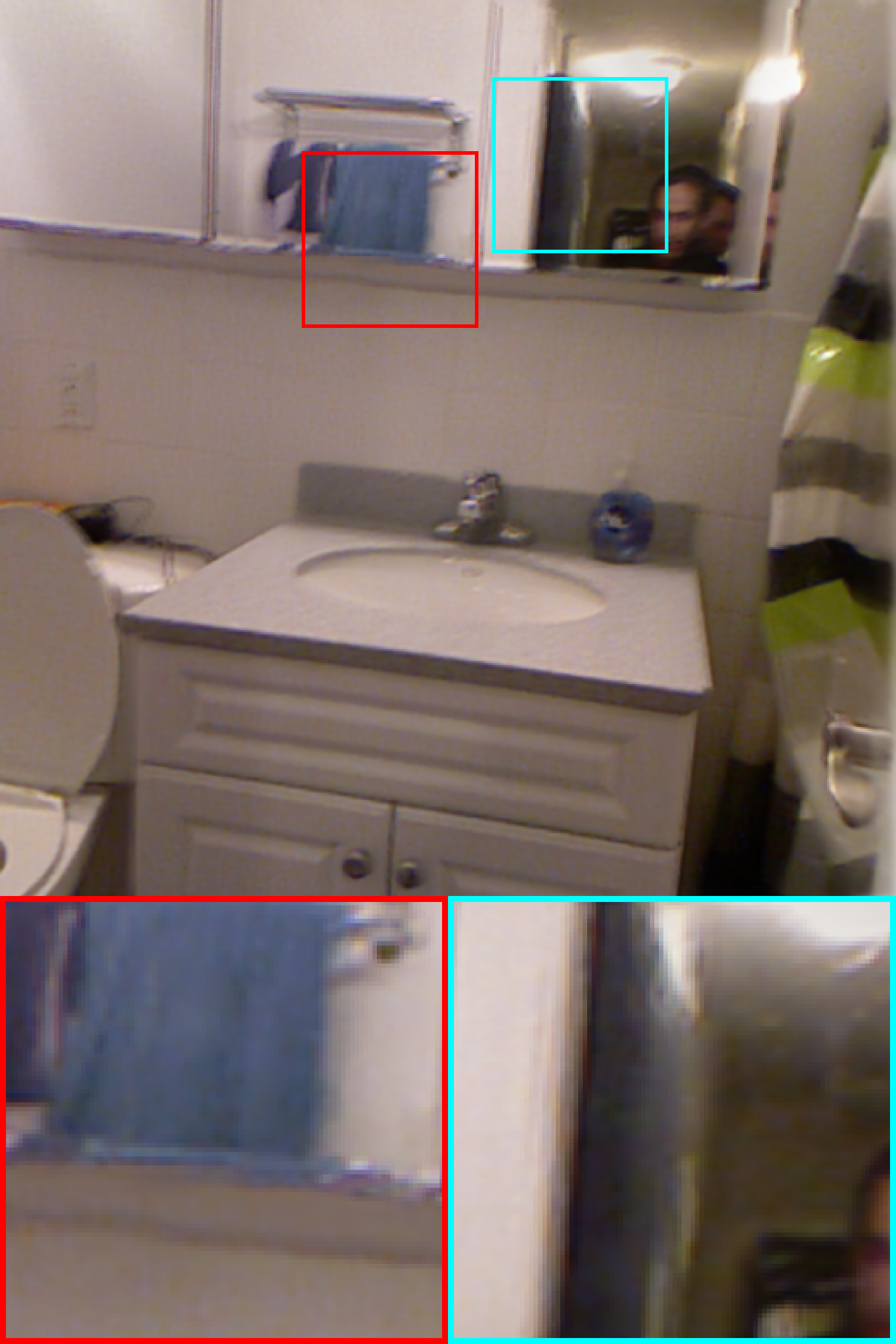} \\
        \includegraphics[width=\imgsize]{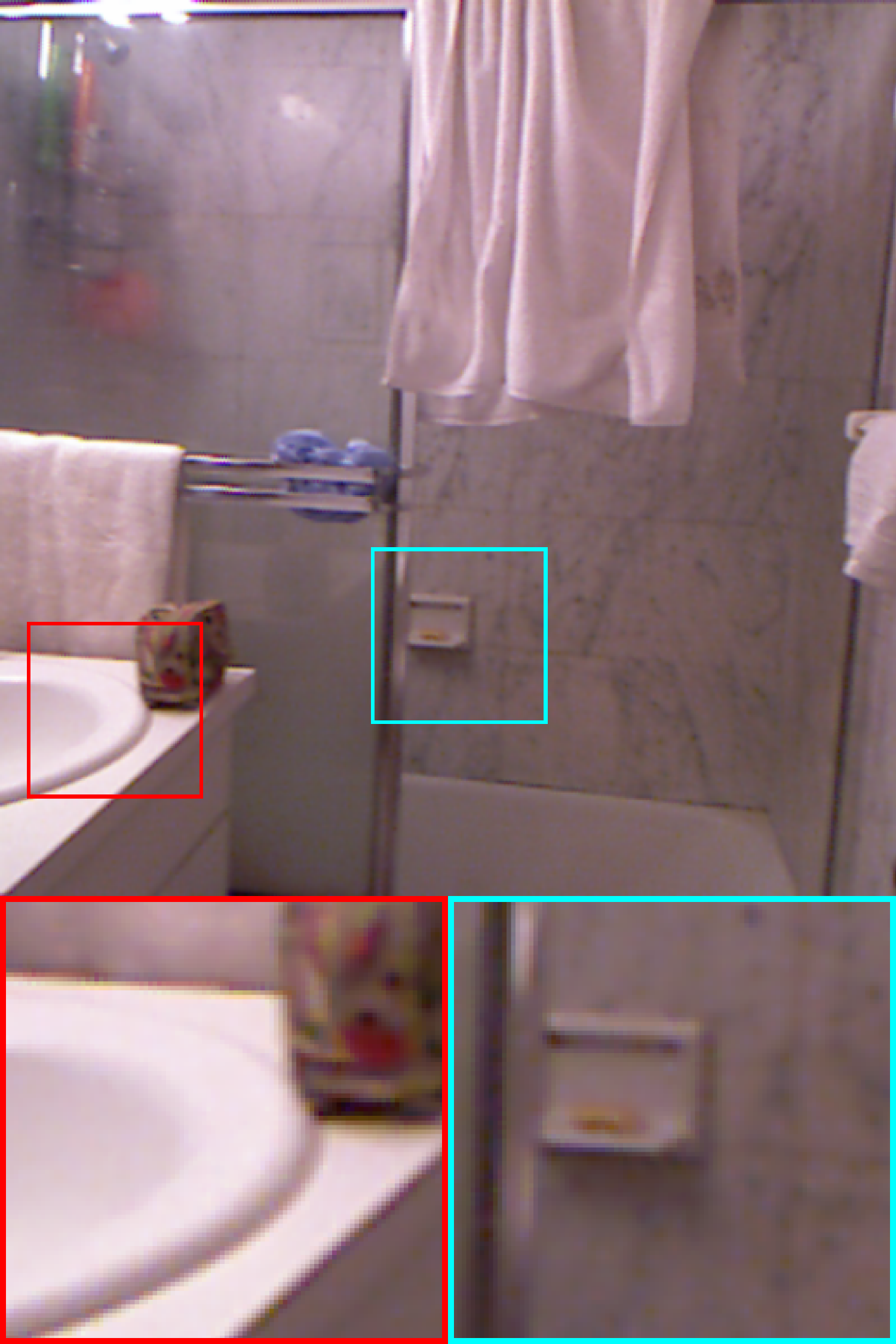} &
        \includegraphics[width=\imgsize]{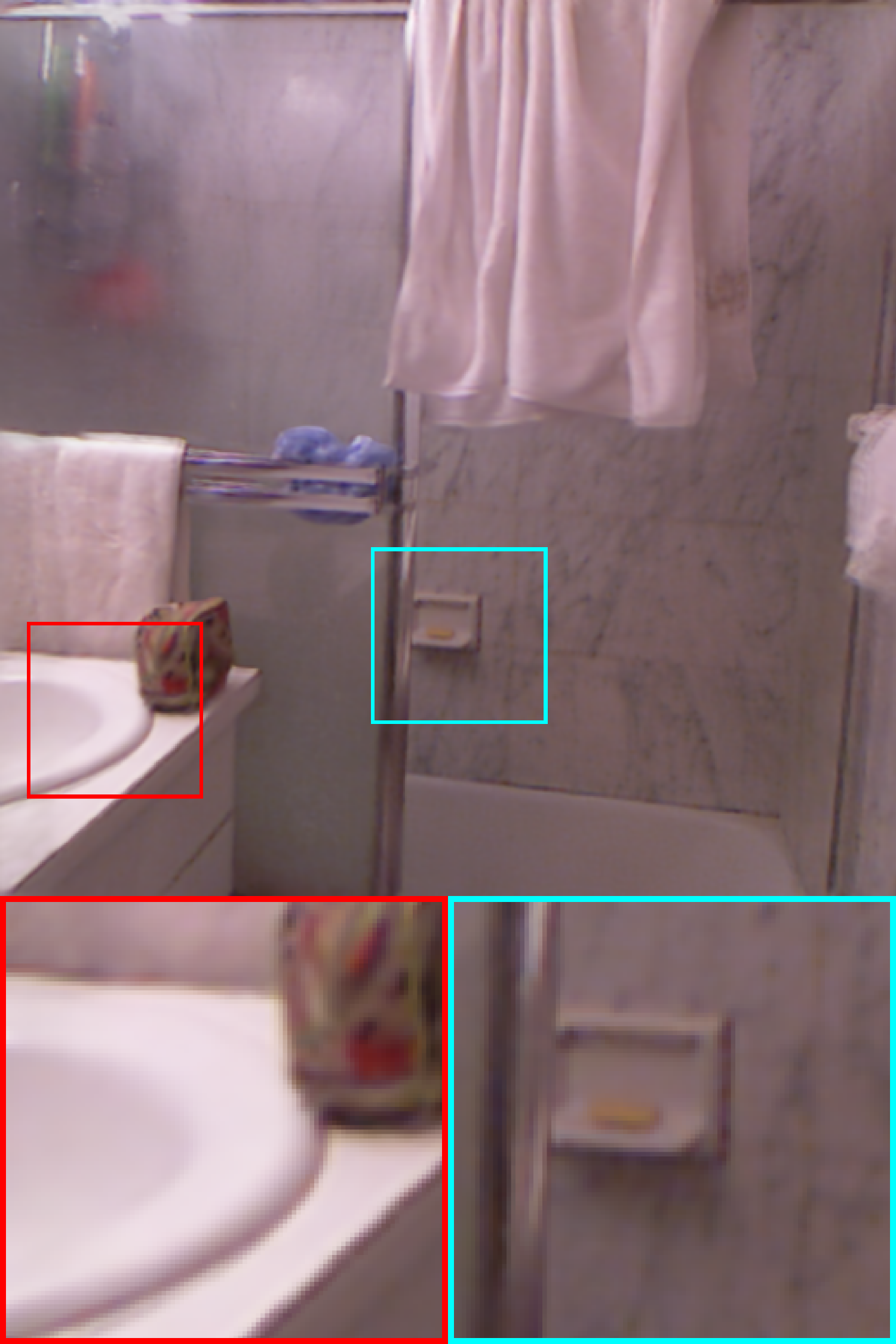} &
        \includegraphics[width=\imgsize]{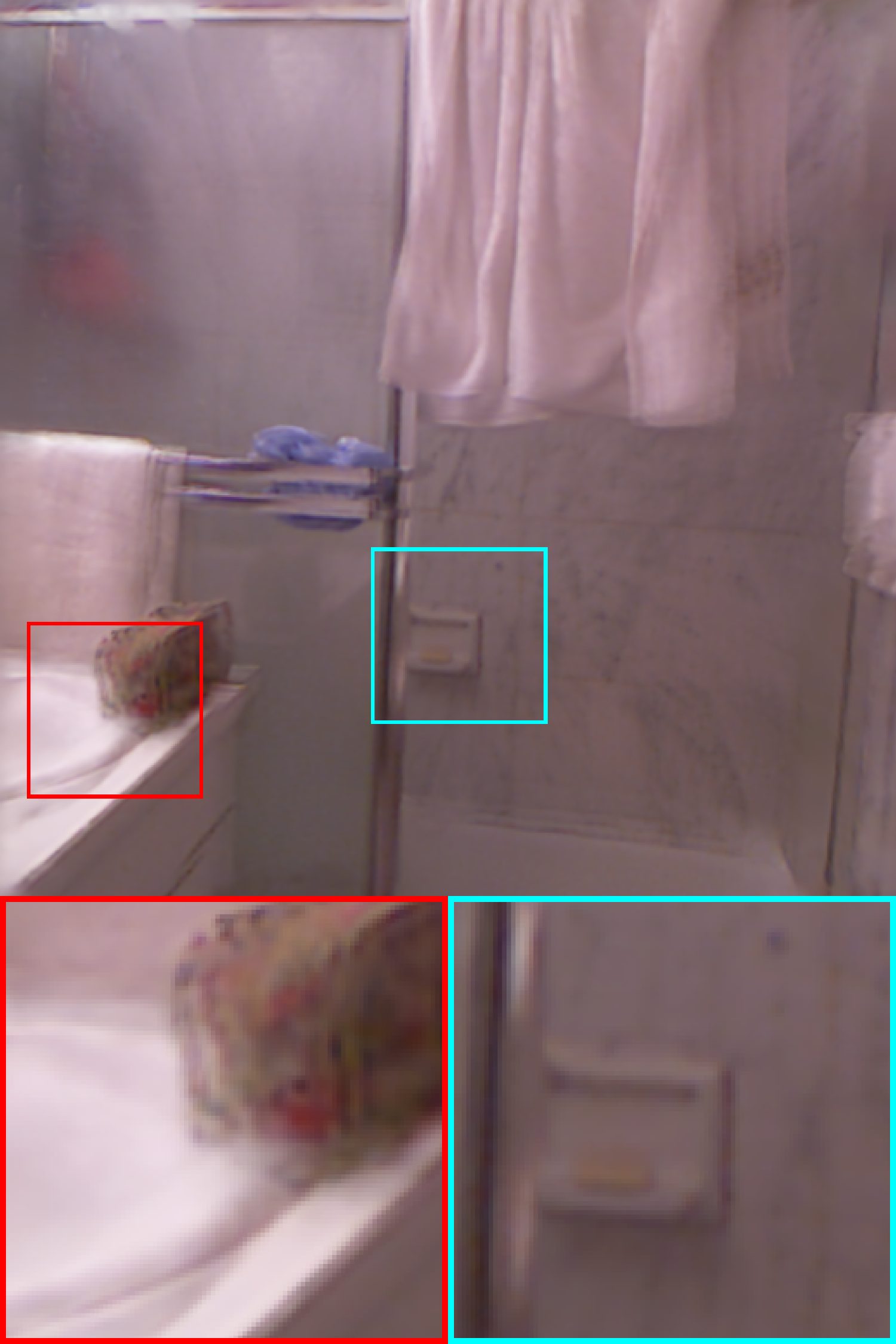} &
        \includegraphics[width=\imgsize]{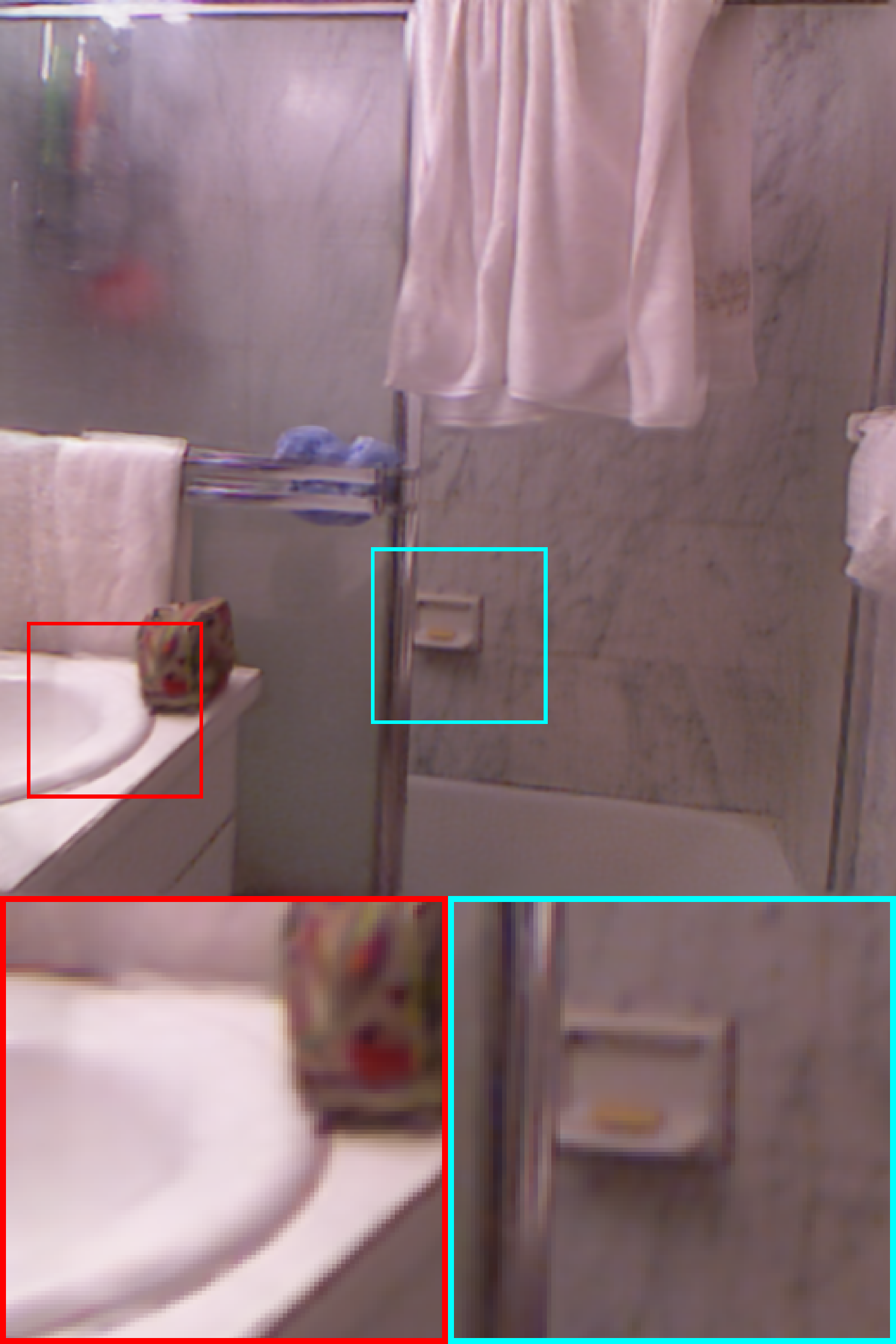} &
        \includegraphics[width=\imgsize]{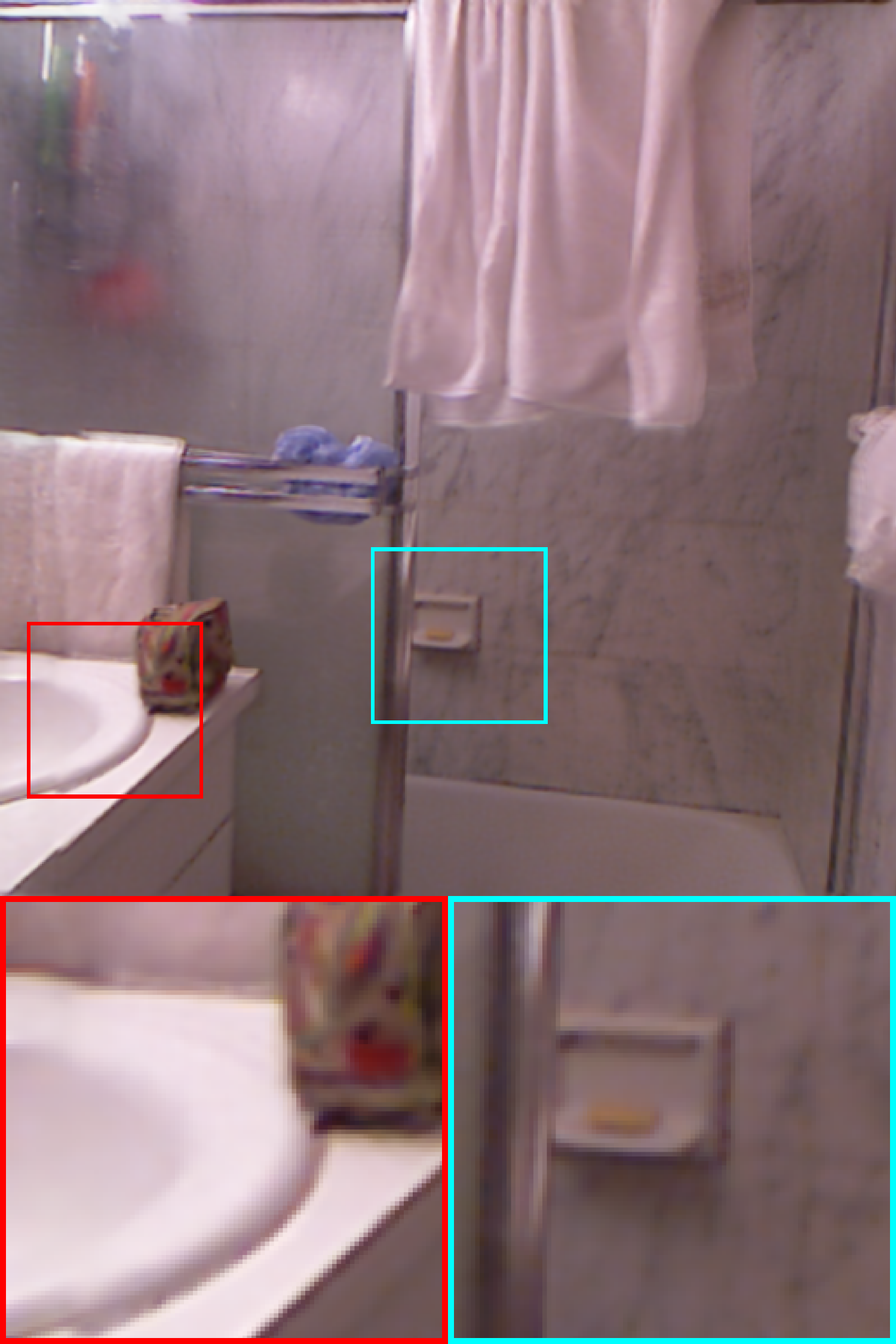} \\
        \includegraphics[width=\imgsize]{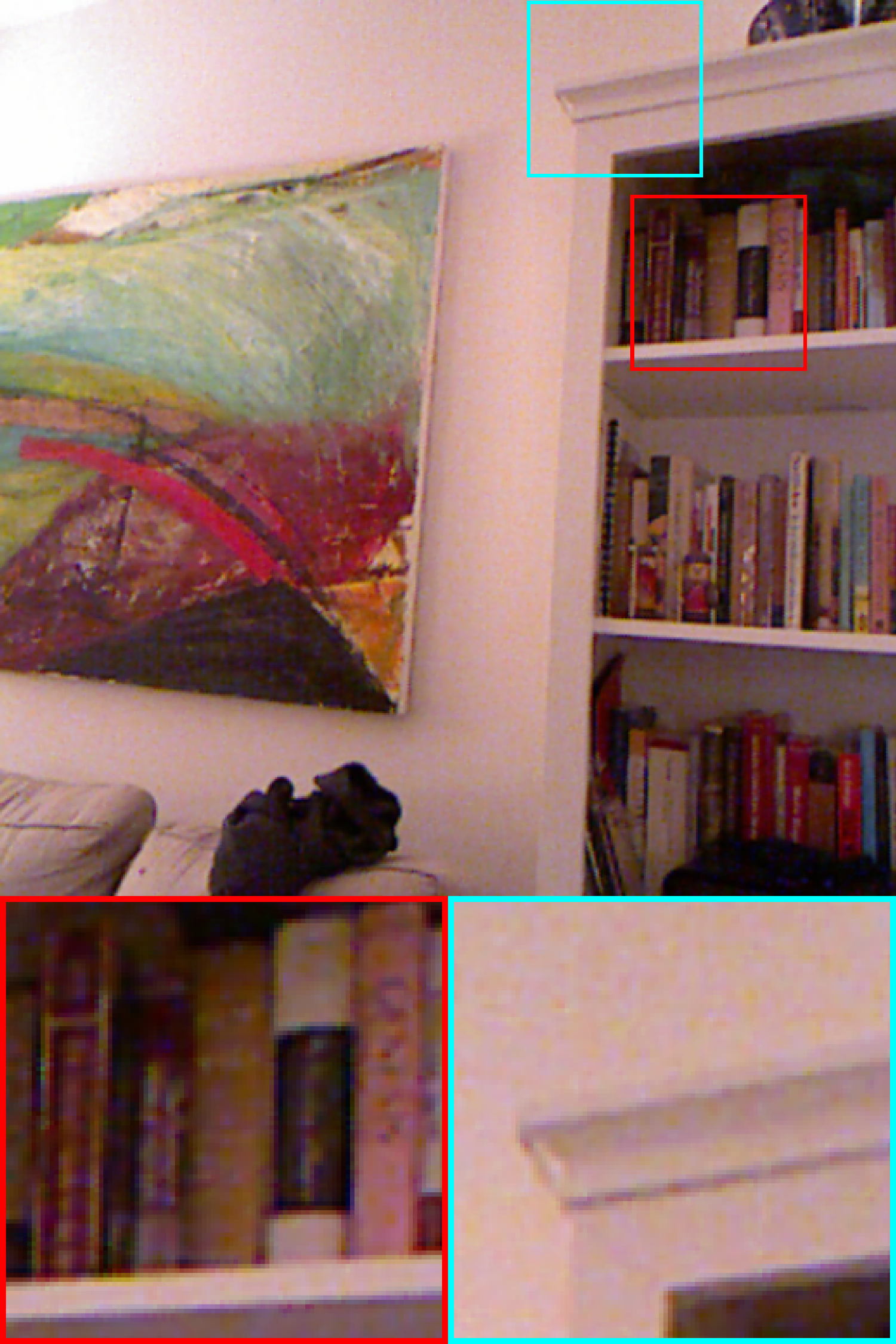} &
        \includegraphics[width=\imgsize]{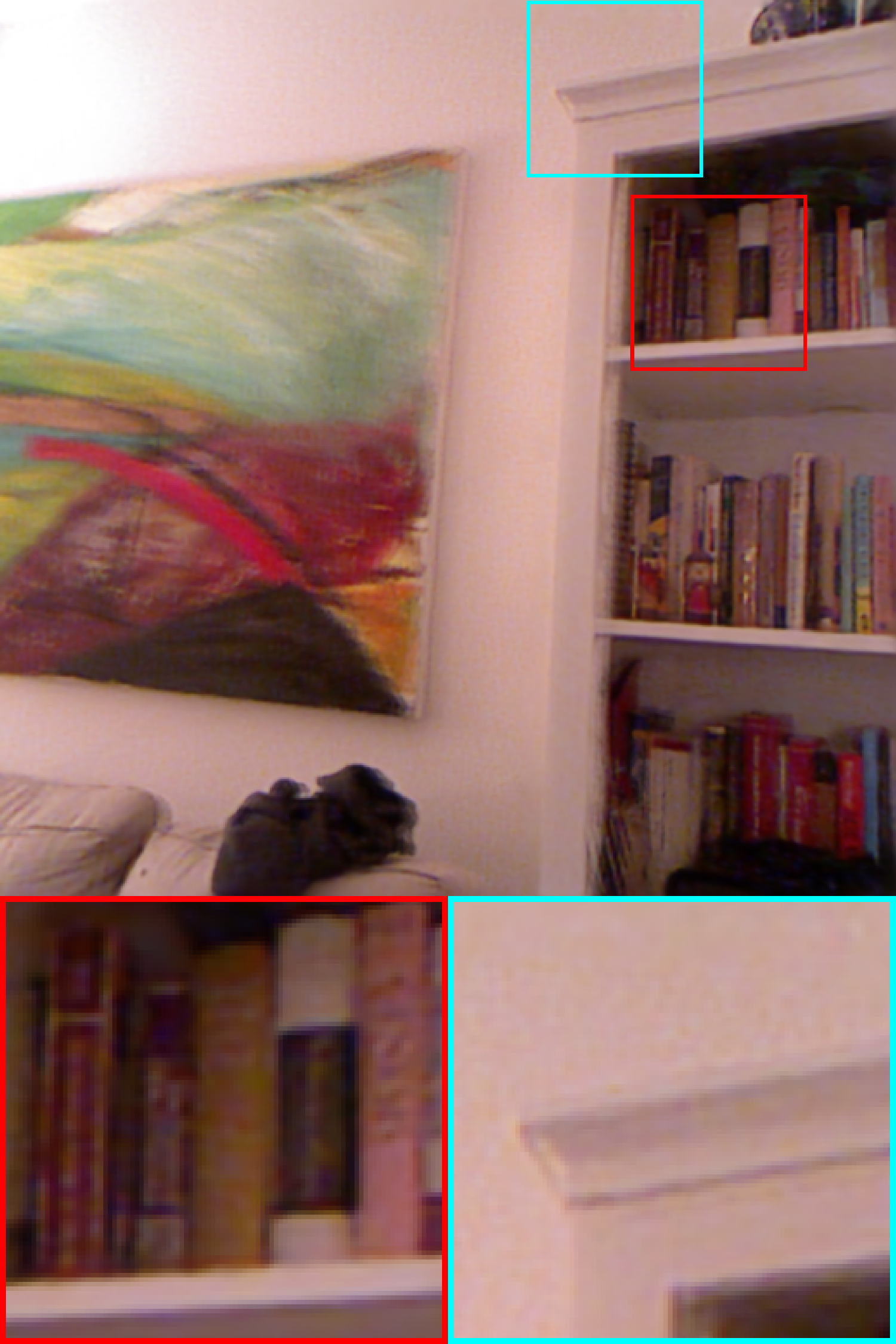} &
        \includegraphics[width=\imgsize]{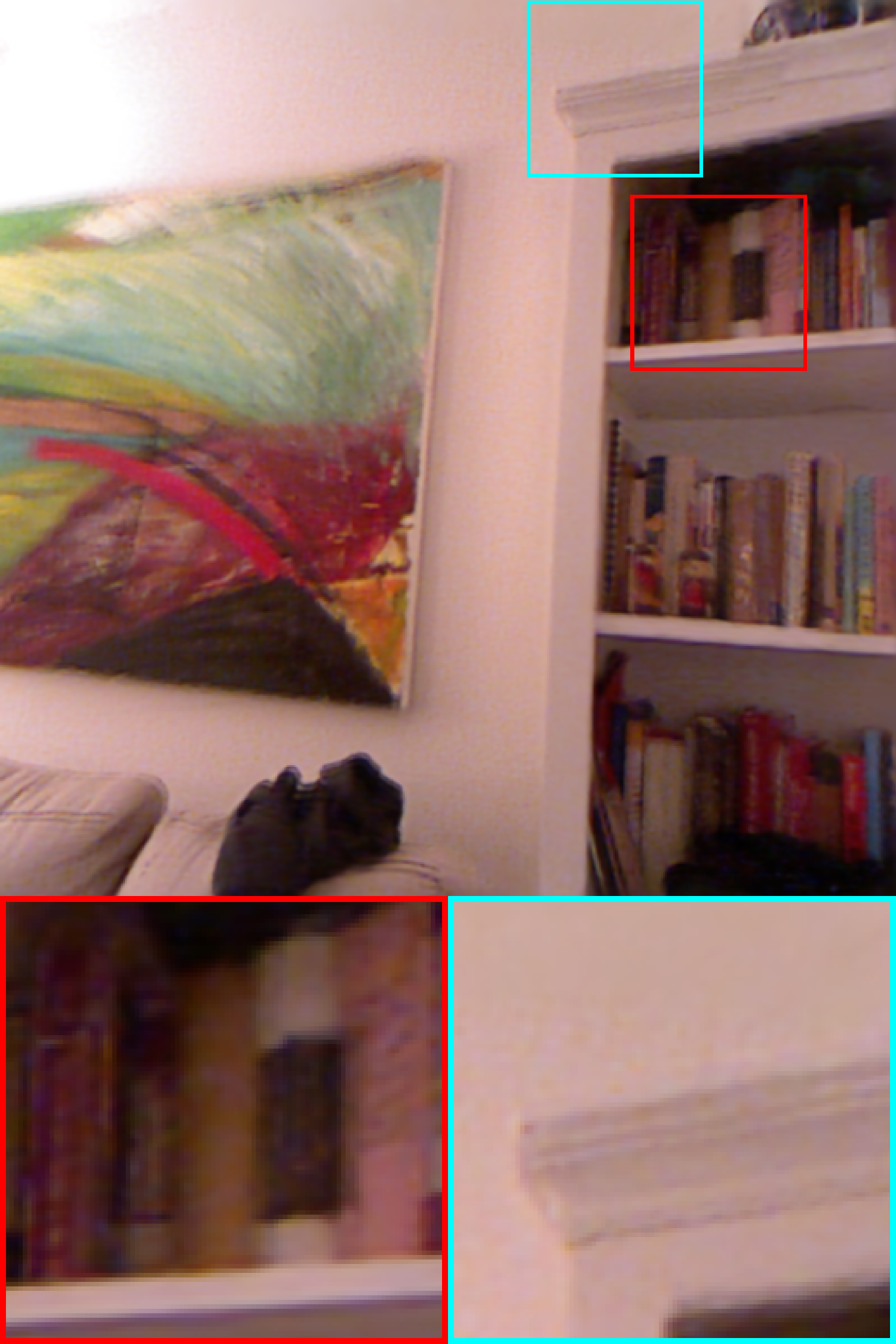} &
        \includegraphics[width=\imgsize]{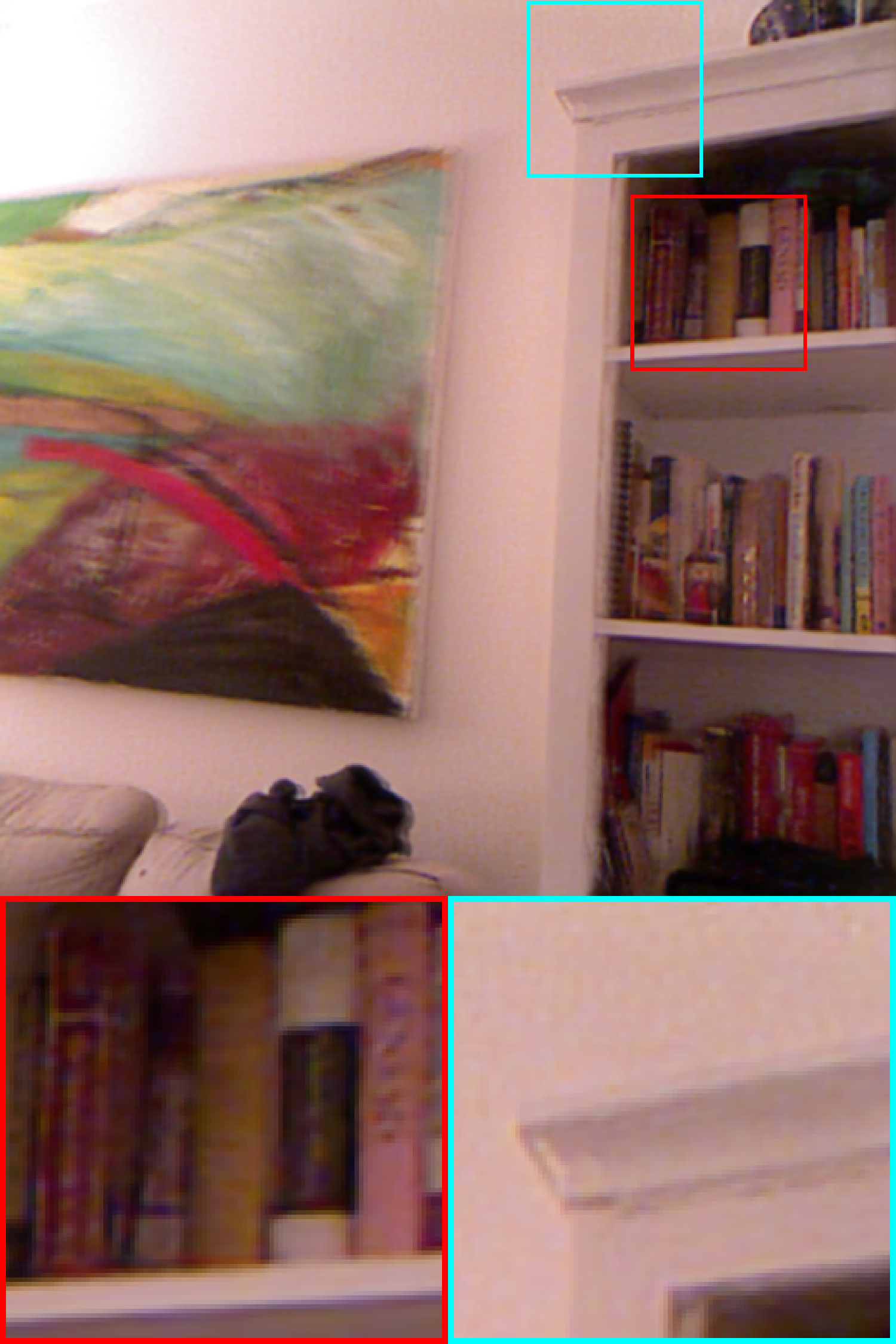} &
        \includegraphics[width=\imgsize]{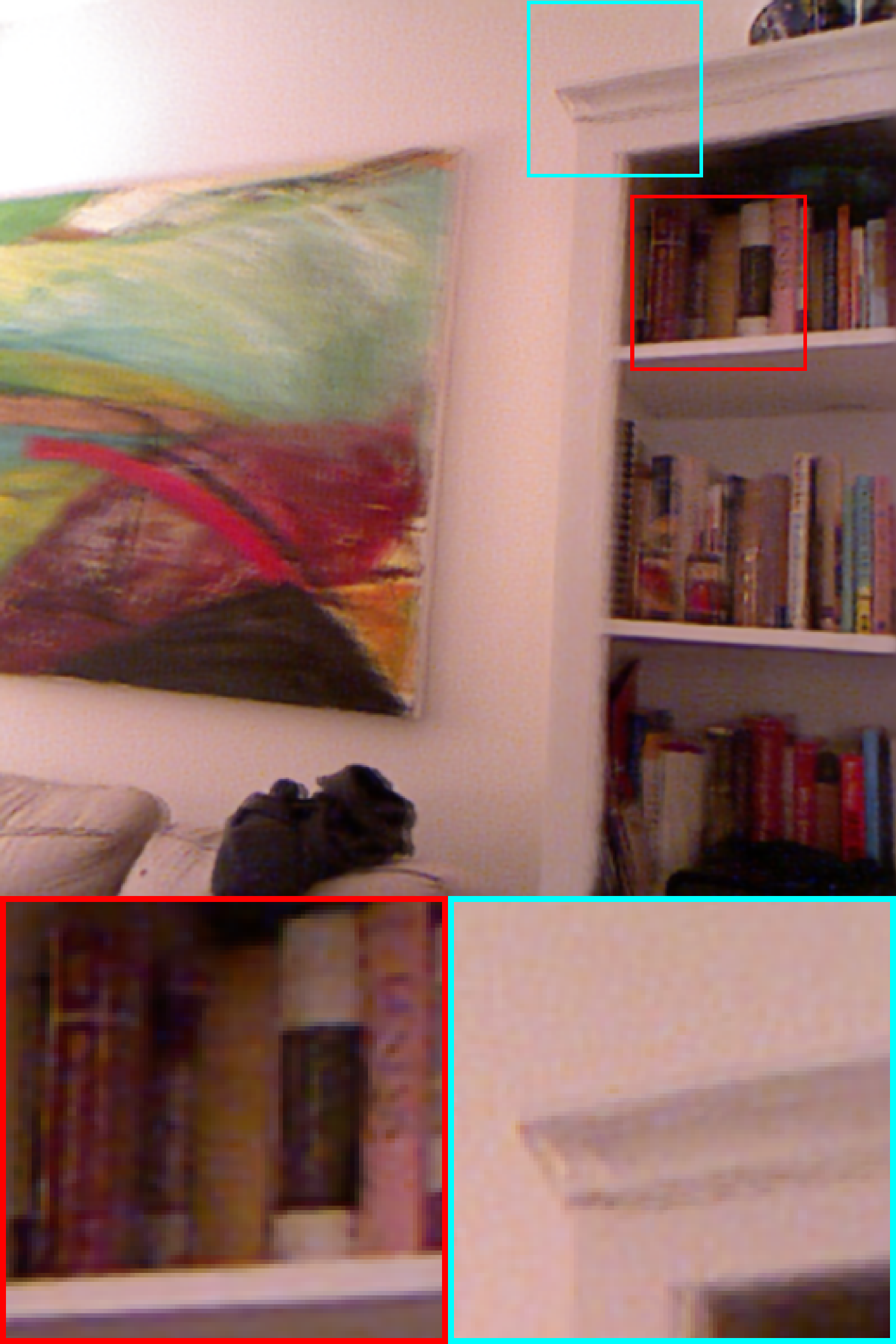} \\
        
        \scriptsize GT & 
        \scriptsize Ours (Full) & 
        \scriptsize w/o DIR-Head &
        \scriptsize w/o DPR-Offsets &
        \scriptsize w/o NV-Sup 
         \\
    \end{NiceTabular}
    
    \caption{Qualitative ablation study on \nyu datasets under 5 input views.}
    \label{fig:fig_abl_nyuv2}
\end{figure*}
To further validate the contribution of each component, we provide qualitative ablation results on the \dlbench, \re, and \nyu datasets in \fref{fig:fig_abl_dl3dv}, \ref{fig:fig_abl_re10k} and \ref{fig:fig_abl_nyuv2}. These visual comparisons reinforce the quantitative findings in the main paper and highlight the specific geometric challenges addressed by our framework.

\noindent\textbf{Impact of DIR-Head.} 
As illustrated in the third column of \fref{fig:fig_abl_dl3dv}, \ref{fig:fig_abl_re10k} and \ref{fig:fig_abl_nyuv2}, removing the DIR-Head leads to severe geometric drift and structural warping. Without the first-frame feature $\mathbf{F}_0$ to anchor the global coordinate system, frame-wise intrinsic instabilities accumulate throughout the sequence. This results in significant misalignment of global structures, such as ghosting artifacts on window frames (Fig.~\ref{fig:fig_abl_dl3dv}) and duplicated structural edges around picture frames and door (Fig.~\ref{fig:fig_abl_re10k}), confirming that DIR-Head is essential for long-term coordinate consistency in an unposed online setting.

We also find teacher forcing essential for training DIR-Head from scratch. Without the teacher prior, early erratic focal length predictions would lead to chaotic unprojections, preventing the downstream decoder from learning consistent scene geometry. The focal-gap analysis shows only minor inference-time mismatch between student and teacher focal lengths, with mean relative gaps of $2.15\%$, $3.90\%$, and $2.40\%$ on \dlbench, \re, and \nyu, respectively.

\noindent\textbf{Impact of DPR-Offsets.} 
Excluding the DPR-Offsets (the fourth column of \fref{fig:fig_abl_dl3dv}, \ref{fig:fig_abl_re10k} and \ref{fig:fig_abl_nyuv2}) results in noticeable ghosting artifacts and blurred boundaries. This is because conventional per-pixel 3DGS updates are often too rigid to compensate for the joint drift inherent in online pose and depth estimation. Consequently, newly spawned Gaussians fail to align precisely with historical geometry, leading to ghosting artifacts on thin structures like chair legs (Fig.~\ref{fig:fig_abl_dl3dv}) and door edges (Fig.~\ref{fig:fig_abl_re10k}). Our DPR-Offsets provide the necessary spatial flexibility to refine primitive positions for seamless fusion.

\noindent\textbf{Impact of NV-Sup.} 
The absence of Novel-View-Weighted Supervision (NV-Sup) typically leads to over-smoothed textures and reduced structural consistency in unseen regions. Without explicitly supervising extrapolated viewpoints, the model tends to overfit the appearance of input views rather than recovering accurate multi-view geometric cues. This results in a loss of crisp details, particularly visible on complex surfaces like bookshelves (Fig.~\ref{fig:fig_abl_nyuv2}).

\section{Failure Cases and Analysis}
\label{sec:supp_fail}

\begin{figure*}[t!]
    \centering
    \newcommand{\failimgsize}{0.22\textwidth} 
    \setlength{\tabcolsep}{2pt} 
    \renewcommand{\arraystretch}{1.2} 
    
    \begin{NiceTabular}{ccccc}
        \raisebox{3.5em}{\rotatebox[origin=c]{90}{GT}} &
        \includegraphics[width=\failimgsize]{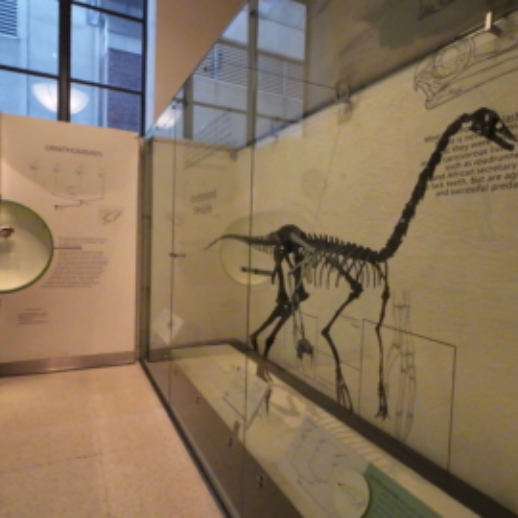} &
        \includegraphics[width=\failimgsize]{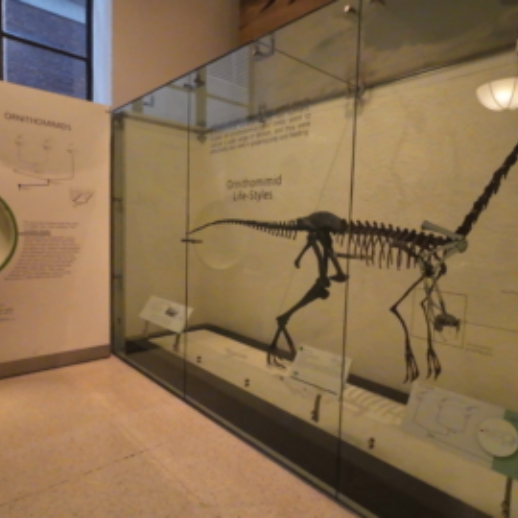} &
        \includegraphics[width=\failimgsize]{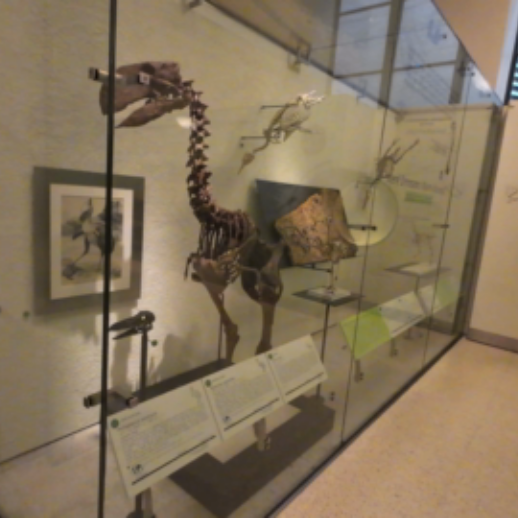} &
        \includegraphics[width=\failimgsize]{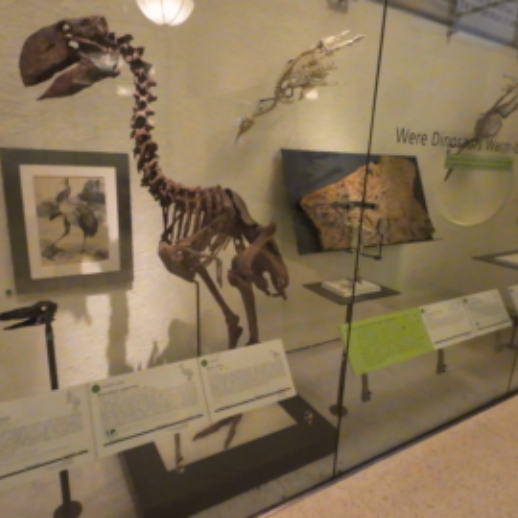} \\
        
        \raisebox{3.5em}{\rotatebox[origin=c]{90}{\small Novel Views}} &
        \includegraphics[width=\failimgsize]{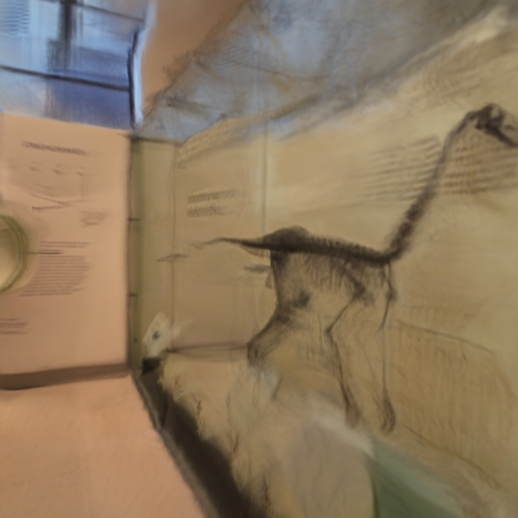} &
        \includegraphics[width=\failimgsize]{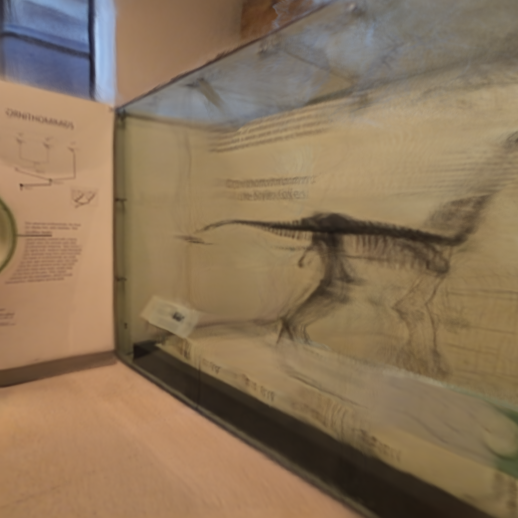} &
        \includegraphics[width=\failimgsize]{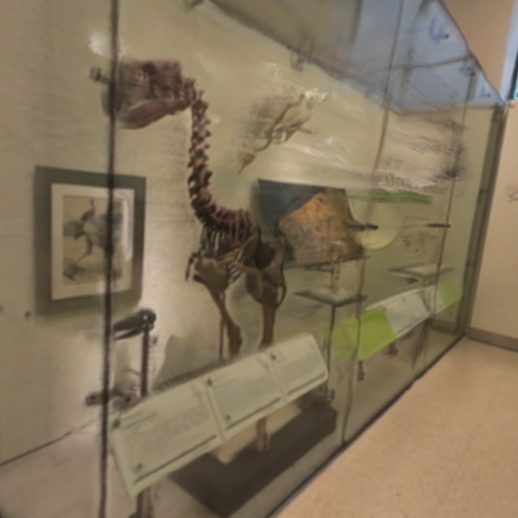} &
        \includegraphics[width=\failimgsize]{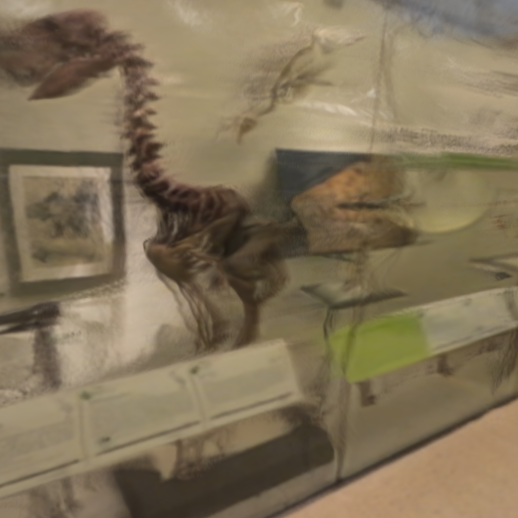} \\
        
        & \multicolumn{4}{c}{\small (a) Challenges in scenes with transparency.} \\
        \vspace{2pt} \\ 

        \raisebox{3.5em}{\rotatebox[origin=c]{90}{GT}} &
        \includegraphics[width=\failimgsize]{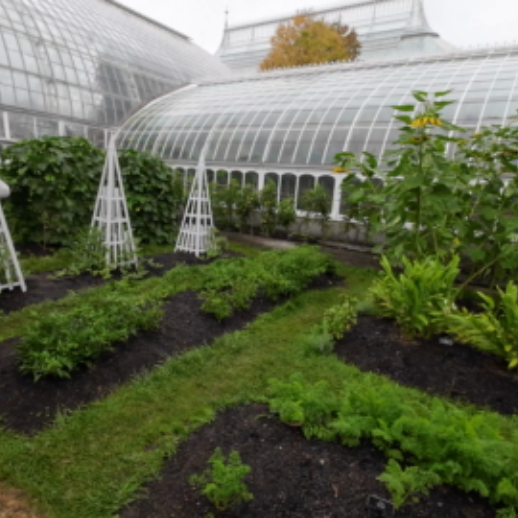} &
        \includegraphics[width=\failimgsize]{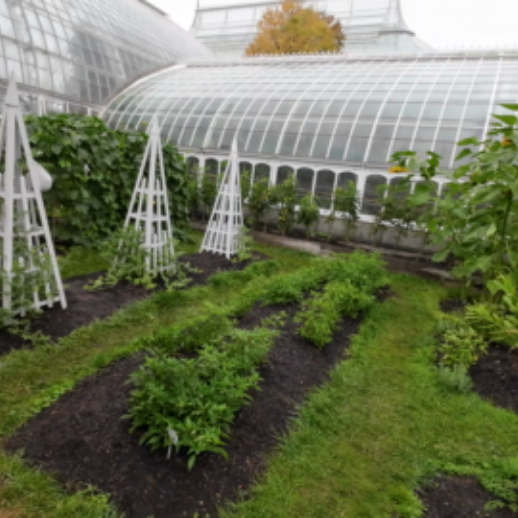} &
        \includegraphics[width=\failimgsize]{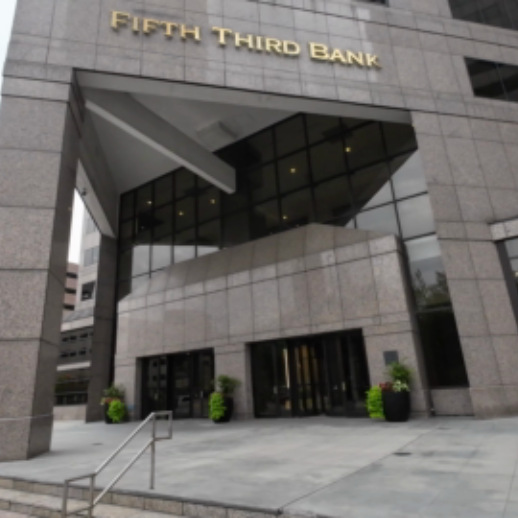} &
        \includegraphics[width=\failimgsize]{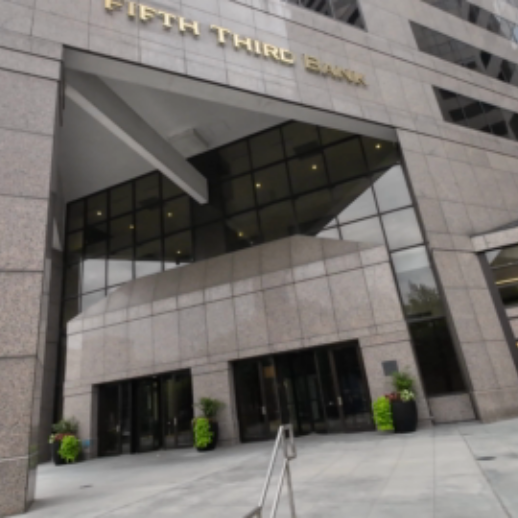} \\
        
        \raisebox{3.5em}{\rotatebox[origin=c]{90}{Novel Views}} &
        \includegraphics[width=\failimgsize]{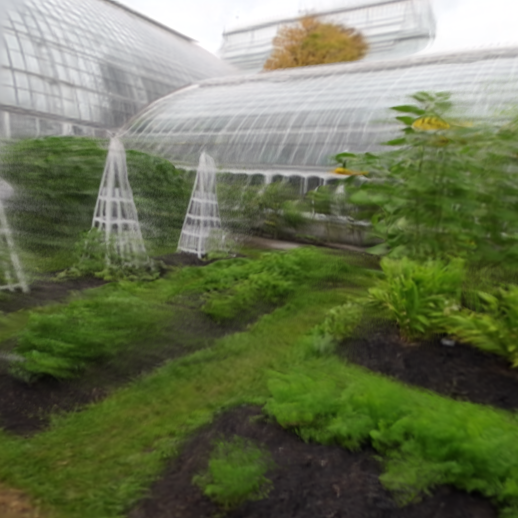} &
        \includegraphics[width=\failimgsize]{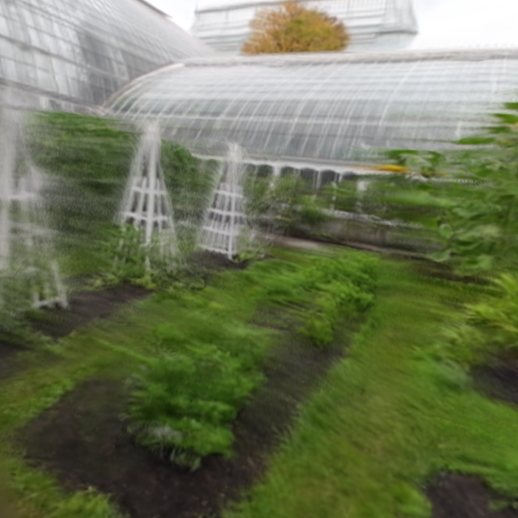} &
        \includegraphics[width=\failimgsize]{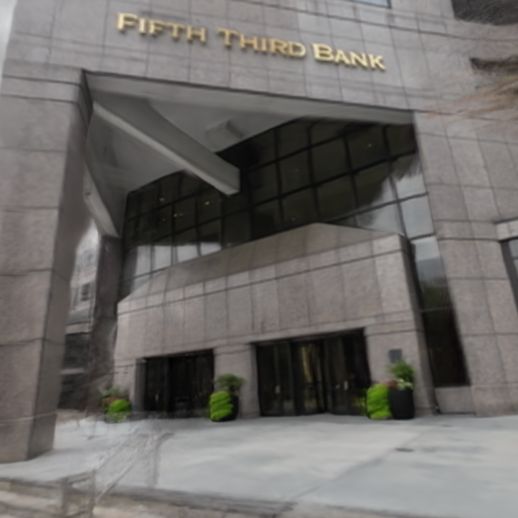} &
        \includegraphics[width=\failimgsize]{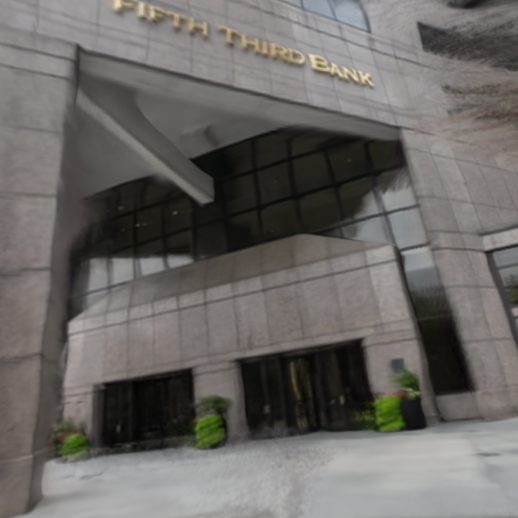} \\

        & \multicolumn{4}{c}{\small (b) Challenges in thin structures.} \\
        
    \end{NiceTabular}
    
    \caption{Visual analysis of typical failure cases. (a) The presence of transparent surfaces (\eg, glass windows) leads to suboptimal reconstruction due to the inherent ambiguity in depth estimation and complex light transport. (b) The presence of thin structures and high-frequency natural elements (\eg, dense grass) leads to suboptimal reconstruction due to the insufficient detail in depth priors and the ineffective fusion of redundant Gaussians.}
    \label{fig:fig_fail}
\end{figure*}
As illustrated in \fref{fig:fig_fail}, our method exhibits suboptimal performance in specific challenging scenarios. For regions with transparency (a), the inherent ambiguity in depth estimation—which often perceives depths behind the glass surface—coupled with complex light transport like reflections leads to blurred geometric artifacts. For scenarios containing thin structures or dense natural elements like grass (b), the limited resolution of depth estimation fails to capture fine-grained geometry. Moreover, the lack of a more sophisticated fusion mechanism to effectively consolidate redundant Gaussians in these high-frequency areas results in reduced structural fidelity and over-smoothing.

Beyond these visual artifacts, our framework shares the limitations discussed in the main paper. First, it is not designed to scale to ultra-long videos, where bounded-memory maintenance and long-range consistency become critical. Second, it remains challenging on highly dynamic scenes and videos with time-varying camera intrinsics, where stronger temporal modeling and more flexible intrinsic recovery are still needed.
\section{More Qualitative Experiments}
\label{sec:supp_qual}

\begin{figure*}[t!]
    \makeatletter\setlength{\@fptop}{0pt}\makeatother
    \centering
    
    \newcommand{\smallw}{0.12\textwidth} 
    \newcommand{\bigw}{0.365\textwidth}  
    \newcommand{\inputw}{0.365\textwidth} 
    
    \setlength{\tabcolsep}{1pt} 
    \begin{NiceTabular}{cccc}
        \Block{2-1}{\includegraphics[width=\inputw]{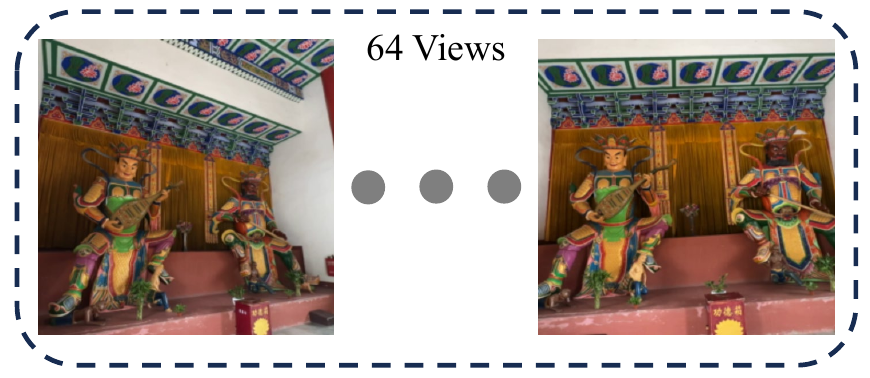}} &
        \Block{2-1}{\includegraphics[width=\bigw]{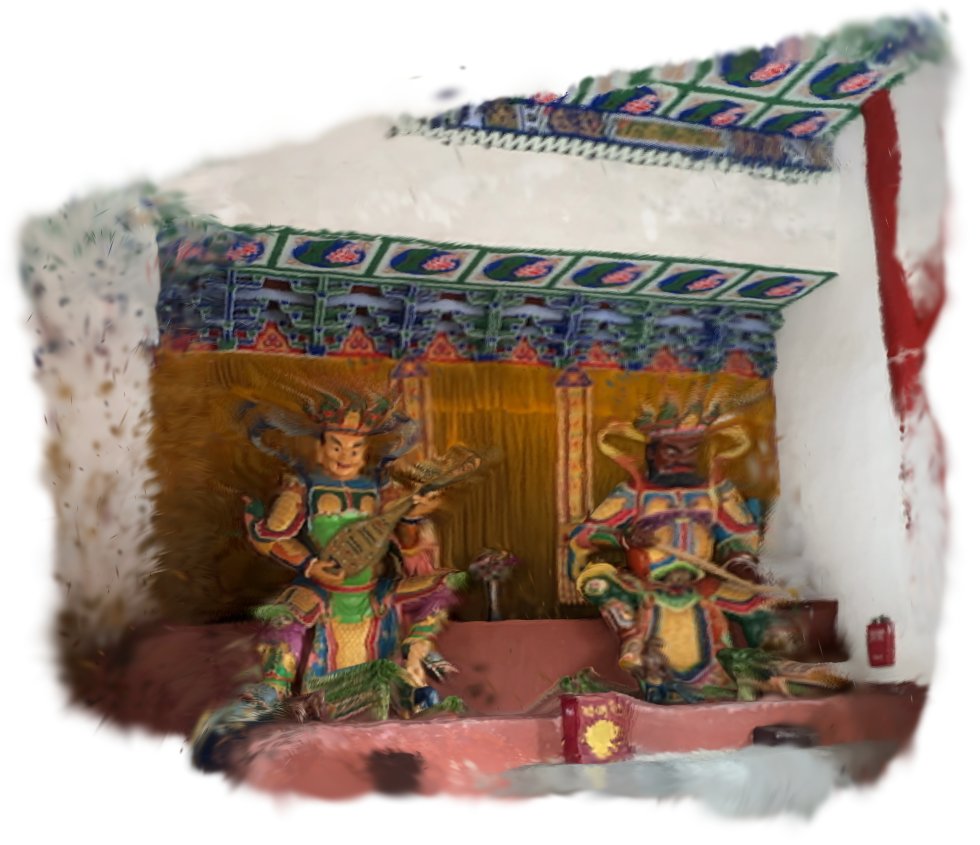}} &
        \includegraphics[width=\smallw]{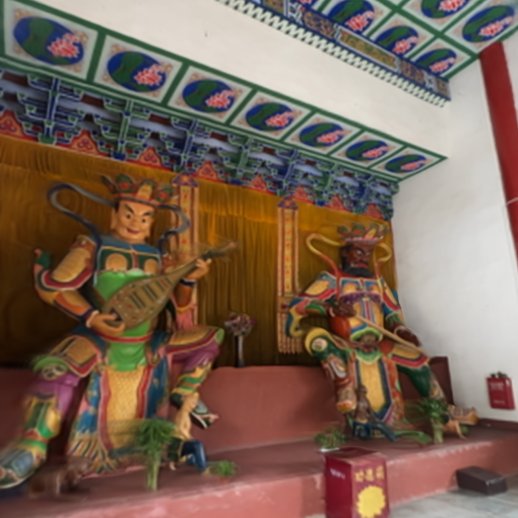} &
        \includegraphics[width=\smallw]{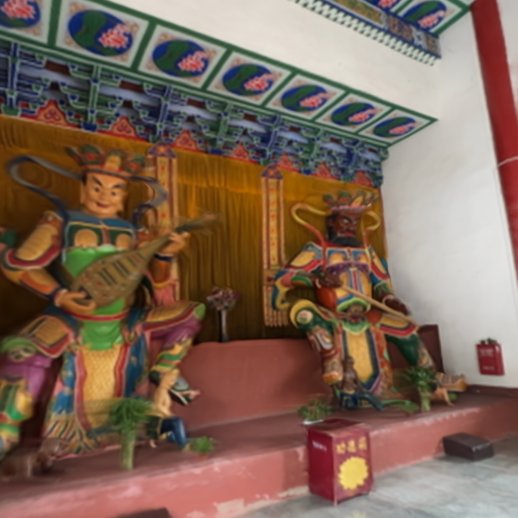} \\
        
         & & 
        \includegraphics[width=\smallw]{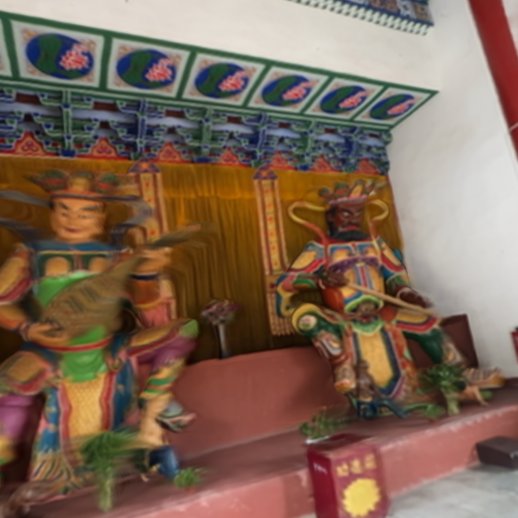} &
        \includegraphics[width=\smallw]{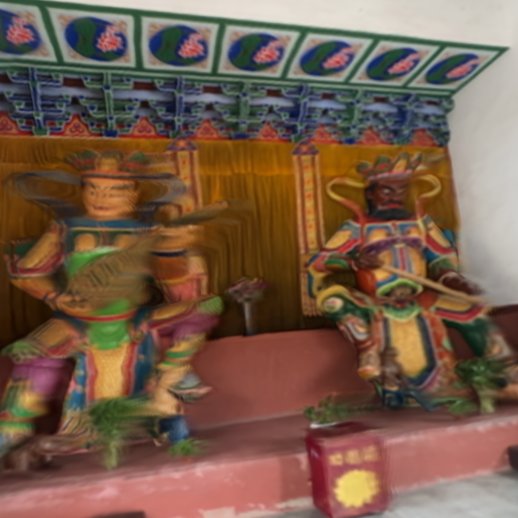} \\
        

        \Block{2-1}{\includegraphics[width=\inputw]{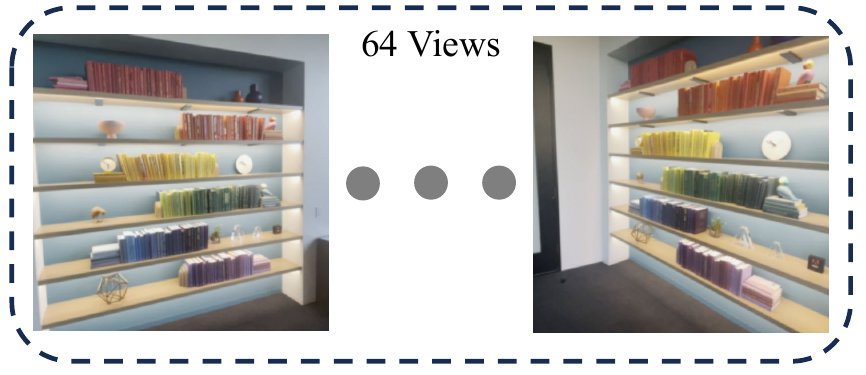}} &
        \Block{2-1}{\includegraphics[width=\bigw]{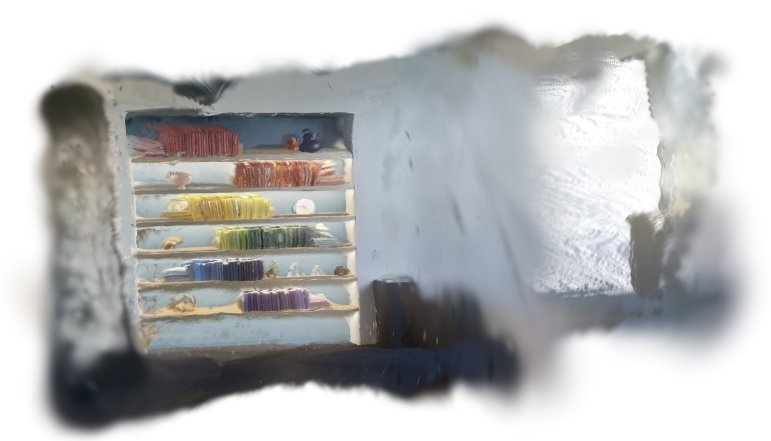}} &
        \includegraphics[width=\smallw]{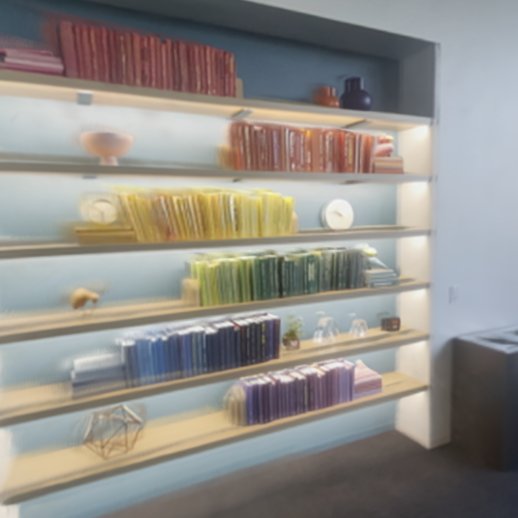} &
        \includegraphics[width=\smallw]{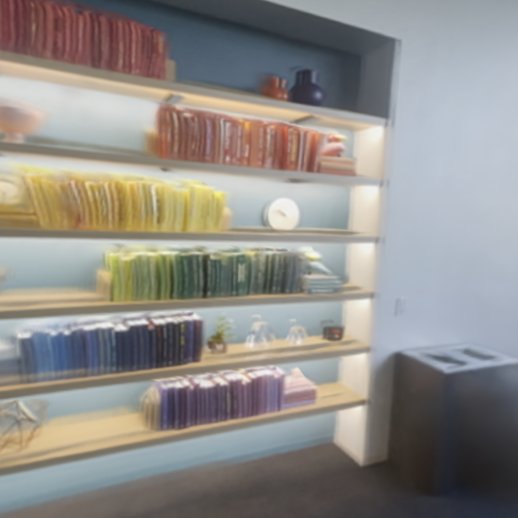} \\
         & & 
        \includegraphics[width=\smallw]{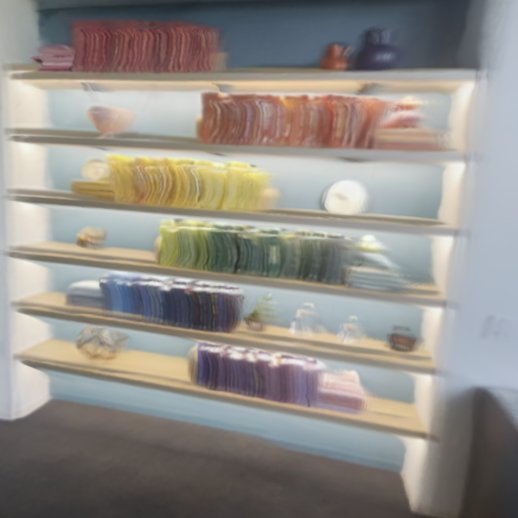} &
        \includegraphics[width=\smallw]{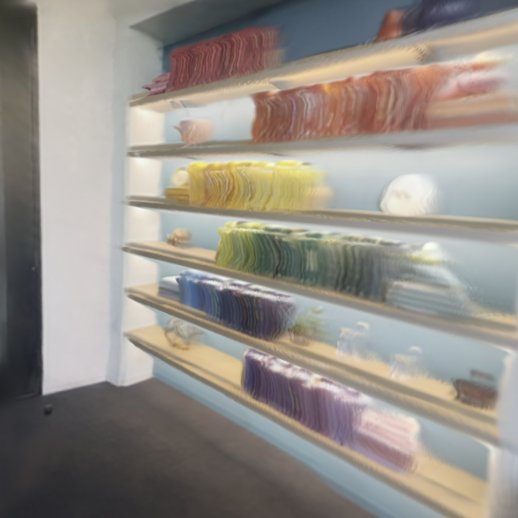} \\

        \noalign{\vspace{3pt}}

        \Block{2-1}{\includegraphics[width=\inputw]{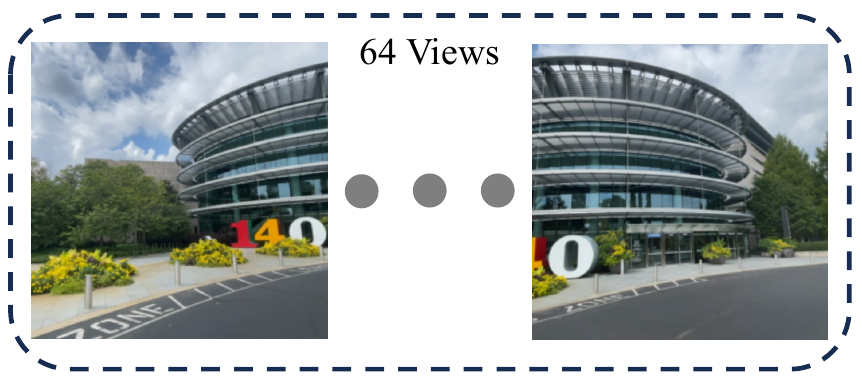}} &
        \Block{2-1}{\includegraphics[width=\bigw]{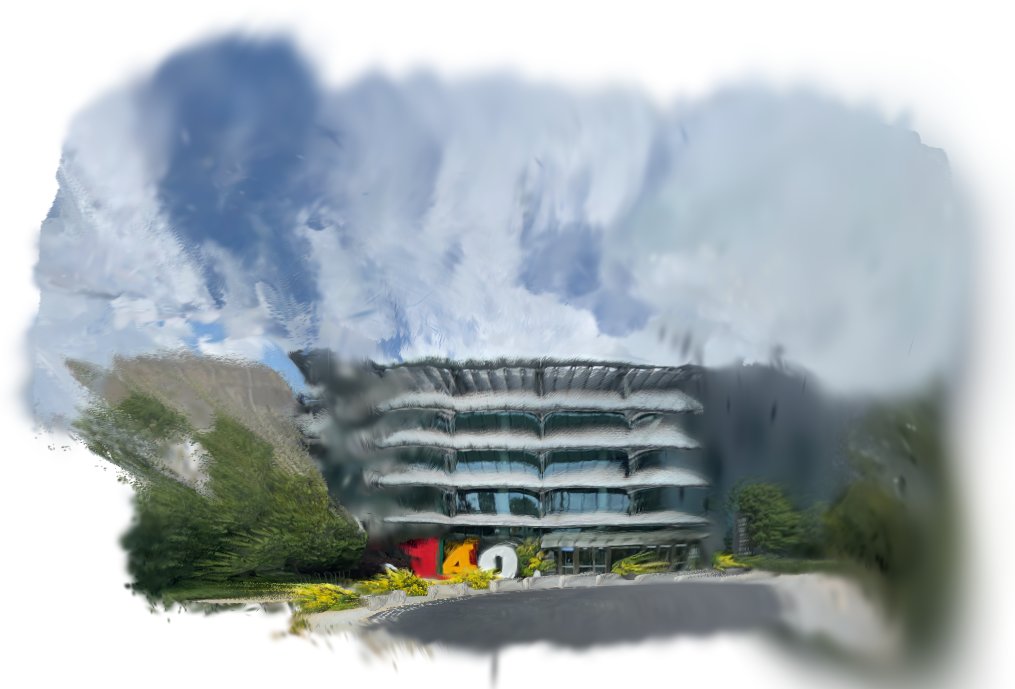}} &
        \includegraphics[width=\smallw]{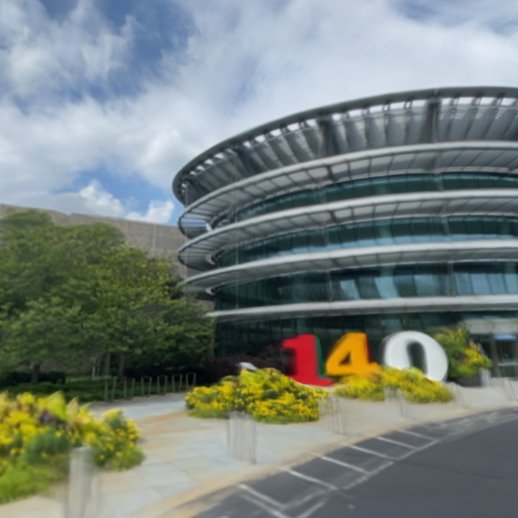} &
        \includegraphics[width=\smallw]{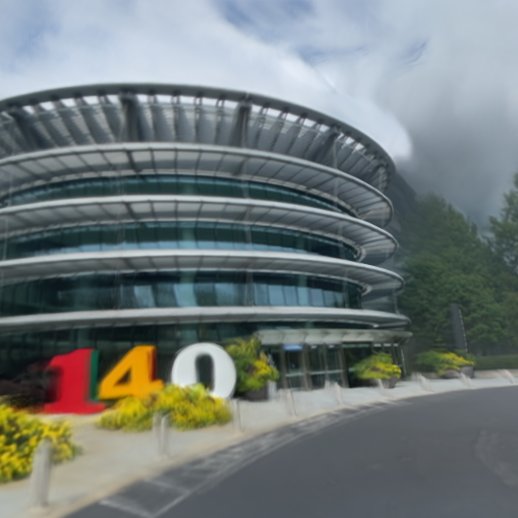} \\
         & & 
        \includegraphics[width=\smallw]{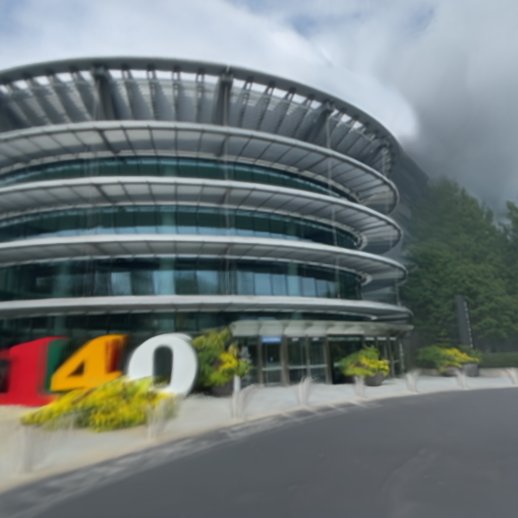} &
        \includegraphics[width=\smallw]{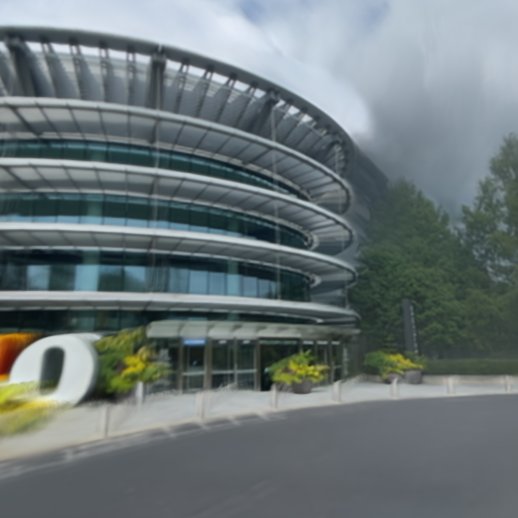} \\
        
        \noalign{\vspace{3pt}}
        \scriptsize Inputs & 
        \scriptsize Feed-forward Online 3DGS & 
         \\
    \end{NiceTabular}
    
    \caption{Qualitative comparison on \dlbench datasets under 64 input views. Each row shows one scene. Left: Input views. Middle: Reconstructed 3DGS. Right: Rendered novel views from ours.}
    \label{fig:fig_nvs_dl3dv64}
\end{figure*}

\begin{figure*}[t!]
    \makeatletter\setlength{\@fptop}{0pt}\makeatother
    \centering
    
    \newcommand{\smallw}{0.12\textwidth} 
    \newcommand{\bigw}{0.345\textwidth}  
    \newcommand{\inputw}{0.345\textwidth} 
    
    \setlength{\tabcolsep}{1pt} 
    \begin{NiceTabular}{cccc}
        \Block{2-1}{\includegraphics[width=\inputw]{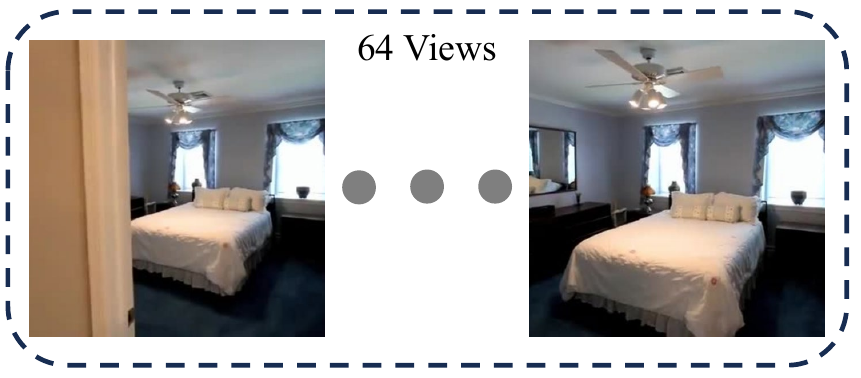}} &
        \Block{2-1}{\includegraphics[width=0.305\textwidth]{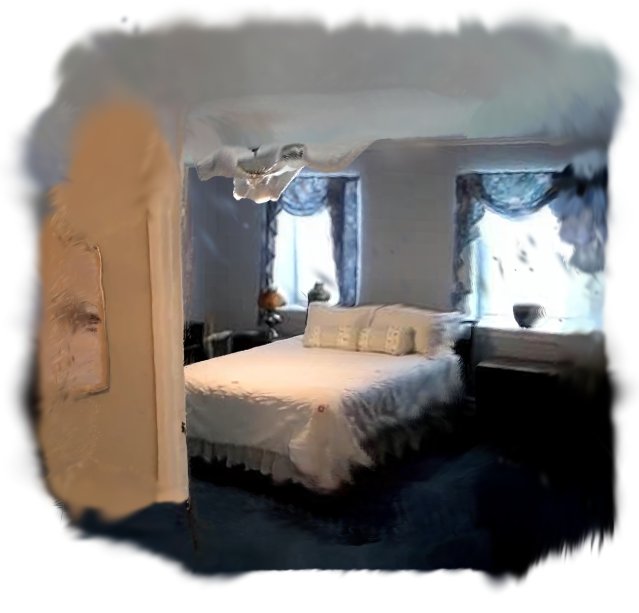}} &
        \includegraphics[width=\smallw]{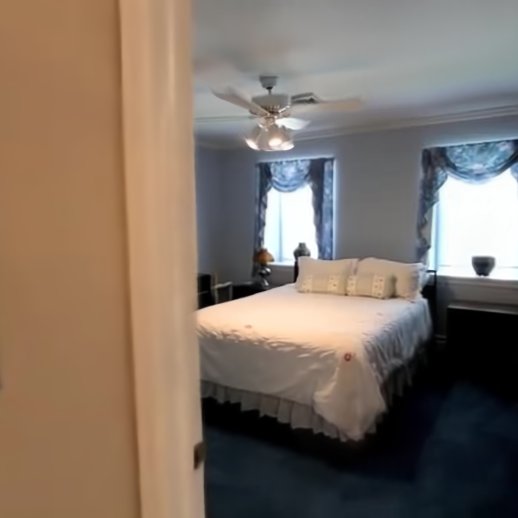} &
        \includegraphics[width=\smallw]{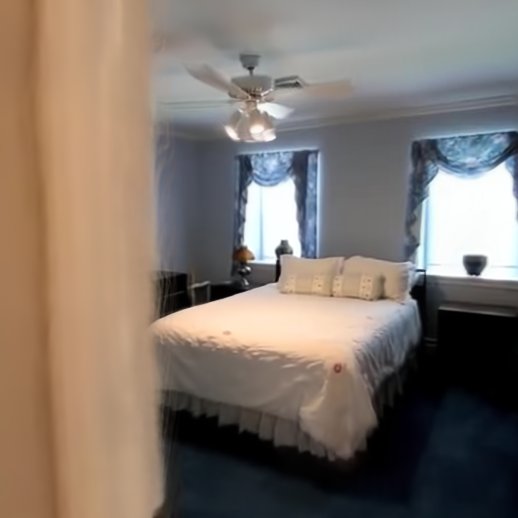} \\
        
         & & 
        \includegraphics[width=\smallw]{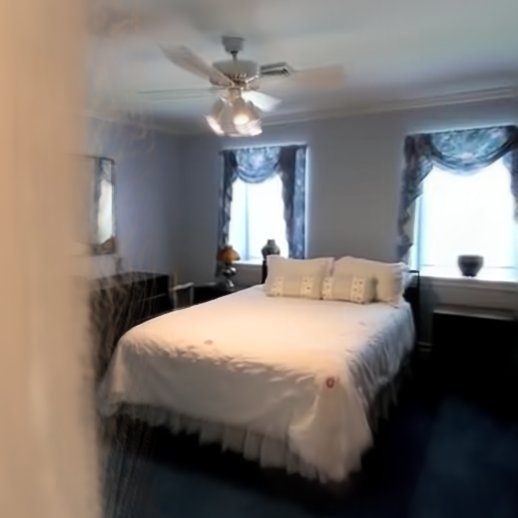} &
        \includegraphics[width=\smallw]{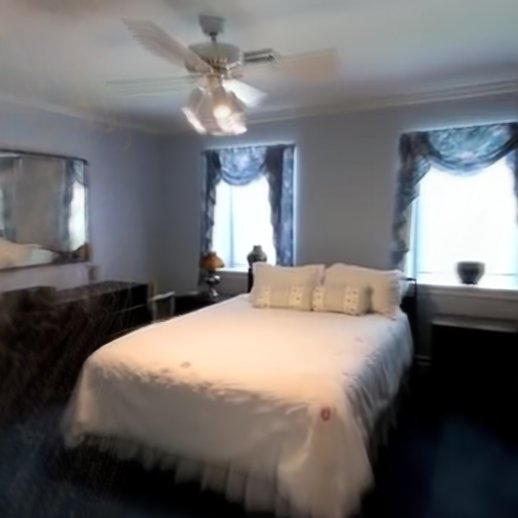} \\

        \noalign{\vspace{3pt}}

        \Block{2-1}{\includegraphics[width=\inputw]{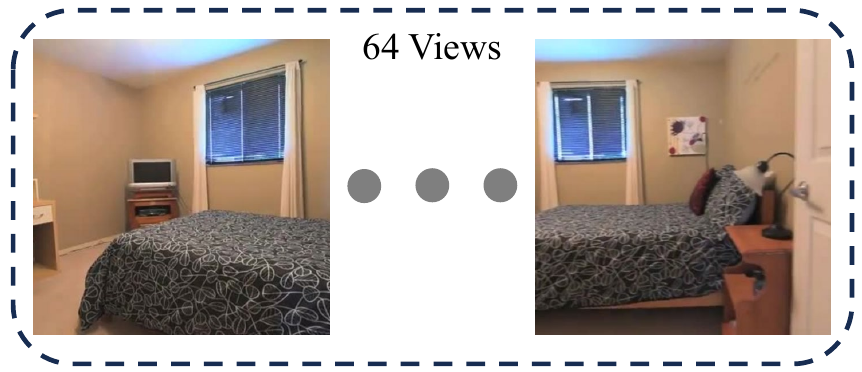}} &
        \Block{2-1}{\includegraphics[width=\bigw]{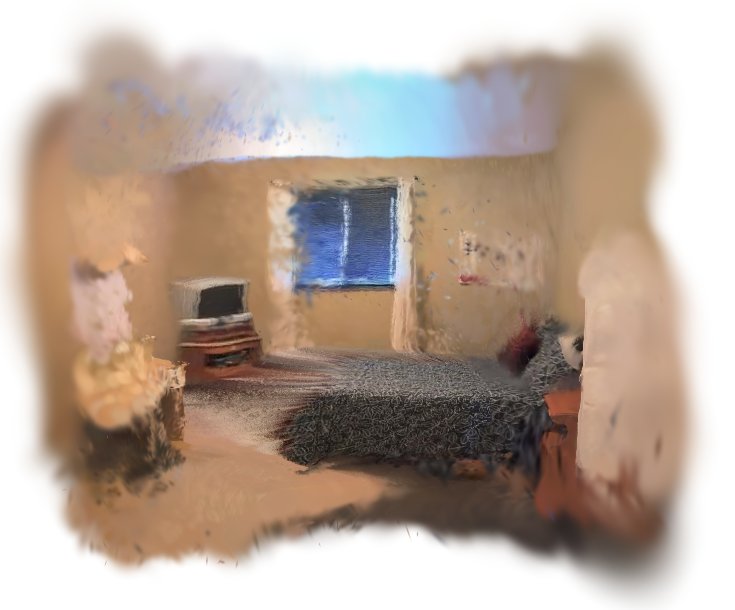}} &
        \includegraphics[width=\smallw]{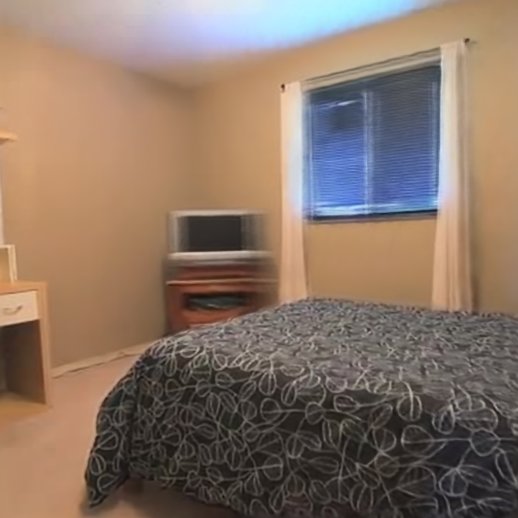} &
        \includegraphics[width=\smallw]{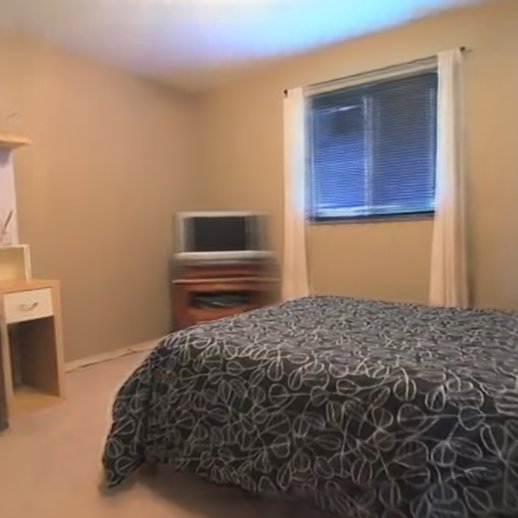} \\
        
         & & 
        \includegraphics[width=\smallw]{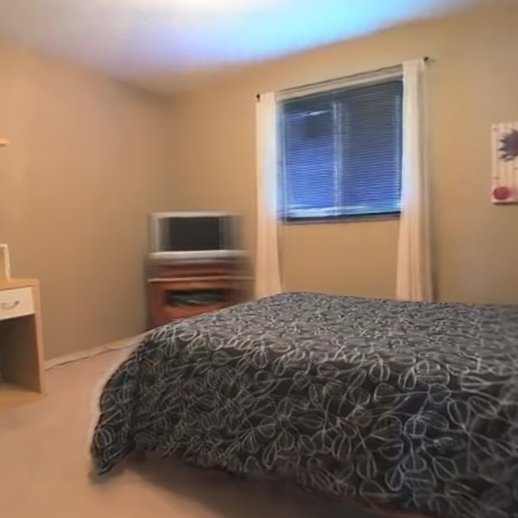} &
        \includegraphics[width=\smallw]{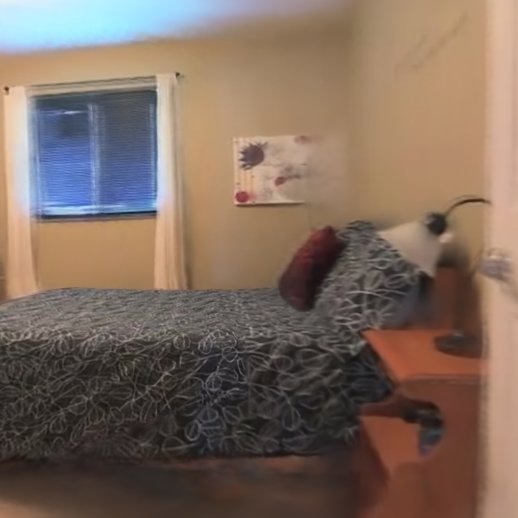} \\

        \noalign{\vspace{3pt}}

        \Block{2-1}{\includegraphics[width=\inputw]{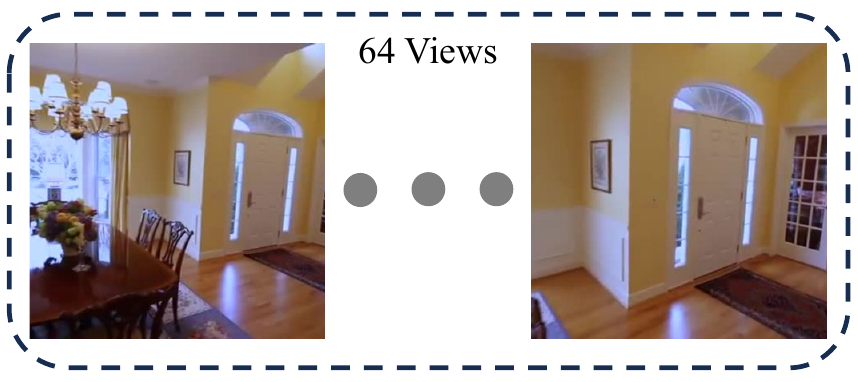}} &
        \Block{2-1}{\includegraphics[width=\bigw]{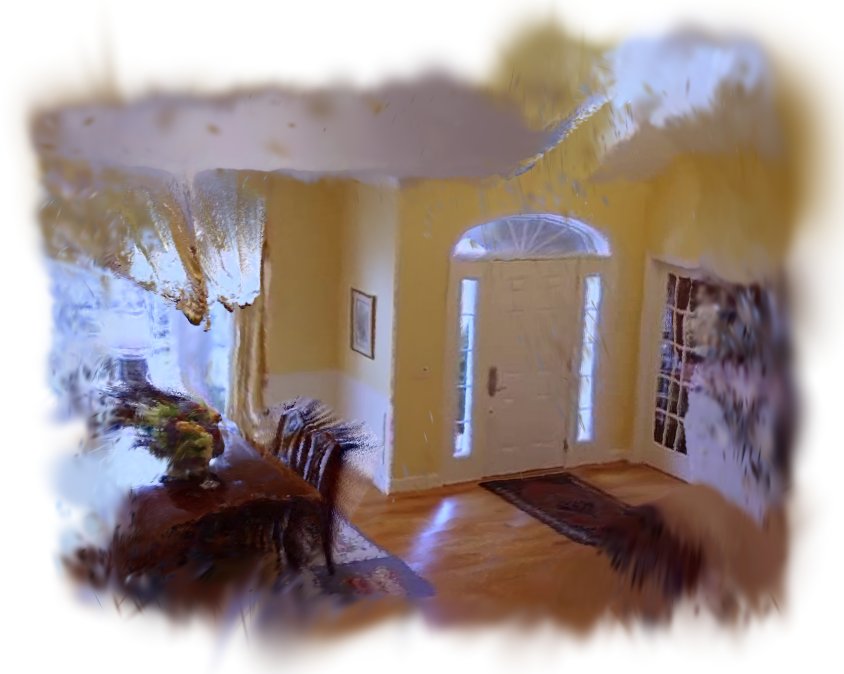}} &
        \includegraphics[width=\smallw]{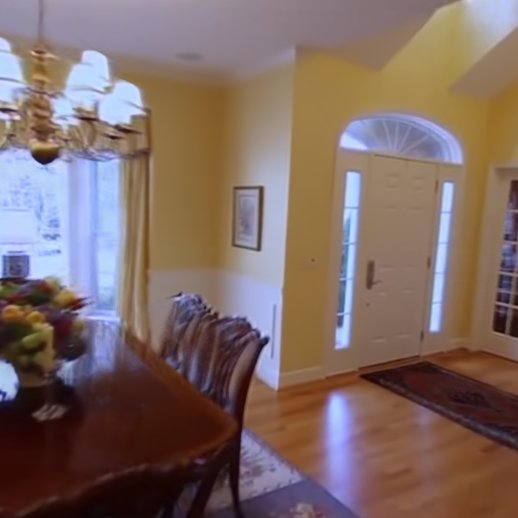} &
        \includegraphics[width=\smallw]{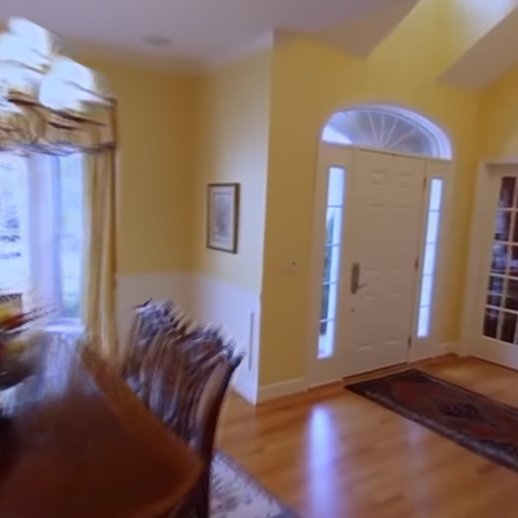} \\
        
         & & 
        \includegraphics[width=\smallw]{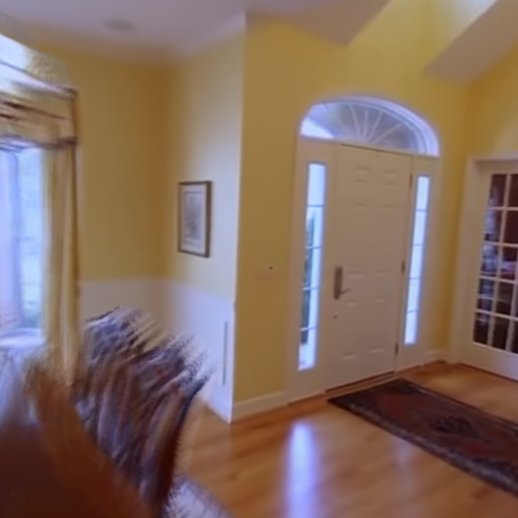} &
        \includegraphics[width=\smallw]{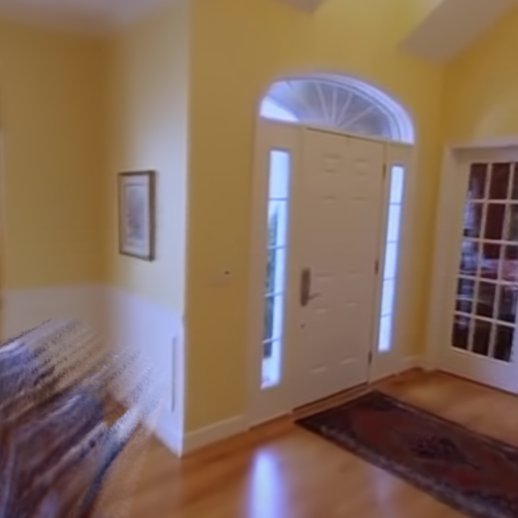} \\

        \noalign{\vspace{3pt}}
        \scriptsize Inputs & 
        \scriptsize Feed-forward Online 3DGS & 
         \\
    \end{NiceTabular}
    
    \caption{Qualitative comparison on \re datasets under 64 input views. Each row shows one scene. Left: Input views. Middle: Reconstructed 3DGS. Right: Rendered novel views from ours.}
    \vspace{-15pt}
    \label{fig:fig_nvs_re10k64}
\end{figure*}

\begin{figure*}[t!]
    \makeatletter\setlength{\@fptop}{0pt}\makeatother
    \centering
    
    \newcommand{\smallw}{0.12\textwidth} 
    \newcommand{\bigw}{0.345\textwidth}  
    \newcommand{\inputw}{0.345\textwidth} 
    
    \setlength{\tabcolsep}{1pt} 
    \begin{NiceTabular}{cccc}
        \Block{2-1}{\includegraphics[width=\inputw]{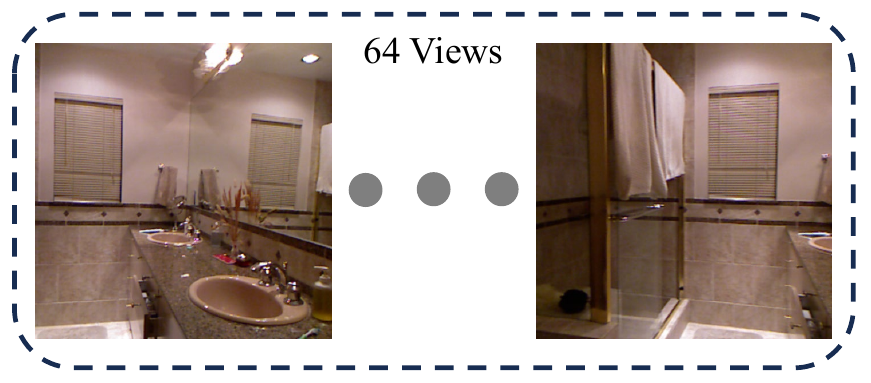}} &
        \Block{2-1}{\includegraphics[width=\bigw]{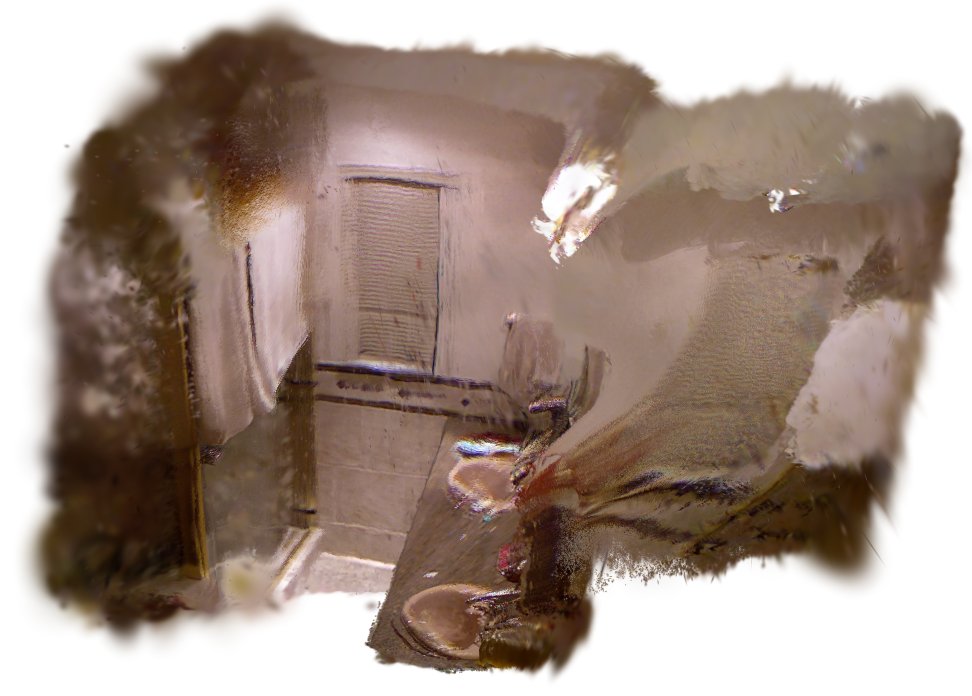}} &
        \includegraphics[width=\smallw]{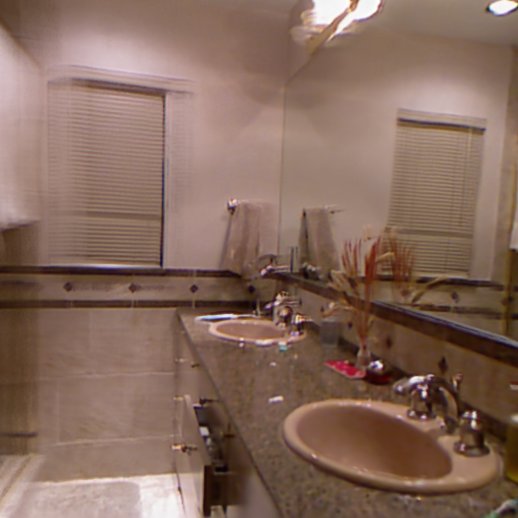} &
        \includegraphics[width=\smallw]{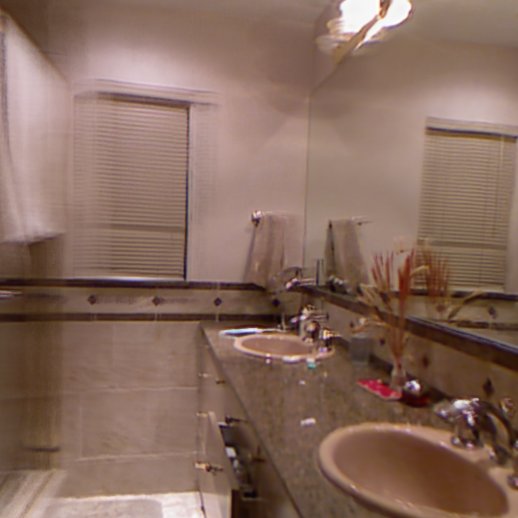} \\
        
         & & 
        \includegraphics[width=\smallw]{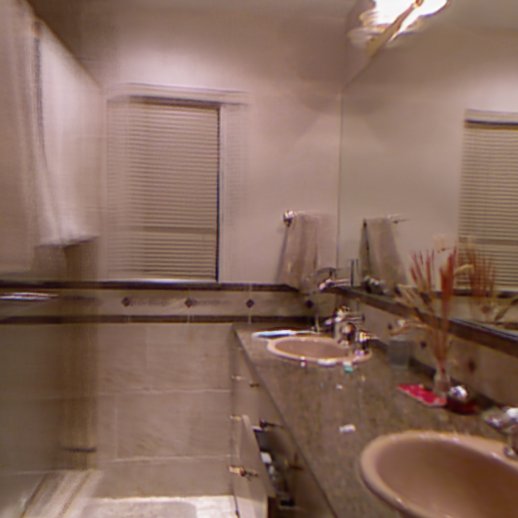} &
        \includegraphics[width=\smallw]{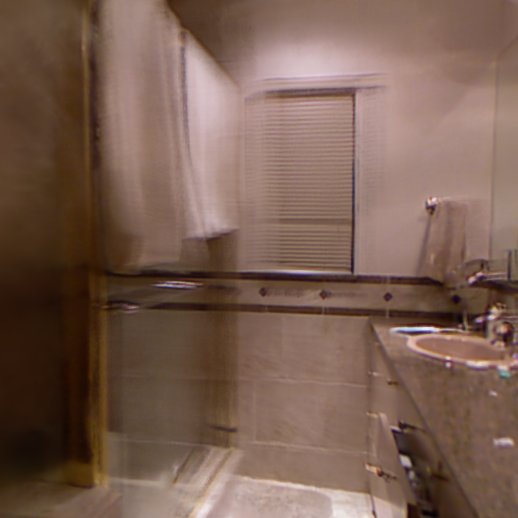} \\

        \noalign{\vspace{3pt}}

        \Block{2-1}{\includegraphics[width=\inputw]{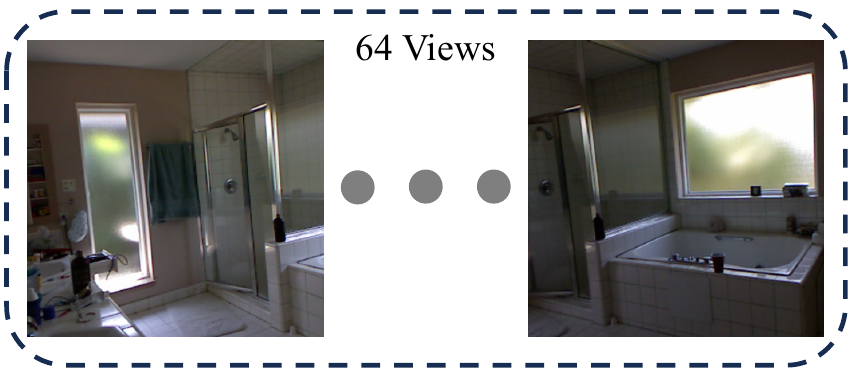}} &
        \Block{2-1}{\includegraphics[width=\bigw]{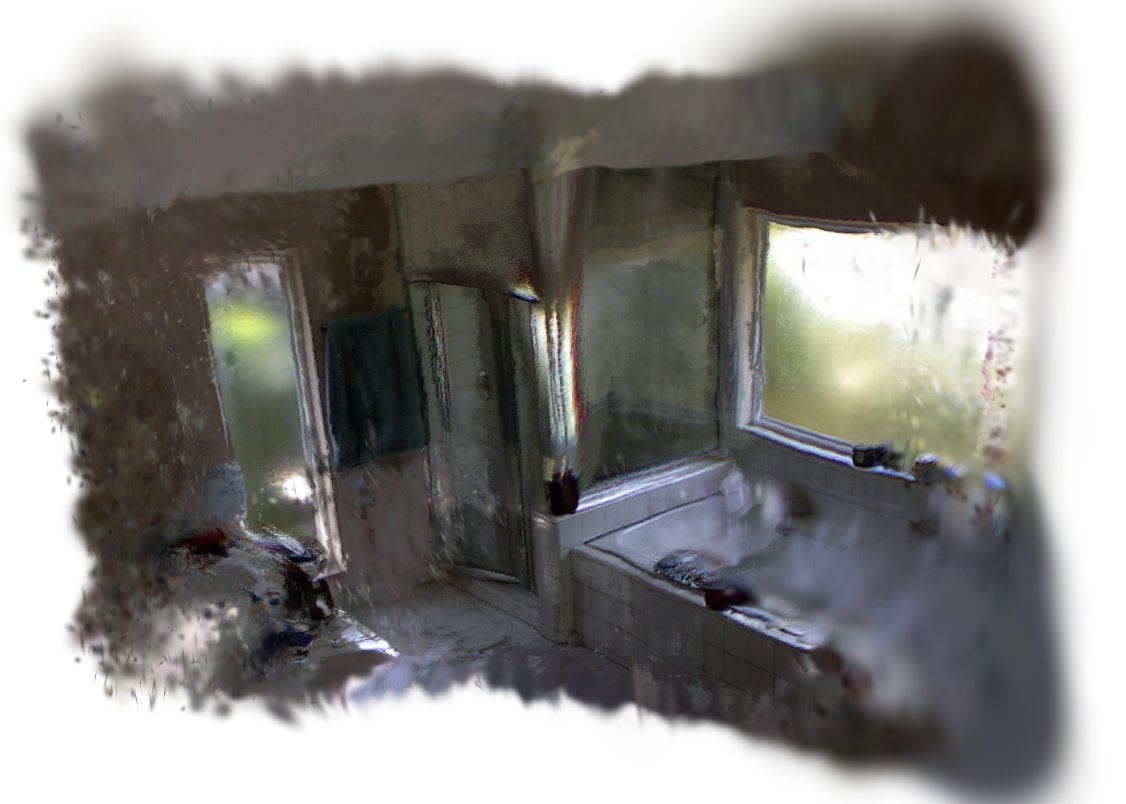}} &
        \includegraphics[width=\smallw]{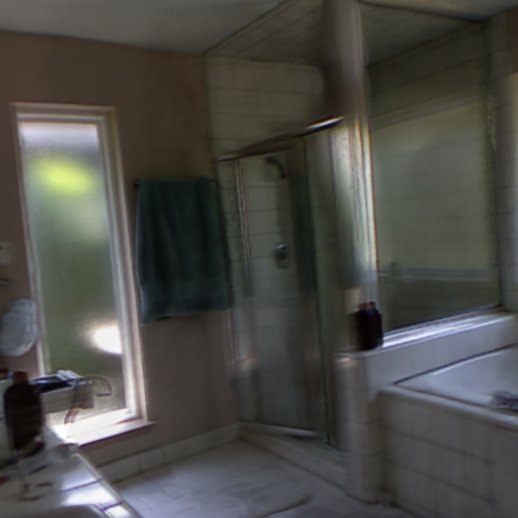} &
        \includegraphics[width=\smallw]{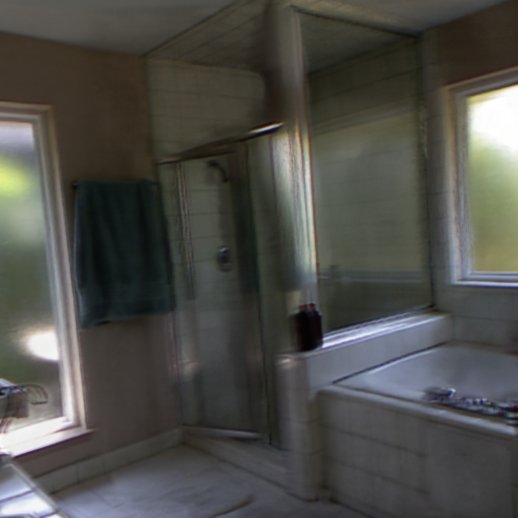} \\
        
         & & 
        \includegraphics[width=\smallw]{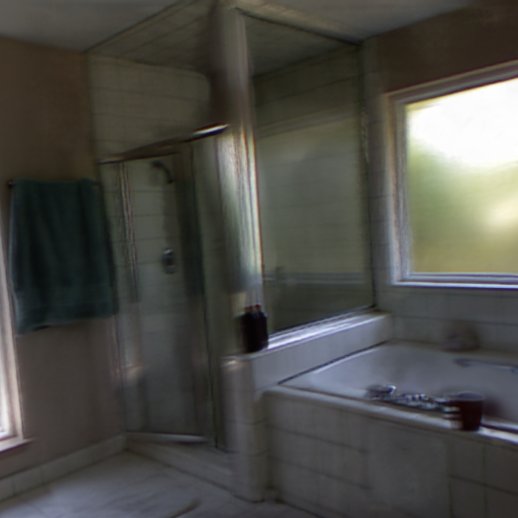} &
        \includegraphics[width=\smallw]{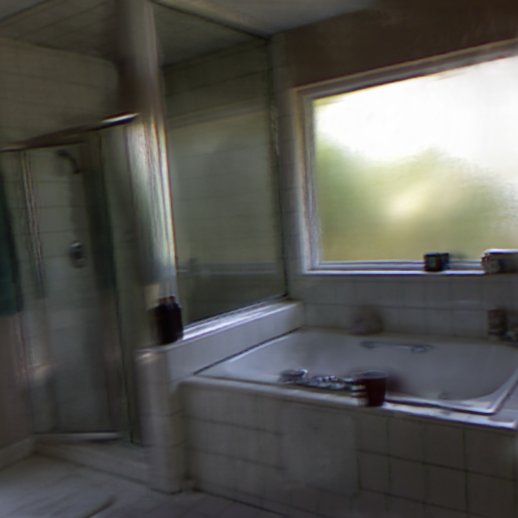} \\

        \noalign{\vspace{3pt}}

        \Block{2-1}{\includegraphics[width=\inputw]{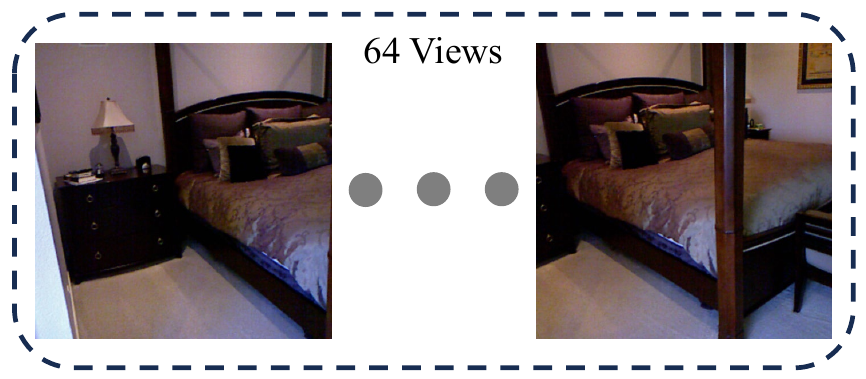}} &
        \Block{2-1}{\includegraphics[width=\bigw]{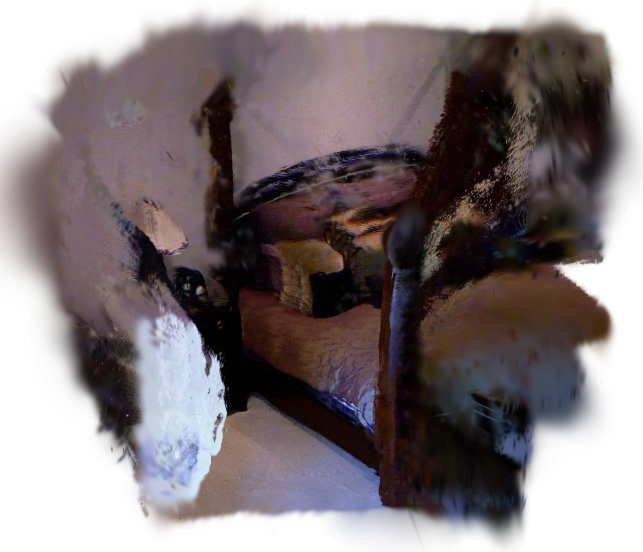}} &
        \includegraphics[width=\smallw]{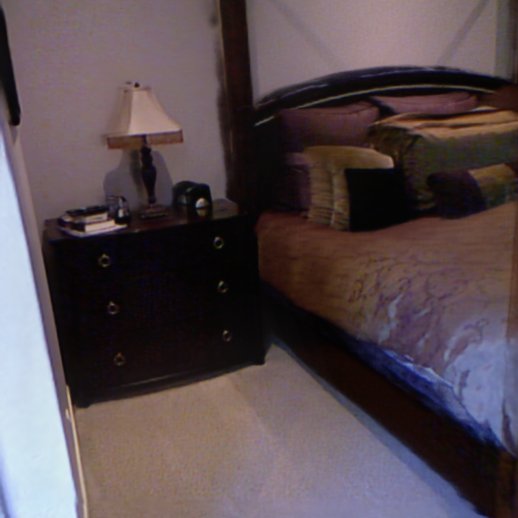} &
        \includegraphics[width=\smallw]{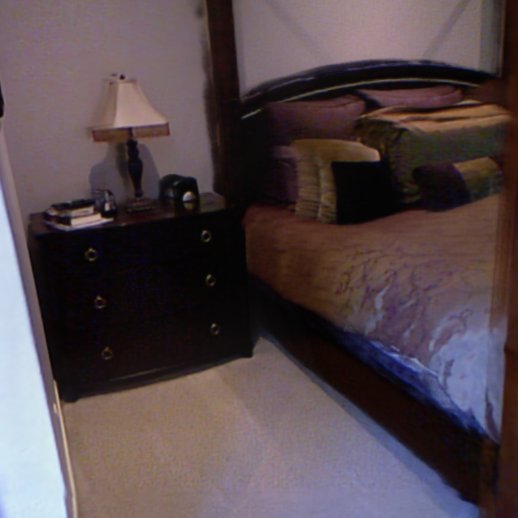} \\
        
         & & 
        \includegraphics[width=\smallw]{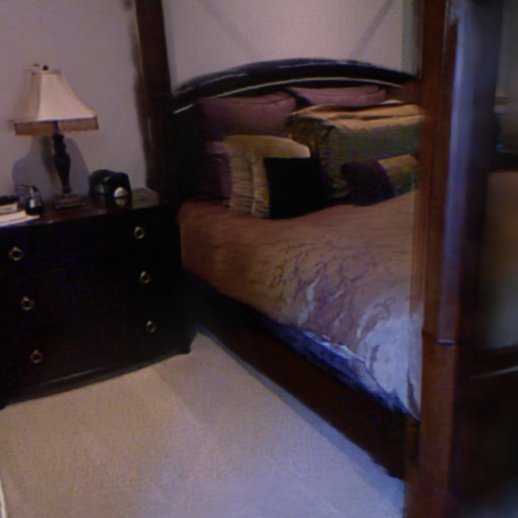} &
        \includegraphics[width=\smallw]{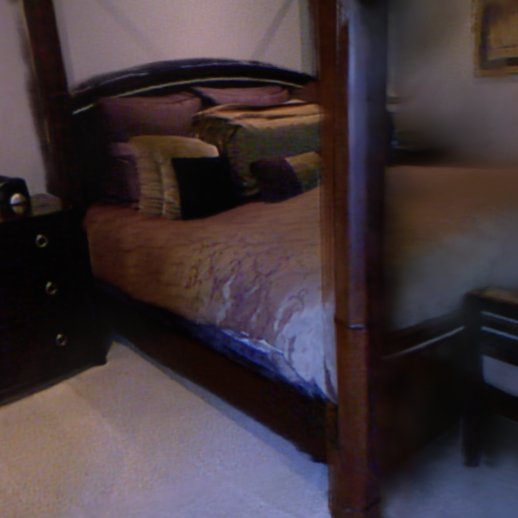} \\

        \noalign{\vspace{3pt}}
        \scriptsize Inputs & 
        \scriptsize Feed-forward Online 3DGS & 
         \\
    \end{NiceTabular}
    
    \caption{Qualitative comparison on \nyu datasets under 64 input views. Each row shows one scene. Left: Input views. Middle: Reconstructed 3DGS. Right: Rendered novel views from ours.}
    \vspace{-15pt}
    \label{fig:fig_nvs_nyuv264}
\end{figure*}

\noindent\textbf{Novel View Synthesis with $64$ Inputs.} 
\Fref{fig:fig_nvs_dl3dv64}, \ref{fig:fig_nvs_re10k64} and \ref{fig:fig_nvs_nyuv264} extend our evaluation on \dlbench, \re and \nyu test datasets to a dense input setting with 64 views. This setting evaluates resilience against catastrophic geometric collapse and error accumulation as redundant observations and state size increase. As illustrated in the middle column, our feed-forward online approach maintains a globally consistent Gaussian representation and avoids severe causal drift or blurring. Consequently, the rendered novel views (right column) preserve photorealistic appearance and structural integrity across different viewing angles.

\clearpage
\twocolumn
\bibliography{main}

\end{document}